\documentclass[11pt]{article}

\PassOptionsToPackage{table}{xcolor}

\usepackage[final]{acl}

\usepackage{times}
\usepackage{latexsym}

\usepackage[T1]{fontenc}

\usepackage[utf8]{inputenc}

\usepackage{microtype}

\usepackage{inconsolata}

\usepackage{graphicx}

\usepackage{amsmath}
\usepackage{amssymb}
\usepackage{amsfonts}
\usepackage{mathtools}
\usepackage{algorithm}
\usepackage{algorithmic}
\usepackage{adjustbox}
\usepackage{booktabs}
\usepackage{multirow}
\usepackage{xcolor}
\usepackage{colortbl}
\usepackage{tikz}
\usepackage{pgfplots}
\usepackage{float}
\usepackage{afterpage}
\usepackage{placeins}
\usepackage{dblfloatfix}
\usepackage{pifont}
\usepackage{multicol}
\usepackage{makecell}
\usepackage[most]{tcolorbox}
\usepackage{titletoc}
\usepackage{enumitem}

\definecolor{slmnavy}{HTML}{1F3A5F}
\definecolor{slmrow}{HTML}{E1E8F3}
\definecolor{slmhl}{HTML}{BFCFEA}
\colorlet{slmblue}{slmnavy}
\colorlet{slmcream}{slmrow}

\arrayrulecolor{slmnavy}
\colorlet{bluerow}{slmrow}
\colorlet{bluecol}{slmhl}
\definecolor{slmmuted}{HTML}{4A5A70}
\newcommand{\prettytable}{\rowcolors{2}{white}{slmrow}}
\newcommand{\hlcell}{\cellcolor{slmhl}}
\newcommand{\cipad}[4]{{\footnotesize\textcolor{slmmuted}{\phantom{#1}[#2,\,#3]\phantom{#4}}}}

\newtcolorbox{promptflow}[1]{
  enhanced, breakable,
  colback=slmcream,
  colframe=slmblue,
  coltitle=white,
  colbacktitle=slmblue,
  fonttitle=\bfseries\small,
  fontupper=\fontsize{7.5pt}{9pt}\selectfont,
  title={#1},
  boxrule=0.8pt,
  titlerule=0pt,
  arc=1.5mm,
  toptitle=2pt, bottomtitle=2pt,
  left=3pt, right=3pt, top=3pt, bottom=3pt,
  before skip=6pt, after skip=8pt
}
\newtcolorbox{promptbox}[1]{
  enhanced,
  colback=slmcream,
  colframe=slmblue,
  coltitle=white,
  colbacktitle=slmblue,
  fonttitle=\bfseries,
  title={#1},
  boxrule=0.8pt,
  titlerule=0pt,
  arc=1.5mm,
  toptitle=3pt, bottomtitle=3pt,
  left=4pt, right=4pt, top=3pt, bottom=3pt
}

\newcommand{\slmjury}{\textsc{SLMJury}}
\newcommand{\yes}{\textcolor[HTML]{3A7D5C}{\ding{51}}}
\newcommand{\no}{\textcolor[HTML]{A84B4B}{\ding{55}}}

\pgfplotsset{compat=1.18}
\renewcommand{\dbltopfraction}{0.80}

\renewcommand{\floatpagefraction}{0.85}
\title{\slmjury{}: Can Small Language Models Judge as Well as Large Ones?}

\author{
  \textbf{Anish Laddha}\textsuperscript{1} \qquad
  \textbf{Nitesh Pradhan}\textsuperscript{1} \qquad
  \textbf{Gaurav Srivastava}\textsuperscript{2} \\
  \mdseries \textsuperscript{1}Department of Computer Science and Engineering, LNMIIT, India \\[-1pt]
  \mdseries \textsuperscript{2}Department of Computer Science, Virginia Tech, USA \\[1pt]
  \mdseries\footnotesize \texttt{anshladdha15@gmail.com} \qquad
  \texttt{nitesh.pradhan@lnmiit.ac.in} \qquad
  \texttt{gks@vt.edu} \\[1pt]
  \mdseries\footnotesize \textbf{Code:} \href{https://github.com/anishh15/SLMJury}{\texttt{github.com/anishh15/SLMJury}} \qquad
  \textbf{Leaderboard:} \href{https://anishh15.github.io/SLMJury/}{\texttt{anishh15.github.io/SLMJury}} \\[-3pt]
  \mdseries\footnotesize \textbf{PyPI:} \href{https://pypi.org/project/slmjury/}{\texttt{pypi.org/project/slmjury}}
}

\begin{document}
\raggedbottom
\maketitle
\begin{abstract}
Large language model judges are costly, slow, and opaque. We introduce \slmjury{}\footnote{Code, leaderboard, and \texttt{pip} package: \url{https://github.com/anishh15/SLMJury}}, a framework for evaluating small language models (SLMs) as budget-conditioned judges of \textbf{closed-ended} correctness and \textbf{open-ended} quality. We compare \textbf{16 SLM judges} (0.6B--14B, four families) with \textbf{six large open-weight judges} (70B--122B) under one pipeline on \textbf{ten benchmarks} (64{,}824 closed-ended judgments per configuration). \textbf{(1)} On reference-guided verification, Phi-4 (14B) is equivalent to the strongest large judge, Qwen3.5-122B, within $\pm$0.5 points and significantly more accurate than the other five; accuracy saturates near 8--14B, and its quick verdict costs an estimated 9 to 32 times less. \textbf{(2)} Overthinking is \textit{domain-dependent} and prompt-mediated: budget alone leaves accuracy unchanged for 11 of 13 judges. \textbf{(3)} Math-to-general gaps reach 38 points and stay at 8.5--12 even for the strongest judges (14B--122B). \textbf{(4)} On 1,600 human-annotated summaries, the top SLMs match or beat large judges in 25 of 30 comparisons, yet closed- and open-ended rankings diverge. \textbf{(5)} Debate degrades accuracy in every configuration; the top two judges hold aggregate accuracy within a point across six adversarial personas. \textbf{(6)} Judges rank students by solver accuracy, without lineage preference at the quick verdict. Reference-guided evaluation does not need large models, yet no single SLM dominates.
\end{abstract}

\section{Introduction}

Large language models (LLMs) are now the default automated judges for evaluating model outputs \citep{zheng2023judging, kim2024prometheus, zhu2025judgelm}. Systems such as MT-Bench \citep{zheng2023judging} and JudgeLM \citep{zhu2025judgelm} score student solutions with proprietary models like GPT-4 \citep{achiam2023gpt}, reaching high agreement with human annotations. The cost structure is the problem. Judging $N$ solutions at token budget $B$ costs $C_\text{total} = N (\alpha B_{\text{in}} + \beta B)$ for per-token prices $\alpha, \beta$ and prompt length $B_{\text{in}}$, which at evaluation scale ($N {>} 10^4$) becomes a bottleneck for RLHF \citep{ouyang2022training} and constitutional AI \citep{bai2022constitutional}. Proprietary judges also send inputs off the local network, change version silently, and add latency \citep{wang2024pandalm, chen2024humans, szymanski2025limitations}.

\begin{table*}[t]
\centering
\small
\rowcolors{2}{white}{slmrow}
\begin{adjustbox}{max width=\textwidth}
\begin{tabular}{lccccccc c}
\toprule
\textbf{Feature} & \textbf{MT-Bench} & \textbf{JudgeLM} & \textbf{Prometheus\,2} & \textbf{JudgeBoard} & \textbf{JudgeBench} & \textbf{PoLL} & \textbf{Explicit Reason.} & \hlcell\hspace*{.5\fontdimen2\font}\textbf{\slmjury}\hspace*{.5\fontdimen2\font} \\
\midrule
Off-the-shelf SLM judges ($\leq$14B)\textsuperscript{$\dagger$} & \no & \no & \no & \yes & \no & \no & \yes & \hlcell \textbf{\yes} \\
Binary correctness eval. & \no & \no & \no & \yes & \yes & \yes & \no & \hlcell \textbf{\yes} \\
Open-ended quality scoring & \yes & \yes & \yes & \no & \no & \yes & \no & \hlcell \textbf{\yes} \\
Token budget analysis & \no & \no & \no & \no & \no & \no & \yes & \hlcell \textbf{\yes} \\
Persona robustness & \no & \no & \no & \no & \no & \no & \no & \hlcell \textbf{\yes} \\
Ensemble / majority voting & \no & \no & \no & \no & \no & \yes & \no & \hlcell \textbf{\yes} \\
Multi-agent debate & \no & \no & \no & \yes & \no & \no & \no & \hlcell \textbf{\yes} \\
Multi-family evaluation & \no & \no & \no & \yes & \yes & \yes & \no & \hlcell \textbf{\yes} \\
\midrule
Num.\ judge models & 1 & 3 & 2 & 8 & 14 & 3 & 3 & \hlcell \textbf{22} \\
Num.\ benchmarks & 1 & 2 & 8 & 5 & 1 & 6 & 1 & \hlcell \textbf{10} \\
\bottomrule
\end{tabular}
\end{adjustbox}
\vspace{-2pt}
{\raggedright\scriptsize\textsuperscript{$\dagger$}Non-fine-tuned models with $\leq$14B parameters. JudgeLM and Prometheus\,2 train \emph{custom} judges (7--33B and 7B/8$\times$7B); PoLL uses off-the-shelf models but includes proprietary LLMs (GPT-3.5, Claude-3-Haiku) rather than SLMs. \slmjury{}'s 22 judges are 16 SLMs and six large open-weight judges (70B--122B).\par}
\caption{\textbf{Feature comparison of \slmjury{} with related LLM-as-a-Judge work.} To our knowledge, \slmjury{} is the first framework to evaluate off-the-shelf SLM judges across all of these dimensions in one place. See Appendix~\ref{app:datasets} for dataset details and Appendix~\ref{app:models} for model configurations.}
\label{tab:related_comparison}
\end{table*}

Small language models (SLMs) attack both terms at once. Phi-4 \citep{abdin2024phi}, Phi-4-mini \citep{abouelenin2025phi}, Phi-4-Reasoning \citep{abdin2025phi}, Qwen2.5 \citep{qwen2025qwen25technicalreport}, Qwen3 \citep{yang2025qwen3}, and Llama 3 \citep{grattafiori2024llama} reason well at a fraction of the cost on a single local GPU, and prior work shows them matching LLM-level performance on reasoning \citep{srivastava-etal-2025-thinkslm}, contamination-resistant evaluation \citep{ICLR2026_22b4e30a}, and agentic generation \citep{pmlr-v306-srivastava26a, latimer2025hindsight2020buildingagent}. \textit{If SLMs can reason, can they also judge?} Despite growing interest in LLM-as-a-judge \citep{bi2026judgeboard, jayarao2025explicit, li2026rethinking}, no study has evaluated SLM judges across task domains, paradigms, token budgets, persona conditions, and debate protocols together. Nor can the question behind it, how much accuracy a small judge gives up, be answered without running large judges through the same pipeline.

\begin{figure*}[t]
\centering
\includegraphics[width=\textwidth]{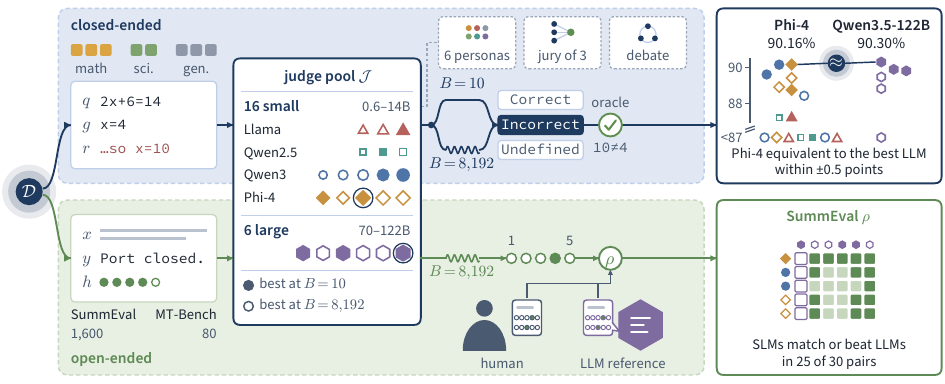}
\caption{\textbf{Overview of \slmjury{}.} Closed-ended items $(q, g, r)$ from 8 benchmarks and open-ended SummEval and MT-Bench items $(x, y, h)$ go to one judge pool $\mathcal{J}$ of 22 judges (16 SLMs, 0.6B--14B; six large, 70B--122B). Closed-ended judges give a quick ($B{=}10$) or reasoned ($B{=}8{,}192$) verdict, checked by a programmatic oracle ($N{=}64{,}824$ judgments per configuration). The SLMs also run with six personas, in juries of three, and in debate (the dashed boxes). Open-ended judges reason at $B{=}8{,}192$ and score 1--5, compared by Spearman $\rho$ with human scores (SummEval) or an LLM reference, Qwen3.5-397B-A17B (MT-Bench). \textit{Top right}: each judge's best accuracy (filled = best at $B{=}10$, open = best at $B{=}8{,}192$); Phi-4 (90.16\%) and Qwen3.5-122B (90.30\%) are equivalent within $\pm$0.5 points, and Phi-4 costs an estimated 32 times less to serve. \textit{Bottom right}: SummEval comparisons of the five strongest SLMs with the six large judges (dark = SLM higher, light = not distinguishable, outlined = large judge higher). Both example items are illustrative.}
\label{fig:overview}
\end{figure*}

\slmjury{} closes that gap across two paradigms. For \textbf{closed-ended} evaluation a judge performs reference-guided verification: it is a function $f_{\mathcal{J}}^{(B)} : \mathcal{Q} \times \mathcal{G} \times \mathcal{R} \rightarrow \mathcal{V}$ mapping a question, a reference, and student reasoning to a verdict under token budget $B$. For \textbf{open-ended} evaluation judges assign numerical scores, correlated against human annotations (SummEval) or a large reference scorer (MT-Bench). We center $\mathcal{J}$ on models with $|\theta_\mathcal{J}| \leq 14\text{B}$ and compare them with six open-weight judges at 70B--122B, run under the identical pipeline, students, prompts, and oracle, at two budgets, a quick and a reasoned verdict ($B \in \{10, 8192\}$), across the evaluation dimensions of Figure~\ref{fig:overview}. Our contributions:

\textbf{(1)} We benchmark \textbf{16 SLM judges} from four families spanning 0.6B--14B against six large open-weight judges (70B--122B) from five developers, across \textbf{ten benchmarks} (12{,}029 model runs; $N{=}64{,}824$ judgments per configuration on the eight closed-ended ones). On reference-guided verification the best 14B judge, Phi-4 (90.16\%), is equivalent to the strongest 70B--122B judge, Qwen3.5-122B, within $\pm$0.5 points, and the other five are significantly less accurate. Within the Qwen3 family accuracy saturates near 8--14B, and a 14B judge costs an estimated 9 to 32 times less to serve than the large judges that can give a quick verdict (\S\ref{sec:individual}, \S\ref{sec:scaling}). \textbf{(2)} We show the overthinking effect is \textit{domain-dependent} and locate its cause by single-factor decomposition: budget alone leaves accuracy unchanged for 11 of 13 judges, while the reasoned prompt at ten tokens collapses the parseable-verdict rate and accuracy with it (\S\ref{sec:individual}). \textbf{(3)} We give the first cross-paradigm analysis of the same SLM judges: Phi-4 leads the SLMs on binary judging and on human-annotated SummEval but ranks sixth on MT-Bench, where its reasoning-trained sibling Phi-4-Reasoning scores significantly higher while trailing it on SummEval, so the paradigms draw on different capabilities (\S\ref{sec:openended}). \textbf{(4)} We probe these findings from five directions (adversarial personas, prompt formulations, alternative protocols, a widened student pool with a strong same-family student, and multi-agent debate): the aggregate accuracy of Qwen3-14B and Phi-4 moves under a point across personas, judge rankings hold across prompt formulations, the best small judges keep pace with large ones when the reference is withheld, no family favors its own lineage at the quick verdict, and debate is the only collaborative strategy that reliably hurts (\S\ref{sec:robustness}, \S\ref{sec:debate}, \S\ref{sec:scaling}). Reliable reference-guided evaluation does not require large models, yet no single SLM excels everywhere.

\section{Related Work}
\label{sec:related}

\textbf{LLM- and SLM-as-a-Judge.} MT-Bench and Chatbot Arena \citep{zheng2023judging} established GPT-4 as a proxy for human evaluation; JudgeLM \citep{zhu2025judgelm}, Prometheus \citep{kim2024prometheus}, and PandaLM \citep{wang2024pandalm} explored open alternatives, while SummEval \citep{fabbri2021summeval} and MT-Bench supply the annotated benchmarks for open-ended scoring. LLM judges still degrade on expert tasks \citep{szymanski2025limitations} and carry position and verbosity biases \citep{chen2024humans, gu2024survey, li2025generation, yang2026any}, driving a turn toward smaller judges: JudgeBoard \citep{bi2026judgeboard} benchmarks SLMs for reasoning, JudgeBench \citep{tan2025judgebench} shows GPT-4o near-random on hard pairs, and \citet{wang2026small} report small reward models beating LLM-as-a-judge baselines, alongside representation-based judging \citep{li2026rethinking}, code \citep{crupi2026improving, jiang2025codejudgebench}, contextual settings \citep{xu2025does}, and training-based gains \citep{xu2025j4r, zhang2025compassjudger}. Closest to us, \citet{jayarao2025explicit} study reasoning effects on Qwen3 judges. \textit{We show the token-budget effect is domain-dependent for the 13 small judges that run at both budgets, evaluate 16 small judges on ten benchmarks spanning both paradigms, and compare them with 70B--122B judges run under the same pipeline, using paired tests on identical items.}

\textbf{Token Budget, Overthinking, and Prompt Sensitivity.} Longer reasoning does not always help: \citet{han2025token} introduce token-budget-aware prompting, \citet{wang2024reasoning} tie strategy efficiency to task type, and \citet{srivastava-etal-2026-llms} show extended chain-of-thought hurting smaller models on math, alongside reports of diminishing returns from test-time scaling \citep{sui2025stop, ma2025reasoning, ghosal2026does, shojaee2026illusion, aggarwal2025optimalthinkingbench}. Judges are also sensitive to framing, and persona prompts shift their behavior \citep{tseng2024two, salewski2023context, abdullahi2026persona, ryan2025synthesizeme, laddha2026positionrolelabelcontrolled}. \textit{We bring both lenses to judging, separate the budget from its prompt, and find the strongest judges' accuracy stable under six personas, with judge rankings stable across prompt reformulations.}

\textbf{Ensembles and Multi-Agent Debate.} Self-consistency \citep{wang2022self} and repeated sampling \citep{brown2024large} improve reasoning, SLM ensembles can match larger models \citep{cho2025cosmosfl}, and PoLL \citep{verga2024replacing} shows panels outperforming one large judge; debate improves factual accuracy and divergent thinking \citep{du2024improving, liang2024encouraging}, with ChatEval \citep{chan2024chateval} applying it to open-ended evaluation and \citet{srivastava-etal-2025-debate} introducing RCR \citep{hu2026multi, lin2025efficient, alam2026beyond}. \textit{We find ensembles add $+$0.11 points once accuracy saturates near 90\%, and debate degrades binary judging \citep{wynn2025talk, zhang2025stop}.} Table~\ref{tab:related_comparison} positions \slmjury{} against this work.

\section{The \slmjury{} Framework}
\label{sec:method}

\slmjury{} is a modular framework for studying SLMs as judges across closed-ended and open-ended paradigms (Figure~\ref{fig:overview}). We fix notation: $\mathcal{Q}$, $\mathcal{G}$, $\mathcal{R}$ denote the spaces of questions, ground-truth references, and student responses, and $\mathcal{V} = \{\texttt{Correct}, \texttt{Incorrect}, \texttt{Undefined}\}$ the verdict space. The judges under study have $|\theta_{\mathcal{J}}| \le 14\text{B}$ parameters and a token budget $B \in \{10, 8192\}$; larger judges run through the same function for comparison. We describe the pipeline, judge inference, metrics, evaluation approaches, and open-ended scoring below; the full formal treatment is in Appendix~\ref{app:framework}.

\subsection{Evaluation Pipeline}
\label{sec:pipeline}

The pipeline runs in three stages: student solution generation, judge evaluation, and metric computation. A \textit{student model} $\mathcal{S}$ solves problems from benchmark $\mathcal{D}$ under $\tau_s {=} 0.7$, top-$p {=} 0.9$, and $B_s {=} 1{,}024$ tokens, using dataset-specific templates (Appendix~\ref{app:prompts}) and vLLM with PagedAttention \citep{kwon2023efficient}. Each solution is stored with its problem, ground truth, and response (Appendix~\ref{app:student}).

\subsection{Judge Inference and Token Budget}
\label{sec:judge}

Following the reference-based verification paradigm of MT-Bench \citep{zheng2023judging}, JudgeLM \citep{zhu2025judgelm}, and PoLL \citep{verga2024replacing}, each judge $\mathcal{J}$ maps a question, a ground-truth reference, and a student reasoning chain $(q, g, r_s)$ to a verdict $v \in \mathcal{V} = \{\texttt{Correct}, \texttt{Incorrect}, \texttt{Undefined}\}$, with $g = r^*$ when gold reasoning ships with the dataset and $g = a^*$ otherwise. The closed-ended task is therefore reference-guided verification, the setting our closed-ended conclusions address. The budget-conditioned judge function is the composition
\begin{equation}
    f_{\mathcal{J}}^{(B)} = \textsc{Parse} \circ \mathcal{D}_B \circ \mathcal{T}_B
    \;:\; \mathcal{Q} \times \mathcal{G} \times \mathcal{R} \rightarrow \mathcal{V},
    \label{eq:judgefn}
\end{equation}
where $\mathcal{T}_B$ formats the prompt, $\mathcal{D}_B$ decodes at most $B$ tokens, and $\textsc{Parse}$ extracts the verdict through the cascade of Appendix~\ref{app:verdict_cascade}. At the \textbf{quick verdict} setting ($B{=}10$) the judge answers in one word under deterministic decoding, attending only to the final answer; at the \textbf{reasoned verdict} setting ($B{=}8{,}192$) it reasons step by step before emitting a boxed verdict. Models with native chain-of-thought run with thinking enabled at $B{=}8{,}192$ and sample ($\tau{=}0.6$, top-$p{=}0.95$, top-$k{=}20$), except GPT-OSS-120B, which decodes greedily. Always-thinking models cannot answer within ten tokens, so they are evaluated at $B{=}8{,}192$ only and excluded from $\Delta$ comparisons; serving details for the large judges are in Appendix~\ref{app:models}.

\paragraph{Single-factor decomposition.} The two settings differ in more than budget: prompt format, thinking mode, and temperature move together, so attributing the accuracy delta to any one of them requires holding the rest fixed. From the $B{=}10$ baseline we evaluate four configurations: three change exactly one factor (budget, prompt format, temperature), and the fourth turns on thinking mode, which brings the full budget and sampling with it (for the eight judges without a thinking mode it measures sampling at the full budget). A nine-point budget sweep runs from 10 to 8{,}192 tokens (\S\ref{sec:individual}, Appendix~\ref{app:budget_factorial}).

\subsection{Ground Truth and Metrics}
\label{sec:metrics}

Verdicts are scored against a programmatic oracle: the oracle verdict is \texttt{Correct} if $\textsc{Equiv}(\hat{a}, a^*, \delta) = \top$, where $\hat{a}$ is the student's extracted answer (Appendix~\ref{app:answer_extraction}), $\delta{=}10^{-9}$ is the numeric tolerance, and $\textsc{Equiv}$ is dataset-type-specific (normalized numeric, symbolic for MATH, label matching for multiple choice; Appendix~\ref{app:normalization}). The 437 student answers the extractor cannot read (0.67\%, the same for every judge) are scored \texttt{Incorrect}; excluding them moves no judge, at its best budget, by more than 0.33 points (Appendix~\ref{app:oracle_reliability}). \textbf{Accuracy} measures agreement with the oracle:
\begin{equation}
    \text{Acc}(\mathcal{J}, B) = \frac{1}{N} \sum_{i=1}^{N} \mathbf{1}\!\left[f_{\mathcal{J}}^{(B)}(q_i, g_i, r_{s,i}) = v_i^*\right]
\end{equation}
\textbf{Instruction Following Rate (IFR)} is the fraction of parseable responses. Both are computed over $N {=} 64{,}824$ judgments (2 students $\times$ 8 datasets) per configuration and micro-averaged, with macro averages in Appendix~\ref{app:large_judges}.

\paragraph{Comparing judges.} Every judge scores the same items, so comparisons are paired \citep{dror2018hitchhiker}: McNemar's exact test on the discordant items \citep{mcnemar1947note}, a 95\% bootstrap interval that resamples items within each dataset (10{,}000 resamples), and two one-sided tests (TOST; \citealp{schuirmann1987comparison}) for equivalence within $\pm$0.5 points, with Benjamini--Hochberg correction \citep{benjamini1995controlling} over all 231 judge pairs. We call two judges \emph{equivalent} when TOST rejects and \emph{not distinguishable} when McNemar does not.

\subsection{Evaluation Approaches}
\label{sec:strategies}

Every judge is evaluated independently at each budget it supports, and we then vary the setup. \textbf{Personas}: six adversarial system prompts (\textit{Strict}, \textit{Lenient}, \textit{Industry}, \textit{Logic}, \textit{Safety}, \textit{Helpfulness}) with everything else fixed (Appendix~\ref{app:personas}). \textbf{Ensembles}: the top-5 judges at their best budget, over all $\binom{5}{3} {=} 10$ three-model juries, taking the majority verdict with ties resolved to \texttt{Undefined} (Appendix~\ref{app:ensemble}). \textbf{Debate}: the RCR (Reflect-Critique-Refine) protocol of \citet{srivastava-etal-2025-debate}, where three agents give independent verdicts then critique and refine over up to $T{=}5$ rounds, in a cross-architecture variant ($\tau{=}0$) and a same-model variant (Appendix~\ref{app:debate}).

\paragraph{Prompt formulations and protocols.} We also re-run judges under five formulations of the closed-ended instruction (the original, a stripped-down minimal version, a phrasing in the style of MT-Bench's GPT-4 judge prompt, a rubric-based variant, and a chain-of-thought variant), under two stress variants that rewrite it rather than reword it (an adversarial framing, a few-shot prompt), and under three alternative protocols: reference-free judging with the gold answer withheld, G-Eval-style rubric scoring, and pairwise comparison in both answer orders, the last two with the reference still in the prompt (\S\ref{sec:robustness}).

\subsection{Open-Ended Evaluation}
\label{sec:openended_method}

Each judge also realizes a scoring map $s_{\mathcal{J}}: \mathcal{X} \rightarrow \{1,\dots,5\}$ rating a summary or conversation on a five-point scale. Judges score at $B{=}8{,}192$, with thinking and sampling as in \S\ref{sec:judge}. We report the Spearman rank correlation $\rho$ between a judge's scores and a human (SummEval) or large-model (MT-Bench) reference, with a bootstrap 95\% interval, and Pearson $r$, Cohen's $\kappa$, accuracy, and MSE as secondary measures (Appendix~\ref{app:scoring_metrics}).

On \textbf{SummEval} \citep{fabbri2021summeval} judges score 1,600 article-summary pairs on coherence, consistency, fluency, and relevance against human annotations, giving an SLM-human correlation averaged over the four dimensions. On \textbf{MT-Bench} \citep{zheng2023judging} they score 80 multi-turn conversations on overall quality, referenced against Qwen3.5-397B-A17B, a large model that is not itself a judge in the study, applied to both students' responses; a second scorer, GPT-OSS-120B \citep{agarwal2025gptoss}, measures how reliable that reference is. Reference scorers and SLM judges use the \emph{identical} scoring prompt, so the correlation isolates capability rather than prompt design (Appendix~\ref{app:scoring_prompts}).

\section{Results and Insights}
\label{sec:results}

Throughout, accuracy is measured against the oracle of \S\ref{sec:metrics}, so a judge's leniency toward any student lowers its accuracy rather than inflating it. Where the oracle and a judge disagree, the judge is usually right: a manual audit of sampled disagreements sides with the judge in most cases for every audited judge, small and large (Appendix~\ref{app:oracle_audit}). The same oracle scores every judge, so the comparisons below are like for like.

\subsection{Evaluation Setup}
\label{sec:setup}

We evaluate 16 small judges across 2 students and 8 closed-ended datasets spanning mathematical reasoning (GSM8K, GSM-Plus, MATH), scientific reasoning (ARC-Easy, ARC-Challenge), and general reasoning (HellaSwag, WinoGrande, TruthfulQA), giving $N {=} 64{,}824$ judgments per configuration (Appendix~\ref{app:datasets}). The students are Qwen2.5-32B-GPTQ-Int8 (95.45\% solver accuracy on GSM8K) and Llama-3.1-8B-Instruct (82.49\%), a strong and a weaker source (Appendix~\ref{app:student}). Six large open-weight judges from five developers run under the identical pipeline: Llama-3.3-70B, DeepSeek-R1-Distill-Llama-70B \citep{deepseek2025r1}, Qwen2.5-72B, GLM-4.5-Air (106B; \citealp{glm2025glm45}), GPT-OSS-120B, and Qwen3.5-122B-A10B. Qwen3-32B fills the one gap in the Qwen3 family between 14B and the large judges and is reported separately. \S\ref{sec:scaling} widens the student pool to five families. Results average over both students and all datasets unless stated otherwise (Appendices~\ref{app:per_dataset} and~\ref{app:implementation}).

\subsection{Individual Judging and Token Budget Effects}
\label{sec:individual}

\begin{table*}[t]
\centering
\begin{adjustbox}{width=\textwidth,center}
\small
\rowcolors{2}{white}{slmrow}
\begin{tabular}{llccccccc}
\toprule
\hiderowcolors
& & &
    \multicolumn{2}{c}{\textbf{Accuracy (\%)}} &
    &
    \multicolumn{2}{c}{\textbf{IFR (\%)}} \\
\cmidrule(lr){4-5} \cmidrule(lr){7-8}
\textbf{Family} & \textbf{Judge Model} & \textbf{Parameters}
    & \textbf{10 tokens} & \textbf{8{,}192 tokens}
    & \textbf{$\Delta$ (points)}
    & \textbf{10 tokens} & \textbf{8{,}192 tokens} \\
\midrule
\noalign{\global\rownum=2}\showrowcolors
    & Llama-3.2-1B-Instruct          & 1B   & 43.25 & 65.61 & $-$22.36 & \phantom{0}94.02 & \phantom{0}97.41 \\
    & Llama-3.2-3B-Instruct          & 3B   & 79.92 & 82.14 & \phantom{0}$-$2.22 & \textbf{100.00} & \phantom{0}99.92 \\
\multirow{-3}{*}{Llama 3.x}
    & Llama-3.1-8B-Instruct          & 8B   & 87.29 & 84.20 & \phantom{0}$+$3.10 & \phantom{0}99.98 & \phantom{0}99.79 \\
\midrule
    & Qwen2.5-1.5B-Instruct          & 1.5B & 74.32 & 80.39 & \phantom{0}$-$6.07 & \textbf{100.00} & \phantom{0}99.21 \\
    & Qwen2.5-3B-Instruct            & 3B   & 78.16 & 77.22 & \phantom{0}$+$0.93 & \phantom{0}99.99 & \phantom{0}99.80 \\
\multirow{-3}{*}{Qwen2.5}
    & Qwen2.5-7B-Instruct            & 7B   & 78.01 & 87.23 & \phantom{0}$-$9.22 & \phantom{0}99.99 & \phantom{0}99.94 \\
\bottomrule
\end{tabular}
\end{adjustbox}
\caption{\textbf{Individual judge performance.} Accuracy and IFR (\%) averaged over 2 student models and 8 datasets ($N{=}64{,}824$ per configuration). \textbf{Bold} = column best. $^\dagger$ = always-thinking models (8,192 tokens only). $\Delta$ = Acc($B{=}10$) $-$ Acc($B{=}8{,}192$); positive $\Rightarrow$ quick verdict wins.}
\label{tab:individual}
\end{table*}

\textsc{Small Models Are Capable Judges Across Domains.} \textbf{A 14B open judge (Phi-4, 90.16\% at $B{=}10$) averages just over 90\% agreement with the oracle across eight benchmarks spanning three domains, without any judge-specific fine-tuning.} Qwen3-14B follows at 90.15\%, the best sub-5B model (Phi-4-mini) reaches 88.75\%, and Qwen3-4B at 4B comes within 1.72 points of the best SLM at its better budget (Table~\ref{tab:individual}). For the top models domain matters more than scale: they score 96--98\% on math but only 77--80\% on HellaSwag, which demands commonsense rather than verifiable computation.

\textsc{The Overthinking Effect Is Domain-Dependent.} \textbf{Quick verdicts beat extended reasoning on mathematical judging for 8 of 13 dual-setting judges, yet reasoning wins on general tasks for 8 of 13, so the two effects partially cancel in any single-number summary.} Write $\Delta(\mathcal{J}, \mathcal{D}) = \text{Acc}(\mathcal{J}, 10) - \text{Acc}(\mathcal{J}, 8192)$ on domain $\mathcal{D}$, with domain accuracy averaged over its items. On \textit{math}, quick verdicts win for 8 of 13 judges and gain up to 10.6 points where they help, since correctness is structurally verifiable and deliberation has limited upside. On \textit{general} tasks the sign flips for 8 of 13; reasoning lifts Qwen3-4B by 23.09 points on WinoGrande ($71.47 \to 94.55$), where commonsense inference is hard to settle in ten tokens. Because $\Delta$ flips sign \textit{within the same judge} across domains, task structure drives the effect rather than model capability (Appendix~\ref{app:per_dataset}).

\begin{figure*}[t]
\centering
\includegraphics[width=\textwidth]{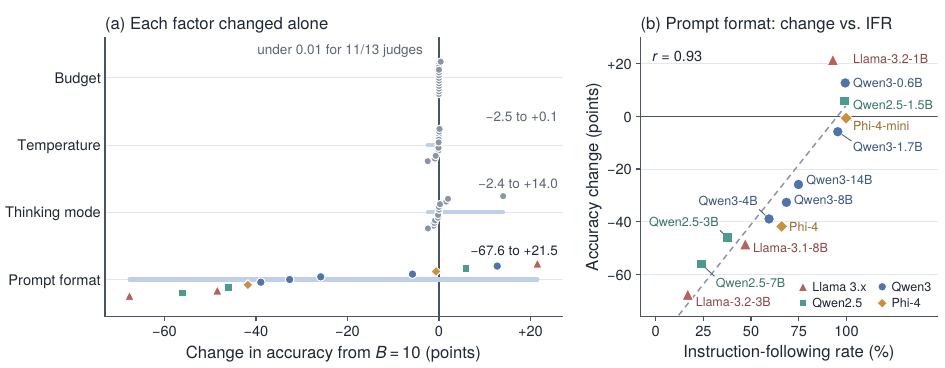}
\caption{\textbf{Single-factor decomposition of the $B{=}10 \to B{=}8{,}192$ delta} (13 judges, all eight datasets, 64{,}824 judgments per cell). \textit{Left}: each change from the $B{=}10$ baseline, one dot per judge. Budget moves to 8{,}192 tokens with the quick prompt held fixed; prompt format moves to the reasoned prompt with the ten-token budget held fixed; thinking mode brings the full budget and sampling with it. Budget alone leaves accuracy unchanged to two decimals for 11 of 13 judges and moves it by at most 0.37 points, while prompt format moves it by tens of points. \textit{Right}: the prompt-only condition cuts a reasoned response off at ten tokens, and the accuracy change tracks the fraction of parseable outputs ($r{=}0.93$), so the collapse is an instruction-following failure.}
\label{fig:factor_decomposition}
\end{figure*}

\textsc{The Delta Comes From the Prompt, Not the Budget.} \textbf{Raising the token budget with the quick-verdict prompt held fixed changes almost nothing: accuracy is unchanged for 11 of 13 judges, verdicts are identical for 9, and the largest movement is $+0.37$ points.} The single-factor decomposition of \S\ref{sec:judge} settles what the aggregate $\Delta$ cannot (Figure~\ref{fig:factor_decomposition}, left). Temperature and thinking mode each shift accuracy by at most 2.5 points, except thinking mode for the smallest judge (Qwen3-0.6B, $+14.0$). The prompt is the active factor, and it acts through instruction following: at ten tokens a reasoned response is cut off mid-explanation, so for the eight judges that explain before answering, parseable verdicts fall to 17--75\% and accuracy follows in near-lockstep ($r{=}0.93$; Figure~\ref{fig:factor_decomposition}, right). In a nine-point sweep with the reasoned prompt, most judges score lowest at 64--256 tokens, where the explanation is cut short, and 12 of the 14 swept judges are back on plateau by 512 (Appendix~\ref{app:budget_factorial}). Overthinking is therefore a prompt$\times$budget interaction, and the domain-dependent sign reversal survives, since a constant prompt or decoding difference cannot flip sign within one judge.

\textsc{Domain Generalization Separates Model Families.} \textbf{The math-to-general accuracy gap runs from 9 points for the strongest judges at any scale to 38 for Qwen2.5-7B, so this gap, more than raw accuracy, separates a broadly applicable judge from a domain specialist.} Here each domain is the unweighted mean of its three datasets. Qwen2.5 sits at the specialist end, with Qwen2.5-7B at 97.1\% on math but 58.7\% on general reasoning, which we attribute to math-heavy instruction tuning. Phi-4, Qwen3-8B and Qwen3-14B hold it between 9 and 12 points and stay reliable on benchmarks they were not optimized for; the per-dataset breakdown is in Appendix~\ref{app:per_dataset}. Instruction following exceeds 99.8\% for every top judge (Table~\ref{tab:individual}), so these gaps reflect disagreement with the oracle, not unparseable output.

\subsection{Robustness Analysis}
\label{sec:robustness}
\label{sec:persona}

\textsc{Well-Trained Judges Resist Persona Manipulation.} \textbf{Qwen3-14B's aggregate accuracy at $B{=}10$ moves by only 0.10 points across all six adversarial personas (64{,}824 judgments each), so the personas we test do not shift a strong judge's accuracy.} Measuring robustness by the range of aggregate accuracy over the six personas, the top judges cluster tightly across all eight benchmarks: Qwen3-14B at 0.10, Phi-4 at 0.56, Qwen3-4B at 1.41 points (Appendix~\ref{app:personas}). They appear to hold a stable correctness boundary rather than a heuristic that surface framing can flip.

\textsc{Lenient and Strict Are the Attack Surfaces.} \textbf{The Lenient persona collapses Llama-3.1-8B by 10.54 points ($87.29 \to 76.76$) but moves Qwen3-14B by only 0.10 points from its baseline.} A weak judge reads ``give partial credit'' as permission to accept wrong answers. With reasoning, the Strict persona costs Phi-4-mini 11.90 points as it second-guesses correct answers, and the Phi-4 judges' persona ranges widen ($2.5\times$ for Phi-4, $10.9\times$ for Phi-4-mini).

\textsc{Rankings Hold Across Prompt Formulations.} \textbf{Across five formulations of the judging instruction, judge rankings stay stable (mean pairwise Kendall's $\tau$ of 0.75 at $B{=}10$, rising to 0.83 for the same judges once they can reason), and at $B{=}10$ the larger Qwen3 judges move least: Qwen3-14B by 1.22 points, Qwen3-8B by 1.30, Qwen3-4B by 1.64.} We re-run every judge under the five formulations of \S\ref{sec:strategies}, each scored on the identical 64{,}824 items (Appendix~\ref{app:prompt_sensitivity}). Once they reason, Phi-4-Reasoning and Phi-4-Reasoning-Plus move by under a point (0.90 and 0.99), while Qwen2.5-7B and Llama-3.2-1B move by 10.7 and 14.4 at $B{=}10$. Qwen3-14B stays in the top three under every formulation. Open-ended rankings are just as stable, at $\tau{=}0.85$ on SummEval and 0.76 on MT-Bench. Rewriting the instruction costs far more than rewording it: adversarial framing and few-shot exemplars take Llama-3.1-8B down 41.8 and 48.4 points at $B{=}10$, and the reasoning budget absorbs most of both, bringing it back within 0.60 points; few-shot breaks its output format, adversarial its verdict.

\textsc{Small Judges Keep Pace Without a Reference.} \textbf{Reference-based pointwise judging is the strongest protocol for 17 of the 18 judges tested, and without the reference the best SLMs land between the two large judges: Qwen3-14B reaches 80.8\%, against 77.2\% for Qwen2.5-72B and 82.2\% for GPT-OSS-120B.} We run all 16 SLMs and two large judges under reference-free judging, G-Eval-style rubric scoring, and pairwise comparison in both answer orders (Appendix~\ref{app:protocols}), each judge at its best budget. Withholding the reference costs every judge (Qwen2.5-72B 12.7 points, GPT-OSS-120B 7.2, the SLMs 10.2 on average), and across the SLMs the drop does not shrink with size (Spearman correlation between parameters and drop magnitude $+0.44$), so it measures how much of the task is verification rather than solving, at every scale. G-Eval, whose prompt keeps the reference, costs far less for most judges (Qwen3-14B reaches 87.2\%, GPT-OSS-120B 88.6\%). Pairwise comparison fails differently: swapping the answers gives Qwen2.5-72B a mean flip rate of 0.83, and GPT-OSS-120B (0.48) is no more consistent than a coin flip, so size does not make a judge consistent. Thinking mode lowers the flip rate, Qwen3-14B's from 0.78 to 0.45, and the most consistent of all 18 judges is the small Phi-4-Reasoning-Plus (0.39).

\subsection{Majority Voting and Multi-Agent Debate}
\label{sec:ensemble}
\label{sec:debate}

\textsc{Ensembles Gain Little at the Accuracy Frontier.} \textbf{The best of all $\binom{5}{3}{=}10$ three-judge juries reaches 90.27\%, just 0.11 points above the best individual SLM judge.} Voting fixes an error only when the other members get that item right, and the top judges share most of their errors: for every pair of the five, 81--97\% of one judge's errors are errors of the other as well, so all ten juries land in 89.50--90.27\% (Appendix~\ref{app:ensemble}).

\textsc{Multi-Agent Debate Degrades Binary Judging.} \textbf{Under the RCR protocol, no debate beats the individual baseline of its strongest member.} The best debate (Qwen3-4B, same-model, 87.67\%) falls 0.77 points below Qwen3-4B alone (88.44\%), and the best cross-architecture debate at $B{=}8{,}192$ loses 3.32 points against its strongest constituent. A binary verdict carries little structured content to exchange, so the critique step can transmit a confident wrong verdict into a previously correct agent rather than repair it (Appendix~\ref{app:error_analysis}), echoing \citet{wynn2025talk} and \citet{zhang2025stop}; open-ended settings differ \citep{chan2024chateval}.

\subsection{Open-Ended Evaluation}
\label{sec:openended}

\looseness=1 \textsc{Small Judges Match Large Ones on Human-Labeled Summaries.} \textbf{On SummEval's 1,600 human-annotated summaries, the five strongest SLMs match or beat the large judges in 25 of 30 paired comparisons (15 higher, 10 not distinguishable), and all five losses are to Qwen3.5-122B.} The comparisons resample the 100 source articles (Appendix~\ref{app:openended_full}). Phi-4 reaches $\rho{=}0.520$ [0.494, 0.544], significantly above five of the six large judges and 0.022 below Qwen3.5-122B. Across the 16 SLMs, closed-ended accuracy predicts SummEval agreement strongly ($r{=}0.84$) and MT-Bench agreement moderately ($r{=}0.66$; Table~\ref{tab:openended} in Appendix~\ref{app:openended_full}).

\textsc{Closed-Ended and Open-Ended Rankings Diverge.} \textbf{Phi-4 leads the SLMs on binary judging and on SummEval but ranks sixth on MT-Bench ($\rho{=}0.269$), while its reasoning-trained sibling Phi-4-Reasoning ranks second ($\rho{=}0.505$).} The Phi-4 family gives a near-controlled comparison, sharing a base model and differing chiefly in always-thinking behavior, and the reversal holds under paired resampling: Phi-4-Reasoning beats Phi-4 on MT-Bench by 0.235 [0.033, 0.396] and trails it on SummEval by 0.040 [0.020, 0.061]. Verifying an answer against a reference and judging coherence, depth, and style are overlapping but distinct skills, which echoes the domain dependence of \S\ref{sec:individual}.

The same split appears within SummEval: the five strongest SLMs reach $\rho{=}0.53$--$0.62$ on coherence but only $0.36$--$0.42$ on fluency, tracking \textit{logical} structure better than \textit{surface} quality, so fluency scoring still benefits from human cross-validation.

\textsc{MT-Bench Is a Noisy Target.} \textbf{Two large scorers given the identical prompt agree on MT-Bench at only $\rho{=}0.574$ [0.450, 0.685], and with 80 conversations the intervals of the top judges overlap: Qwen3-14B reaches 0.516 [0.369, 0.646], against 0.592 [0.454, 0.710] for Qwen3.5-122B and 0.566 [0.443, 0.670] for GPT-OSS-120B.} Qwen3-14B and Phi-4-Reasoning (0.505 [0.333, 0.657]) score above the other four large judges (0.190--0.328; Appendices~\ref{app:openended_full} and~\ref{app:large_judges}), and SummEval's human labels remain the firmer test.

\subsection[Small vs.\ Large Judges, Family Bias, and Cost]{Small vs.\ Large Judges,\\ Family Bias, and Cost}
\label{sec:scaling}

\begin{table}[t]
\centering
\rowcolors{2}{white}{slmrow}
\begin{adjustbox}{max width=\columnwidth}
\small
\setlength{\tabcolsep}{4pt}
\renewcommand{\arraystretch}{1.1}
\begin{tabular}{lcccc}
\toprule
\hiderowcolors
& & & \multicolumn{2}{c}{\textbf{CE Accuracy (\%)}} \\
\cmidrule(lr){4-5}
\textbf{Judge} & \textbf{Params} & \textbf{Type} & $B{=}10$ & $B{=}8{,}192$ \\
\midrule
\noalign{\global\rownum=2}\showrowcolors
Qwen3.5-122B-A10B & 122B & Large & 90.30 & \textbf{89.82} \\
Phi-4 & 14B & SLM & \underline{90.16} & 87.80 \\
Qwen3-14B & 14B & SLM & 90.15 & 88.60 \\
Qwen2.5-72B & 72B & Large & 89.89 & 88.10 \\
Llama-3.3-70B & 70B & Large & 89.81 & 88.86 \\
Qwen3-8B & 8B & SLM & 89.61 & 88.08 \\
GPT-OSS-120B$^\dagger$ & 120B & Large & --- & 89.47 \\
Phi-4-Reasoning-Plus$^\dagger$ & 14B & SLM & --- & \underline{89.42} \\
Phi-4-Reasoning$^\dagger$ & 14B & SLM & --- & 88.91 \\
GLM-4.5-Air$^\dagger$ & 106B & Large & --- & 88.83 \\
Phi-4-mini & 3.8B & SLM & 88.75 & 88.12 \\
Qwen3-4B & 4B & SLM & 88.21 & 88.44 \\
Llama-3.1-8B & 8B & SLM & 87.29 & 84.20 \\
DeepSeek-R1-Distill-70B$^\dagger$ & 70B & Large & --- & 70.74 \\
\midrule
Qwen3-32B & 32B & Ladder & \textbf{90.38} & 88.14 \\
\bottomrule
\end{tabular}
\end{adjustbox}
\caption{\textbf{Small vs.\ large judges} under the identical pipeline, all eight datasets ($N{=}64{,}824$ per configuration). Bold = column best; underline = best SLM. $^\dagger$ = evaluated at $B{=}8{,}192$ only. Qwen3.5-122B, Phi-4, Qwen3-14B, and Qwen2.5-72B are pairwise equivalent within $\pm$0.5 points (TOST); paired tests and every judge's open-ended agreement are in Appendix~\ref{app:large_judges} (Table~\ref{tab:app_all_judges}). Qwen3-32B (Ladder) is the one Qwen3 size between the SLMs and the large judges.}
\label{tab:large_comparison}
\end{table}

\textsc{Judge Accuracy Saturates Near 8--14B.} \textbf{Phi-4 (90.16\%) is equivalent within $\pm$0.5 points to the best large judge, Qwen3.5-122B (90.30\%), and the other five large judges are significantly less accurate.} All six large judges run through the identical pipeline, prompts, students, and oracle as the small ones (Table~\ref{tab:large_comparison}, Appendix~\ref{app:large_judges}). Qwen2.5-72B and Llama-3.3-70B trail Phi-4 by 0.27 and 0.35 points, both equivalent within $\pm$0.5, while GPT-OSS-120B (0.69 below), GLM-4.5-Air, and DeepSeek-R1-Distill-Llama-70B trail by more than the margin. The 0.14-point lead itself shrinks to 0.03 when the 437 answers the oracle cannot read are set aside: Phi-4 then scores 90.32\% and Qwen3.5-122B 90.35\% (Appendix~\ref{app:oracle_reliability}). Inside one family the curve flattens: Qwen3 gains 9.77 points from 0.6B to 1.7B, then 0.54 from 8B to 14B and 0.23 from 14B to 32B (90.38\%), the last step inside the equivalence margin. The next judge up, Qwen3.5-122B, is a newer generation rather than a larger Qwen3 and activates only about 10B parameters per token, and it is 0.08 below the 32B model. Within Qwen2.5, the 72B judge leads the 7B one by 11.89 points at $B{=}10$ but by 2.67 at each judge's best budget, because Qwen2.5-7B's quick verdict is weak outside math (55.35\% on HellaSwag) and it gains 9.2 points once it can reason. The large judges other than DeepSeek-R1-Distill show the same math-to-general gap (8.5--11.2 points) as the strongest small ones (8.5--12.0), and no judge at any scale clears 80\% on HellaSwag. Nor does scale change how judges fail: five of the six large judges disagree with the oracle on 9.9--11.2\% of items over the budgets each supports and the strongest SLMs on 10.6--11.0\%, and those disagreements fall into the same categories at every size we test (Appendix~\ref{app:error_taxonomy}).

\textsc{Judges Rank Students by Quality.} \textbf{Given room to reason, capable judges agree on how to rank the five students at $\tau{=}0.96$, and the ranking they produce is the students' own solver accuracy ($\tau{=}0.98$).} We widen the pool to five students from five families spanning 29.4--84.4\% solver accuracy and compute each judge's accuracy and false-accept rate on every student over all eight datasets (Appendix~\ref{app:self_preference}). To separate lineage from student quality, we add a strong Qwen3-family student (59.3\% solver accuracy, against 29.4\% for the existing one) and difference out each judge's general leniency, asking how much more a family accepts wrong answers from its own lineage student than other families do. At $B{=}10$ this lineage effect on the false-accept rate is $+0.010$ for Qwen3, $-0.075$ for Llama, and $+0.010$ for Qwen2.5, so no family favors its own lineage at the quick verdict. A preference appears only once judges reason, mainly for Qwen3 ($+0.21$; Llama $+0.09$, Qwen2.5 $+0.04$), one more reason to give verifiable tasks a quick verdict. Favoritism could not inflate the main results regardless, since accepting a wrong answer costs accuracy against the oracle.

\textsc{A Quick Verdict Costs an Estimated 9 to 32 Times More at 70B--122B.} \textbf{Phi-4 serves a thousand verdicts for \$0.34; Qwen2.5-72B needs \$3.23, Llama-3.3-70B \$6.28, and Qwen3.5-122B \$10.83 for the same work, at accuracy within 0.35 points of Phi-4 either way.} Cost follows each judge's GPU count and an approximate decode throughput that falls with total parameter count (Appendix~\ref{app:large_judges}): two GPUs at 14B, four to eight for the large judges. Two multipliers compound. Going small cuts the hardware bill by an order of magnitude; going quick cuts the output term $\beta B$ of $C_\text{total}$ by three orders of magnitude ($\tfrac{10}{8192}$), and with a 200-token prompt counted a quick verdict still costs a fortieth of a reasoned one. Because $\Delta_{\text{math}} > 0$ for most judges, that second saving arrives \textit{with} a small accuracy gain on verifiable tasks. The large judges evaluated only with reasoning pay the most: GPT-OSS-120B lands near \$426 per thousand judgments, about 1{,}240 times Phi-4's cost, for 0.69 points less accuracy. Even the mid-size Qwen3-32B adds only 0.22 points over Phi-4, inside the $\pm$0.5-point margin, at 2.15 times its cost.

\section{Conclusion and Future Work}
\label{sec:conclusion}

\slmjury{} casts SLM judging as a budget-conditioned function and evaluates it across both paradigms over 16 small judges, six large ones, 10 benchmarks across three closed-ended domains and two open-ended tasks, and 12{,}029 model runs. We find that (1) on reference-guided verification the best 14B judge, Phi-4, is equivalent to the strongest 70B--122B judge within $\pm$0.5 points and significantly more accurate than the other five, accuracy saturates near 8--14B, and a small judge costs an estimated 9 to 32 times less to serve than a large one giving the same quick verdict; (2) overthinking is \textit{domain-dependent} and prompt-mediated, since budget alone leaves 11 of 13 judges' accuracy unchanged and truncation collapses instruction following; (3) math-to-general gaps separate families and persist for the strongest judges at every scale we test; (4) on human-labeled summaries the top SLMs match or beat large judges in 25 of 30 comparisons, while the two paradigms rely on different capabilities and reorder the judges; (5) debate degrades accuracy everywhere we tested it, while the two strongest judges hold aggregate accuracy within a point across six personas and judge rankings stay stable across prompt formulations; and (6) judges rank students by the students' own solver accuracy, with no lineage preference at the quick verdict. We recommend a \textit{budget-matched} judge over a bigger one: a quick-verdict SLM for verifiable tasks with a reference, and an SLM validated on human labels for open-ended scoring. Adaptive budgets pay off when they retry unreadable verdicts rather than when they allocate by domain (Appendix~\ref{app:adaptive_budget}); in a human audit of sampled judge--oracle disagreements, most cases side with the judge, for small and large judges alike (Appendix~\ref{app:oracle_audit}). Judge-specific fine-tuning and the selection of decorrelated panels remain open.

\section*{Limitations}\label{sec:limitations}

\begin{enumerate}[leftmargin=*,itemsep=2pt,topsep=2pt,parsep=0pt]
\item \textbf{Verification against a programmatic oracle.} Our closed-ended conclusions apply to reference-guided verification and do not support replacing human review in high-stakes settings. The oracle is imperfect: it scores the 437 unreadable student answers \texttt{Incorrect} (Appendix~\ref{app:oracle_reliability}), and an audit sides with the judge in most of the sampled judge--oracle disagreements (Appendix~\ref{app:error_taxonomy}). Because one oracle scores every judge, comparisons are like for like.
\item \textbf{Open-ended references.} MT-Bench is scored against a model reference, two such references agree only moderately ($\rho{=}0.57$), and its 80 conversations give wide intervals, so we treat it as secondary evidence.
\item \textbf{Coverage of judges and settings.} The six large judges are all open-weight; we leave out proprietary API judges, whose versions change without notice. Personas (in English), ensembles, and debate cover the small judges only, and the protocol comparison covers the 16 SLMs and two large judges.
\item \textbf{Students.} Main results average two students (95.45\% and 82.49\% on GSM8K), and with at most two same-family students per family in any one panel and none for Phi-4, family and student quality are only partly separable.
\item \textbf{Run-to-run variation.} Batched inference is not bit-reproducible, even with greedy decoding. Re-running a configuration changes 0.01\% of Phi-4's and Qwen3-14B's quick verdicts but 9.45\% of the weakest judge's (Appendix~\ref{app:run_variance}).
\end{enumerate}

\paragraph{Potential Risks.} Our work poses no direct risks, but SLM judges used in critical applications (e.g., education, hiring) should be tested for robustness before deployment. Adversarial system prompts can lower a weaker judge's accuracy by over 10 points (Llama-3.1-8B under the Lenient persona; Appendix~\ref{app:personas}). The framework is publicly available at \url{https://github.com/anishh15/SLMJury} for transparent benchmarking, and we encourage further research on judge calibration.

\section*{Ethical Considerations}

This study evaluates small language models using standardized benchmarks and publicly available datasets, ensuring transparency and reproducibility. No private or sensitive data was used, and all models were assessed under fair and consistent conditions. We acknowledge potential biases in both programmatic and LLM-based evaluations and encourage further research for mitigation. All model and dataset licenses are documented in Appendix~\ref{app:reproducibility}.

\section*{Acknowledgements}

The core experiments were conducted on the NVIDIA DGX server at the LNM Institute of Information Technology (LNMIIT), Jaipur; the large judges and most extended analyses ran on separate cluster nodes. We thank the institute for providing computational resources.

\paragraph{AI Assistance.} We used AI assistants for parts of the implementation and manuscript preparation, including generating LaTeX code for tables and refining text written by the authors. All AI-generated content was carefully reviewed and revised by the authors to ensure accuracy and clarity.

\bibliography{references}

\setlength{\dbltextfloatsep}{20pt plus 2pt minus 4pt}
\captionsetup{skip=10pt}

\clearpage
\appendix

\setcounter{dbltopnumber}{4}
\renewcommand{\dbltopfraction}{0.95}
\renewcommand{\dblfloatpagefraction}{0.80}
\renewcommand{\floatpagefraction}{0.80}
\setcounter{topnumber}{4}
\setcounter{bottomnumber}{2}
\renewcommand{\bottomfraction}{0.6}

\section*{\centering\LARGE Appendix}
\addcontentsline{toc}{section}{Appendix}

\vspace{0.5em}

\noindent This appendix contains the dataset, model, and implementation details behind the experimental setup; the formal framework, extraction, normalization, and scoring machinery; the full prompt and persona texts; and the complete result tables for every analysis the main text summarizes, including the per-dataset breakdowns, the budget sweep and single-factor decomposition, prompt and protocol robustness, the large-judge comparison with its paired significance tests, the family-bias matrices, the adaptive budget policy, the error taxonomy, and a manual audit of the oracle.

\vspace{1em}

\noindent\textbf{\large Contents of the Appendix}

\vspace{0.5em}

\startcontents[appendix]
{\raggedright\hyphenpenalty=10000\exhyphenpenalty=10000
\printcontents[appendix]{l}{1}{\setcounter{tocdepth}{2}}\par}

\section{Dataset Details}
\label{app:datasets}

We evaluate \slmjury{} on eight closed-ended reasoning benchmarks spanning three domains (mathematical, scientific, and general reasoning), plus two open-ended quality scoring benchmarks.
Each closed-ended dataset is loaded from HuggingFace via the \texttt{datasets} library and cached locally as JSON for reproducibility.
Closed-ended datasets use the \textbf{test} split unless noted otherwise.
Table~\ref{tab:app_datasets} summarizes the datasets.

\begin{table}[!htbp]
\centering
\small
\begin{adjustbox}{max width=\columnwidth}
\rowcolors{2}{white}{slmrow}
\setlength{\tabcolsep}{4pt}
\begin{tabular}{llrl}
\toprule
\textbf{Dataset} & \textbf{Domain} & \textbf{Size} & \textbf{Answer format} \\
\midrule
GSM8K & Math & \phantom{0}1,319 & After \texttt{\#\#\#\#} \\
GSM-Plus & Math & 10,552 & Direct numeric \\
MATH & Math & \phantom{0}5,000 & \texttt{\textbackslash boxed\{\}} \\
ARC-Easy & Science & \phantom{0}2,376 & A / B / C / D \\
ARC-Chall. & Science & \phantom{0}1,172 & A / B / C / D \\
HellaSwag & General & 10,042 & A / B / C / D \\
WinoGrande & General & \phantom{0}1,267 & A / B \\
TruthfulQA & General & \phantom{00,}684 & A / B / C / D \\
\midrule
\multicolumn{2}{l}{\textbf{Total (closed-ended)}} & \textbf{32,412} & \\
\midrule
SummEval & Summarization & 1,600 pairs & 1--5 per dimension \\
MT-Bench & Conversation & 80 questions & 1--5 overall \\
\bottomrule
\end{tabular}
\end{adjustbox}
\caption{\textbf{Benchmark datasets.} Closed-ended evaluations use the test split, except HellaSwag, WinoGrande, and TruthfulQA (validation). With 2 student models, each judge configuration is evaluated over $32{,}412 \times 2 = 64{,}824$ total closed-ended judgments.}
\label{tab:app_datasets}
\end{table}

\paragraph{GSM8K} \citep{cobbe2021training} consists of 1,319 grade-school math word problems requiring multi-step arithmetic reasoning. Ground truth answers follow the \texttt{\#\#\#\#} delimiter. Our loader extracts the final numeric answer after this delimiter.
Source: \texttt{openai/gsm8k}.

\paragraph{GSM-Plus} \citep{li2024gsm} extends GSM8K with 10,552 adversarially perturbed variants covering numerical substitution, digit expansion, fraction conversion, and other perturbation types. Ground truth is provided as a direct numeric value in the dataset schema.
Source: \texttt{qintongli/GSM-Plus}.

\paragraph{MATH} \citep{hendrycks2021measuring} contains 5,000 competition-level mathematics problems across 7 subjects: algebra, counting \& probability, geometry, intermediate algebra, number theory, prealgebra, and precalculus. Answers are LaTeX expressions inside \texttt{\textbackslash boxed\{\}}. Our loader extracts the content of the last \texttt{\textbackslash boxed\{\}} occurrence using a brace-matching parser that handles nested braces.
Source: \texttt{EleutherAI/hendrycks\_math}.

\paragraph{ARC-Easy and ARC-Challenge} \citep{clark2018think} comprise 2,376 and 1,172 multiple-choice science questions, respectively. Each question has 3--5 answer choices, labeled A--D, or A--E for the four ARC-Easy and three ARC-Challenge questions with a fifth choice; 97 ARC-Easy and 22 ARC-Challenge questions label their choices 1--4 instead, which the oracle maps to A--D. Our loader formats the question with inline choices: \texttt{A) text  B) text  C) text  D) text}, and stores the correct label as ground truth. The extractor reads only A--D, so the one question whose answer is E is scored \texttt{Incorrect} for both students.
Source: \texttt{allenai/ai2\_arc}.

\paragraph{HellaSwag} \citep{zellers2019hellaswag} contains 10,042 sentence completion questions requiring commonsense reasoning about everyday activities (\textbf{validation} split; no public test labels). Each question presents a context and four possible continuations (A, B, C, D). The correct continuation requires understanding physical and social commonsense, making it substantially harder for judges than factual recall. Our loader formats the context with inline choices.
Source: \texttt{Rowan/hellaswag}.

\paragraph{WinoGrande} \citep{sakaguchi2021winogrande} provides 1,267 pronoun resolution problems derived from Winograd schemas (\textbf{validation} split; no public test labels). Each problem requires choosing between two possible referents (A or B) for an ambiguous pronoun. This benchmark tests commonsense reasoning about physical interactions and social dynamics. Our loader maps numeric labels (1, 2) to letters (A, B).
Source: \texttt{allenai/winogrande} (\texttt{winogrande\_xl} config).

\paragraph{TruthfulQA} \citep{lin2022truthfulqa} consists of 684 multiple-choice questions designed to test truthfulness on topics where popular misconceptions exist (\textbf{validation} split; no public test labels). Questions span 38 categories (health, law, conspiracies, etc.) with 4 answer choices per question. Our loader uses the \texttt{multiple\_choice} config, reading the \texttt{choices} list and the \texttt{label} field (a ClassLabel index 0--3 mapped to letters A--D).
Source: \texttt{EleutherAI/truthful\_qa\_mc}.

\paragraph{SummEval (SLM-Human Correlation)} \citep{fabbri2021summeval} provides 1,600 machine-generated article-summary pairs annotated by human experts on four dimensions: coherence, consistency, fluency, and relevance (each scored 1--5 and averaged over the three expert annotators). We use these human annotations as ground truth to measure SLM-Human agreement: correlations between SLM judge scores and human expert scores reveal how well small models can replicate nuanced human quality judgments on generated text.
Source: \texttt{mteb/summeval}.

\paragraph{MT-Bench (SLM-LLM Correlation)} \citep{zheng2023judging} contains 80 multi-turn conversational questions across eight categories (coding, extraction, humanities, math, reasoning, roleplay, STEM, writing). The reference scorer is Qwen3.5-397B-A17B (\texttt{Qwen/Qwen3.5-397B-A17B}), queried via the Together API with the judges' own scoring prompt on the responses of both student models, which gives two reference-student combinations per judge. The students answer each question at $\tau{=}0.7$ and top-$p{=}0.9$ with up to 2{,}048 tokens, and the reference scorers run at temperature 0 with up to 2{,}048 tokens, Qwen3.5-397B-A17B with its reasoning mode off. Qwen3.5-397B-A17B is not itself a judge in the study. A second large model, GPT-OSS-120B (\texttt{openai/gpt-oss-120b}), scored the same responses with the same prompt; because it is also one of the large judges, its scores serve only to measure how reliable the reference is (Appendix~\ref{app:large_judges}). Correlations between judge scores and the reference measure SLM-LLM agreement: how well small models can approximate the scoring behavior of a much larger model on multi-turn interactions.
Source: \texttt{philschmid/mt-bench}.

\paragraph{Data Schema.} All closed-ended datasets are normalized to a unified schema with four fields: \texttt{problem\_id} (integer), \texttt{question} (string), \texttt{ground\_truth\_reasoning} (string, empty for non-math datasets), and \texttt{ground\_truth\_answer} (string). This standardization allows all downstream components (judge inference, evaluation, ensemble voting) to operate uniformly across datasets.

\section{Model Configurations}
\label{app:models}

Table~\ref{tab:app_judge_models} gives the configuration of the 16 small judges, Table~\ref{tab:app_large_models} the six large judges and the ladder judge Qwen3-32B, and Table~\ref{tab:app_student_models} the two students behind every main-text result except the family-bias analysis. Two analyses widen that pool: the family-bias panel of Appendix~\ref{app:self_preference} uses five students across all eight datasets, plus a stronger Qwen3-family student for the lineage test, and a wider grid there uses seven across the five datasets they have in common. Table~\ref{tab:app_solver} names all eight and gives their solver accuracy on both grids.
Small-judge and student configurations live in one versioned file, and the large judges, the ladder judge, and the added students have their own; all are parsed at runtime.

\begin{table*}[t]
\centering
\small
\begin{adjustbox}{max width=\textwidth}
\rowcolors{2}{white}{slmrow}
\setlength{\tabcolsep}{4pt}
\begin{tabular}{lllccccc}
\toprule
\textbf{Family} & \textbf{Key} & \textbf{HuggingFace Identifier} & \textbf{Params} & \textbf{TP} & \textbf{Max Len} & \textbf{Thinking} & \textbf{Always} \\
\midrule
& \texttt{llama3.2-1b} & \texttt{meta-llama/Llama-3.2-1B-Instruct} & 1B & 1 & 32,768 & \no & \no \\
& \texttt{llama3.2-3b} & \texttt{meta-llama/Llama-3.2-3B-Instruct} & 3B & 1 & 40,960 & \no & \no \\
\multirow{-3}{*}{Llama 3.x}
& \texttt{llama3.1-8b} & \texttt{meta-llama/Llama-3.1-8B-Instruct} & 8B & 2 & 32,768 & \no & \no \\
\midrule
& \texttt{qwen2.5-1.5b} & \texttt{Qwen/Qwen2.5-1.5B-Instruct} & 1.5B & 1 & 32,768 & \no & \no \\
& \texttt{qwen2.5-3b} & \texttt{Qwen/Qwen2.5-3B-Instruct} & 3B & 1 & 32,768 & \no & \no \\
\multirow{-3}{*}{Qwen2.5}
& \texttt{qwen2.5-7b} & \texttt{Qwen/Qwen2.5-7B-Instruct} & 7B & 2 & --- & \no & \no \\
\midrule
& \texttt{qwen3-0.6b} & \texttt{Qwen/Qwen3-0.6B} & 0.6B & 1 & 40,960 & \yes & \no \\
& \texttt{qwen3-1.7b} & \texttt{Qwen/Qwen3-1.7B} & 1.7B & 1 & 40,960 & \yes & \no \\
& \texttt{qwen3-4b} & \texttt{Qwen/Qwen3-4B} & 4B & 1 & 40,960 & \yes & \no \\
& \texttt{qwen3-8b} & \texttt{Qwen/Qwen3-8B} & 8B & 2 & 32,768 & \yes & \no \\
\multirow{-5}{*}{Qwen3}
& \texttt{qwen3-14b} & \texttt{Qwen/Qwen3-14B} & 14B & 2 & 32,768 & \yes & \no \\
\midrule
& \texttt{phi4-14b} & \texttt{microsoft/phi-4} & 14B & 2 & 16,384 & \no & \no \\
& \texttt{phi4r-14b} & \texttt{microsoft/Phi-4-reasoning} & 14B & 2 & 32,768 & \yes & \yes \\
& \texttt{phi4rp-14b} & \texttt{microsoft/Phi-4-reasoning-plus} & 14B & 2 & 32,768 & \yes & \yes \\
& \texttt{phi4mi-3.8b} & \texttt{microsoft/Phi-4-mini-instruct} & 3.8B & 1 & 40,960 & \no & \no \\
\multirow{-5}{*}{Phi-4}
& \texttt{phi4mr-3.8b} & \texttt{microsoft/Phi-4-mini-reasoning} & 3.8B & 1 & 32,768 & \yes & \yes \\
\bottomrule
\end{tabular}
\end{adjustbox}
\caption{\textbf{Judge model configurations.} TP = tensor parallel size (number of GPUs). Max Len = maximum context length passed to vLLM. Thinking = whether the model's chat template includes a \texttt{<think>} block. Always = models that always generate thinking tokens (cannot be disabled); these are evaluated at 8,192 tokens only (marked $^\dagger$ in Table~\ref{tab:individual}).}
\label{tab:app_judge_models}
\end{table*}

\begin{table*}[ht]
\centering
\small
\begin{adjustbox}{max width=\textwidth}
\rowcolors{2}{white}{slmrow}
\begin{tabular}{llcc}
\toprule
\textbf{Key} & \textbf{HuggingFace Identifier} & \textbf{Params} & \textbf{TP} \\
\midrule
\texttt{qwen2.5-32b} & \texttt{Qwen/Qwen2.5-32B-Instruct-GPTQ-Int8} & 32B & 2 \\
\texttt{llama3.1-8b} & \texttt{meta-llama/Llama-3.1-8B-Instruct} & 8B & 2 \\
\bottomrule
\end{tabular}
\end{adjustbox}

\caption{\textbf{Student model configurations.} The 32B Qwen model uses GPTQ INT8 quantization. Each student model generates solutions for all 8 closed-ended datasets, producing the student answer corpus that judges evaluate.}
\label{tab:app_student_models}
\end{table*}

\paragraph{Thinking Mode Handling.} Qwen3 models support a ``thinking'' mode activated via the chat template parameter \texttt{enable\_thinking=True}. When a thinking-capable model receives \texttt{max\_tokens=10}, we pass \texttt{enable\_thinking=False} to suppress internal reasoning and force a direct verdict. At \texttt{max\_tokens=8192}, thinking is enabled, allowing the model to use internal chain-of-thought before producing its verdict. Three small models (Phi-4-Reasoning, Phi-4-Reasoning-Plus, Phi-4-mini-Reasoning) always generate thinking tokens regardless of the template flag. These always-thinking models are evaluated at 8,192 tokens only and marked with $^\dagger$ in Table~\ref{tab:individual}. Among the larger judges, DeepSeek-R1-Distill-Llama-70B and GPT-OSS-120B always reason and are likewise evaluated at 8,192 tokens only; GLM-4.5-Air has a thinking switch but was evaluated at 8,192 tokens only; Qwen3.5-122B-A10B and Qwen3-32B use the same switch as the Qwen3 judges and run at both budgets. Table~\ref{tab:app_large_models} gives the serving configuration of each of these judges, including the tensor-parallel width that sets its GPU count in the cost estimate.

\begin{table*}[t]
\centering
\small
\rowcolors{2}{white}{slmrow}
\begin{adjustbox}{max width=\textwidth}
\setlength{\tabcolsep}{4pt}
\begin{tabular}{llcccccc}
\toprule
\textbf{Key} & \textbf{HuggingFace Identifier} & \textbf{Params} & \textbf{TP} & \textbf{Weights} & \textbf{Thinking} & \textbf{Always} & \textbf{Budgets} \\
\midrule
\texttt{llama3.3-70b} & \texttt{meta-llama/Llama-3.3-70B-Instruct} & 70B & 8 & FP16 & \no & \no & both \\
\texttt{\makecell[l]{deepseek-r1-distill-\\llama-70b}} & \texttt{\makecell[l]{deepseek-ai/\\DeepSeek-R1-Distill-Llama-70B}} & 70B & 8 & BF16 & \yes & \yes & 8{,}192 \\
\texttt{qwen2.5-72b} & \texttt{Qwen/Qwen2.5-72B-Instruct} & 72B & 4 & BF16 & \no & \no & both \\
\texttt{glm-4.5-air} & \texttt{zai-org/GLM-4.5-Air-FP8} & \makecell{106B\\(12B act.)} & 8 & FP8 & \yes & \no & 8{,}192 \\
\texttt{gpt-oss-120b} & \texttt{openai/gpt-oss-120b} & 120B & 8 & MXFP4 & \yes & \yes & 8{,}192 \\
\texttt{qwen3.5-122b} & \texttt{Qwen/Qwen3.5-122B-A10B-FP8} & \makecell{122B\\(10B act.)} & 8 & FP8 & \yes & \no & both \\
\midrule
\texttt{qwen3-32b} (ladder) & \texttt{Qwen/Qwen3-32B} & 32B & 2 & BF16 & \yes & \no & both \\
\bottomrule
\end{tabular}
\end{adjustbox}
\caption{\textbf{Large and ladder judge configurations.} These judges run through the identical pipeline, prompts, students, and oracle as the small judges, all with a maximum context of 32,768 tokens. Thinking = the model has a reasoning mode; Always = the mode cannot be switched off; Budgets = the token budgets evaluated (both = 10 and 8,192). TP is the tensor-parallel width of each judge's 16-bit serving configuration on 44\,GiB GPUs, which the serving-cost estimate of Appendix~\ref{app:large_judges} uses as its GPU count. GLM-4.5-Air and Qwen3.5-122B-A10B were served from the vendors' FP8 checkpoints and GPT-OSS-120B from its MXFP4 release, which fit on fewer GPUs. Judges with a thinking channel decode at $\tau{=}0.6$, top-$p{=}0.95$, top-$k{=}20$ at $B{=}8{,}192$, except GPT-OSS-120B, which was decoded greedily. Qwen3-32B is the one Qwen3 size between the small judges and the large ones and is reported separately as the ladder judge.}
\label{tab:app_large_models}
\end{table*}

\section{Formal Framework}
\label{app:framework}

This section gives the complete formal description of the \slmjury{} pipeline
that \S\ref{sec:method} summarizes. We define the judging problem, the
budget-conditioned judge function, the programmatic oracle, and the closed-ended
metrics in one place; the open-ended correlation metrics are defined in
Appendix~\ref{app:scoring_metrics}.

\subsection{Spaces and Objects}

Let $\mathcal{Q}$ be the space of questions, $\mathcal{G}$ the space of
ground-truth references, $\mathcal{R}$ the space of student responses, and
\begin{equation}
\mathcal{V} = \{\texttt{Correct},\ \texttt{Incorrect},\ \texttt{Undefined}\}
\end{equation}
the verdict space, where \texttt{Undefined} denotes an unparseable response. A
\emph{benchmark} is a finite set
$\mathcal{D} = \{(q_i, a_i^{*}, r_{s,i})\}_{i=1}^{N_{\mathcal{D}}}$ of questions
$q_i$, gold answers $a_i^{*}$, and student responses $r_{s,i}$ produced by a
student model $\mathcal{S}$ (Appendix~\ref{app:student}). When step-by-step gold
reasoning $r_i^{*}$ is available (the math datasets), the reference passed to the
judge is $g_i = r_i^{*}$; otherwise $g_i = a_i^{*}$.

\subsection{The Judge Function}

A judge is a language model $\mathcal{J}$ with parameters $\theta_{\mathcal{J}}$
constrained to $|\theta_{\mathcal{J}}| \le 14\text{B}$; the comparison judges of
\S\ref{sec:scaling} (Qwen3-32B and the six large judges at 70B--122B) realize the
same function through the same code. Under token budget
$B \in \{10, 8192\}$ it realizes a (stochastic) map
\begin{equation}
f_{\mathcal{J}}^{(B)} : \mathcal{Q} \times \mathcal{G} \times \mathcal{R}
\;\longrightarrow\; \mathcal{V},
\qquad
(q_i, g_i, r_{s,i}) \mapsto \hat{v}_i .
\end{equation}
Concretely, $f_{\mathcal{J}}^{(B)}$ applies the budget-appropriate prompt
template $\mathcal{T}_B$ (Appendix~\ref{app:prompt_quick},
\ref{app:prompt_reasoned}), tokenizes it through the model's native chat
template, decodes a response $y_i \sim \pi_{\mathcal{J}}(\cdot \mid
\mathcal{T}_B(q_i, g_i, r_{s,i}))$ of at most $B$ tokens, and parses a verdict
$\hat{v}_i = \textsc{Parse}(y_i)$ with the priority cascade of
Appendix~\ref{app:verdict_cascade}. Decoding is deterministic ($\tau{=}0$) for the
quick verdict and for non-thinking models at $B{=}8{,}192$, and uses
Qwen3's recommended sampling configuration ($\tau{=}0.6$, top-$p{=}0.95$, top-$k{=}20$,
min-$p{=}0$) when native thinking is enabled. The one exception is GPT-OSS-120B,
which reasons in its own output channels and was decoded greedily at
$B{=}8{,}192$.

\subsection{The Programmatic Oracle}

Ground-truth verdicts come from a deterministic oracle rather than from human
labels, which makes the closed-ended evaluation exactly reproducible. Define a
dataset-type-specific equivalence relation
\begin{equation}
\textsc{Equiv}_{\,\text{type}}(\hat{a}, a^{*}, \delta) \in \{\top, \bot\},
\end{equation}
with tolerance $\delta = 10^{-9}$, where $\text{type} \in
\{\textsc{numeric}, \textsc{latex}, \textsc{mc}\}$ is selected from the dataset
(Appendix~\ref{app:normalization}). The oracle verdict for instance $i$ is
\begin{equation}
v_i^{*} =
\begin{cases}
\texttt{Correct}   & \!\text{if } \textsc{Equiv}_{\text{type}}(\hat{a}_i, a_i^{*}, \delta) = \top,\\
\texttt{Incorrect} & \!\text{otherwise,}
\end{cases}
\end{equation}
where $\hat{a}_i = \textsc{Extract}_{\text{type}}(r_{s,i})$ is the student's
final answer recovered by the answer extractor of
Appendix~\ref{app:answer_extraction}. When every extraction strategy fails,
$\hat{a}_i$ is a sentinel that matches no gold answer, so $v_i^{*} =
\texttt{Incorrect}$; this happens for 437 of the 64{,}824 student answers
(0.67\%), and because extraction depends only on the student response, they are
the same items for every judge (Appendix~\ref{app:oracle_reliability}). The judge never sees $v_i^{*}$; it must
re-derive correctness from $(q_i, g_i, r_{s,i})$ alone.

\subsection{Closed-Ended Metrics}

\paragraph{Accuracy.} With $\hat{v}_i = f_{\mathcal{J}}^{(B)}(q_i, g_i, r_{s,i})$ the
judge's verdict on item $i$, agreement between the judge and the oracle over a
configuration is
\begin{equation}
\text{Acc}(\mathcal{J}, B) = \frac{1}{N} \sum_{i=1}^{N}
\mathbf{1}\!\left[\hat{v}_i = v_i^{*}\right],
\end{equation}
computed over $N = 64{,}824$ judgments (2 students $\times$ 8 datasets) per
configuration.

\paragraph{Instruction-Following Rate.} The fraction of responses that yield a
parseable (non-\texttt{Undefined}) verdict is
\begin{equation}
\text{IFR}(\mathcal{J}, B) = \frac{1}{N} \sum_{i=1}^{N}
\mathbf{1}\!\left[\hat{v}_i \neq \texttt{Undefined}\right].
\end{equation}
Because $\text{IFR} > 99.8\%$ for all top judges (Appendix~\ref{app:per_dataset}),
$\text{Acc}$ and $\text{Acc}/\text{IFR}$ (accuracy among parsed responses)
differ by at most 0.2 points, so we report raw accuracy throughout.

\paragraph{Budget effect.} For domain $\mathcal{D}$ we summarize the
overthinking effect by the signed difference
\begin{equation}
\Delta(\mathcal{J}, \mathcal{D}) = \text{Acc}(\mathcal{J}, 10;\mathcal{D})
- \text{Acc}(\mathcal{J}, 8192;\mathcal{D}),
\end{equation}
where $\Delta > 0$ means quick verdicts win. \S\ref{sec:individual} shows that
$\operatorname{sign}\Delta$ can flip with $\mathcal{D}$ (verifiable vs.
non-verifiable) for the same $\mathcal{J}$.

\paragraph{Persona robustness.} For a judge evaluated under a set $P$ of
persona system prompts (Appendix~\ref{app:personas}), robustness is the range
\begin{equation}
r(\mathcal{J}) = \max_{p \in P} \text{Acc}(\mathcal{J}, p)
- \min_{p \in P} \text{Acc}(\mathcal{J}, p),
\end{equation}
with smaller $r$ indicating a more stable internal correctness boundary.

\subsection{Comparing Two Judges}
\label{app:paired_stats}

Every judge scores the same $N = 64{,}824$ items, so two judges $\mathcal{J}_a$
and $\mathcal{J}_b$ are compared on paired per-item correctness
$c_{a,i}, c_{b,i} \in \{0, 1\}$, each at its best budget. Their accuracy
difference is $\hat{d} = \text{Acc}_a - \text{Acc}_b$.

\paragraph{McNemar's test.} Only the discordant items carry information about
which judge is better. With $n_{10}$ items that only $\mathcal{J}_a$ gets right
and $n_{01}$ that only $\mathcal{J}_b$ gets right, the null hypothesis
$\hat{d} = 0$ makes $n_{10} \sim \text{Binomial}(n_{10} + n_{01}, \tfrac12)$,
and we report the exact two-sided $p$-value \citep{mcnemar1947note}.

\paragraph{Paired bootstrap.} The 95\% interval on $\hat{d}$ resamples item
differences $c_{a,i} - c_{b,i}$ with replacement \emph{within} each of the eight
datasets, holding dataset sizes fixed, over 10{,}000 resamples with a fixed seed.
Items within a dataset are exchangeable; the dataset sizes are a design choice
rather than a sample, so whole datasets are not resampled.

\paragraph{Equivalence.} A non-significant McNemar test says only that two
judges cannot be told apart. To call them \emph{equivalent} we use two one-sided
tests \citep[TOST;][]{schuirmann1987comparison} against a margin of $\pm$0.5
points fixed in advance: the judges are equivalent at $\alpha = 0.05$ when the
90\% paired-bootstrap interval on $\hat{d}$ lies entirely inside
$(-0.5, +0.5)$.

\paragraph{Multiple comparisons.} Over all $\binom{22}{2} = 231$ judge pairs,
McNemar $p$-values are corrected with the Benjamini--Hochberg procedure
\citep{benjamini1995controlling} at a false discovery rate of 0.05. Results are
in Appendix~\ref{app:large_judges}.

\paragraph{Open-ended statistics.} Intervals on SummEval $\rho$ resample the
100 source articles together with their 16 summaries each; intervals on
MT-Bench $\rho$ resample the 80 questions, using the same draw for both
students. Every judge is resampled with the same draws, so the difference
between two judges has a paired interval (10{,}000 resamples). The agreement
between the two MT-Bench reference scorers is bootstrapped separately, with
2{,}000 resamples that draw each student's questions independently.

\section{Prompt Templates}
\label{app:prompts}

This section presents the exact prompt templates used across all experiments.
All prompts are reproduced verbatim from the codebase; no modifications were made between experimental runs.

\subsection[Quick Verdict Prompt]{Judge Prompt: Quick Verdict (10 Tokens)}
\label{app:prompt_quick}

This prompt is used when \texttt{max\_tokens=10}. The judge must respond with exactly one word: \texttt{Correct} or~\texttt{Incorrect}.

\begin{promptflow}{Quick Verdict Prompt (\texttt{max\_tokens=10})}
\begin{verbatim}
Your role is to compare the student's answer
to the ground truth and determine correctness.

Respond with exactly one word:
- 'Correct': If the student's final answer
  matches the ground truth.
- 'Incorrect': If the student's final answer
  does not match.

Focus only on the final answer, ignore
reasoning steps.

Question: {question}

Ground truth answer: {ground_truth}

Student answer: {student_answer}
\end{verbatim}
\end{promptflow}

\subsection[Reasoned Verdict Prompt]{Judge Prompt: Reasoned Verdict (8,192 Tokens)}
\label{app:prompt_reasoned}

This prompt is used when \texttt{max\_tokens=8192}. The judge provides step-by-step reasoning followed by a boxed verdict.

\begin{promptflow}{Reasoned Verdict Prompt (\texttt{max\_tokens=8192})}
\begin{verbatim}
Your role is to compare the student's answer
to the ground truth and determine correctness.

First, explain your reasoning for the judgment.
Then, on a new line, give the final verdict in
the exact format:
\boxed{CORRECT} or \boxed{INCORRECT}

Focus only on the final answer when judging
correctness. Ignore reasoning steps in the
student's solution.

Question: {question}

Ground truth answer: {ground_truth}

Student answer: {student_answer}
\end{verbatim}
\end{promptflow}

\subsection{Student Solver Prompts}
\label{app:student_prompts}

The student models (Qwen2.5-32B and Llama-3.1-8B) generate solutions using dataset-type-specific prompts:

\begin{promptflow}{Numeric Dataset Prompt (GSM8K, GSM-Plus)}
\begin{verbatim}
Think step-by-step to solve this math problem.
Give your final answer after ####.

Problem: {question}

Solution:
\end{verbatim}
\end{promptflow}

\begin{promptflow}{LaTeX Dataset Prompt (MATH)}
\begin{verbatim}
Think step-by-step to solve this math problem.
Put your final answer in \boxed{}.

Problem: {question}

Solution:
\end{verbatim}
\end{promptflow}

\begin{promptflow}{Science Multiple-Choice Prompt (ARC-Easy, ARC-Challenge)}
\begin{verbatim}
Think step-by-step to answer this science
question. Choose A, B, C, or D.

{question}

Think step-by-step, then give your final answer.
\end{verbatim}
\end{promptflow}

\begin{promptflow}{General Multiple-Choice Prompt (HellaSwag, WinoGrande, TruthfulQA)}
\begin{verbatim}
Think step-by-step to answer this question.
Choose the best option.

{question}

Think step-by-step, then give your final answer.
\end{verbatim}
\end{promptflow}

\subsection{Open-Ended Scoring Prompts}
\label{app:scoring_prompts}

The following prompts are used for open-ended quality scoring evaluation (\S\ref{sec:openended}).

\begin{promptflow}{SummEval Scoring Prompt (4 dimensions, single call)}
\begin{verbatim}
Evaluate the following summary of a news article
on four dimensions.
Rate EACH dimension from 1 (very poor) to 5
(excellent).

Source Article:
{source_text}

Summary to Evaluate:
{summary}

Dimensions:
- Coherence: Is the summary well-organized and
  easy to follow?
- Consistency: Does the summary accurately
  reflect the source article without
  contradictions?
- Fluency: Is the summary grammatically correct
  and well-written?
- Relevance: Does the summary capture the
  important information from the source?

Provide your ratings as:
\boxed{coherence=X, consistency=X, fluency=X,
relevance=X}
\end{verbatim}
\end{promptflow}

\begin{promptflow}{MT-Bench Scoring Prompt (2-turn holistic)}
\begin{verbatim}
Rate the overall quality of the assistant's
responses in the following 2-turn conversation
on a scale of 1-5.
1 = Very poor, 2 = Poor, 3 = Average,
4 = Good, 5 = Excellent.

[Turn 1 Question]:
{turn1_question}

[Turn 1 Response]:
{turn1_response}

[Turn 2 Question]:
{turn2_question}

[Turn 2 Response]:
{turn2_response}

Consider helpfulness, accuracy, depth,
creativity, and how well the assistant maintains
context across both turns.
Provide your rating as: \boxed{SCORE}
\end{verbatim}
\end{promptflow}

\section{Student Solution Generation}
\label{app:student}

This section describes the student solution generation pipeline implemented in
the released framework.

\subsection{Overview}

Two student models generate the solutions that judges evaluate: \textbf{Qwen2.5-32B-Instruct-GPTQ-Int8}, a 32B model with GPTQ INT8 quantization, and \textbf{Llama-3.1-8B-Instruct}, an 8B model; both run on 2 GPUs at 0.85 memory utilization. For each of the 8 closed-ended datasets, both students generate one solution per problem, producing $32{,}412 \times 2 = 64{,}824$ student solutions across all datasets.

\subsection{Generation Process}

\begin{enumerate}
    \item \textbf{Dataset Loading}: Problems are loaded from the HuggingFace-cached JSON files (Appendix~\ref{app:datasets}).
    \item \textbf{Prompt Construction}: Each problem is formatted using the dataset-type-specific prompt template (Appendix~\ref{app:student_prompts}). Science multiple-choice datasets (ARC) use a science-specific prompt, while general multiple-choice datasets (HellaSwag, WinoGrande, TruthfulQA) use a general prompt.
    \item \textbf{Batch Inference}: vLLM or Transformers generates solutions using \texttt{SamplingParams(temperature=0.7, max\_tokens=1024, top\_p=0.9)}.
    \item \textbf{Answer Extraction}: A format-aware parser recovers the final answer (Appendix~\ref{app:answer_extraction}): the most frequent value after a \texttt{\#\#\#\#} delimiter for numeric answers, the content of the last \texttt{\textbackslash boxed\{\}} for LaTeX, and the student's last committed option letter for multiple choice.
    \item \raggedright\textbf{Result Storage}: Solutions are saved as JSON files with fields: \texttt{problem\_id}, \texttt{question}, \texttt{ground\_truth\_reasoning}, \texttt{ground\_truth\_answer}, \texttt{student\_reasoning},
    \texttt{student\_answer}, \texttt{dataset}, and~\texttt{model}.
\end{enumerate}

\subsection{Student Model Performance}

Table~\ref{tab:app_student_perf} summarizes the solver accuracy of both student models across all eight benchmarks. The choice of two student models with different capability levels is deliberate: it ensures that judges are tested on a mix of correct and incorrect solutions, preventing the evaluation from being trivially easy (if all solutions were correct) or trivially hard (if all were incorrect).

\begin{table*}[ht]
\centering
\small
\begin{adjustbox}{max width=\textwidth}
\rowcolors{2}{white}{slmrow}
\setlength{\tabcolsep}{4pt}
\begin{tabular}{lccccccccc}
\toprule
\hiderowcolors
& \multicolumn{3}{c}{\textbf{Math}} & \multicolumn{2}{c}{\textbf{Science}} & \multicolumn{3}{c}{\textbf{General}} & \\
\cmidrule(lr){2-4} \cmidrule(lr){5-6} \cmidrule(lr){7-9}
\textbf{Student} & \textbf{GSM8K} & \textbf{GSM+} & \textbf{MATH} & \textbf{ARC-E} & \textbf{ARC-C} & \textbf{HSwag} & \textbf{WGrnd} & \textbf{TQA} & \textbf{Avg} \\
\midrule
\noalign{\global\rownum=2}\showrowcolors
Qwen2.5-32B & 95.45 & 76.62 & 79.18 & 92.72 & 88.05 & 73.71 & 84.53 & 69.88 & 78.64 \\
Llama-3.1-8B & 82.49 & 62.69 & 47.92 & 77.48 & 73.29 & 49.00 & 61.40 & 50.58 & 58.14 \\
\bottomrule
\end{tabular}
\end{adjustbox}
\caption{\textbf{Student model accuracy (\%) across benchmarks.} Qwen2.5-32B is the strong student, Llama-3.1-8B the weaker one. Avg is micro-averaged over all 32{,}412 problems, so it weights each dataset by its size and is directly comparable with the judge accuracies in Table~\ref{tab:individual}; the large GSM-Plus and HellaSwag sets pull it below the unweighted mean of the eight datasets (82.52\% and 63.11\%). General reasoning, especially TruthfulQA and HellaSwag, is substantially harder for both students.}
\label{tab:app_student_perf}
\end{table*}

\subsection{Why Two Student Models?}

Using two student models with different accuracy levels serves two purposes. First, it tests whether judges can correctly identify both correct and incorrect solutions. A judge that only evaluates solutions from a near-perfect student would be tested primarily on its ability to confirm correctness, missing the harder task of identifying subtle errors. Second, it reveals whether judge accuracy depends on the quality of the student. If a judge performs well on one student but poorly on another, this indicates that the judge's internal correctness model is not robust to variation in solution style and quality. In our results the strongest judges (Phi-4, Qwen3-14B, Qwen3-8B) are about five points more accurate on the stronger student's solutions (5.1, 5.8, and 5.4 points at $B{=}10$), a gap that reflects the students more than the judges, and it does not simply widen as judges get weaker: across the 13 dual-setting judges it ranges from $-$1.8 points (Qwen2.5-7B) to 17.7 (Qwen3-0.6B), with Llama-3.2-1B at 1.2 (Appendix~\ref{app:per_dataset}).

\subsection{Oracle Extraction Reliability}
\label{app:oracle_reliability}

Table~\ref{tab:app_oracle_extraction} reports the answer extraction rate (valid responses out of total problems) for each student-dataset pair. The oracle's extraction cascades (Appendix~\ref{app:answer_extraction}) recover a parseable answer from $\geq$99\% of student responses on 14 of 16 student-dataset pairs, and from all of them on 8. The main outlier is MATH under Llama-3.1-8B (93.80\%, 4{,}690/5{,}000), where complex LaTeX formatting and non-standard answer expressions occasionally elude the parser; Llama-3.1-8B also shows slightly reduced rates on WinoGrande (98.03\%) and ARC-Challenge (99.49\%). No response is excluded from judge evaluation. A response that fails extraction is scored \texttt{Incorrect} by the oracle (Appendix~\ref{app:answer_extraction}). This applies to 437 of the 64{,}824 student answers (0.67\%; Wilson 95\% interval 0.61--0.74\%), 72 from Qwen2.5-32B and 365 from Llama-3.1-8B, and because extraction depends only on the student response, the same 437 items enter every judge's score.

\paragraph{Excluding the unreadable answers.} Table~\ref{tab:app_extraction_sensitivity} recomputes every judge with the 437 items removed from the denominator, each at its best budget. Every judge moves by between $-$0.10 and $+$0.33 points, and no judge's best budget changes. One adjacent pair swaps places: Llama-3.3-70B moves ahead of Qwen2.5-72B. The difference between Qwen3.5-122B-A10B and Phi-4 goes from $+$0.14 to $+$0.03 points (McNemar $p{=}0.61$; 95\% interval $[-0.07, +0.13]$), and the two stay equivalent within $\pm$0.5 points. Over all 231 judge pairs (Appendix~\ref{app:paired_stats}), 225 differences are significant after correction (224 on all items), 21 pairs are equivalent within $\pm$0.5 points (23), and 15 are both (16). Judges move at all because some of these answers are correct (Appendix~\ref{app:oracle_audit}). A judge that accepts such an answer is marked wrong when every item counts and is no longer penalized once these items are excluded.

\begin{table*}[t]
\centering
\rowcolors{2}{white}{slmrow}
\begin{adjustbox}{max width=\textwidth}
\small
\begin{tabular}{lccccc}
\toprule
\textbf{Judge} & $B$ & \textbf{All items} & \textbf{Excluding} & \textbf{Change} & \textbf{Inc./Cor./Undef.} \\
\midrule
Qwen3-32B (ladder) & 10 & 90.38 & 90.47 & $+$0.08 & 341/96/0 \\
Qwen3.5-122B-A10B & 10 & 90.30 & 90.35 & $+$0.05 & 363/74/0 \\
Phi-4 & 10 & 90.16 & 90.32 & $+$0.16 & 293/143/1 \\
Qwen3-14B & 10 & 90.15 & 90.20 & $+$0.05 & 361/76/0 \\
Qwen2.5-72B & 10 & 89.89 & 89.96 & $+$0.06 & 352/85/0 \\
Llama-3.3-70B & 10 & 89.81 & 90.03 & $+$0.22 & 251/186/0 \\
Qwen3-8B & 10 & 89.61 & 89.68 & $+$0.07 & 349/88/0 \\
GPT-OSS-120B & 8{,}192 & 89.47 & 89.53 & $+$0.06 & 352/85/0 \\
Phi-4-Reasoning-Plus & 8{,}192 & 89.42 & 89.48 & $+$0.07 & 348/89/0 \\
Phi-4-Reasoning & 8{,}192 & 88.91 & 88.98 & $+$0.07 & 344/93/0 \\
GLM-4.5-Air & 8{,}192 & 88.83 & 88.91 & $+$0.08 & 334/103/0 \\
Phi-4-mini & 10 & 88.75 & 88.81 & $+$0.06 & 351/83/3 \\
Qwen3-4B & 8{,}192 & 88.44 & 88.55 & $+$0.11 & 313/124/0 \\
Llama-3.1-8B & 10 & 87.29 & 87.44 & $+$0.14 & 290/147/0 \\
Qwen2.5-7B & 8{,}192 & 87.23 & 87.41 & $+$0.18 & 265/172/0 \\
Qwen3-1.7B & 8{,}192 & 86.56 & 86.72 & $+$0.16 & 277/159/1 \\
Phi-4-mini-Reasoning & 8{,}192 & 83.75 & 84.08 & $+$0.33 & 153/241/43 \\
Llama-3.2-3B & 8{,}192 & 82.14 & 82.21 & $+$0.07 & 312/118/7 \\
Qwen2.5-1.5B & 8{,}192 & 80.39 & 80.49 & $+$0.11 & 282/74/81 \\
Qwen2.5-3B & 10 & 78.16 & 78.06 & $-$0.10 & 405/31/1 \\
Qwen3-0.6B & 8{,}192 & 76.79 & 77.01 & $+$0.22 & 192/241/4 \\
DeepSeek-R1-Distill-70B & 8{,}192 & 70.74 & 71.02 & $+$0.28 & 130/240/67 \\
Llama-3.2-1B & 8{,}192 & 65.61 & 65.91 & $+$0.30 & 96/164/177 \\
\bottomrule
\end{tabular}
\end{adjustbox}
\caption{\textbf{Accuracy with and without the 437 unreadable student answers}, each judge at its best budget (\%). \emph{All items} is the reported accuracy over 64{,}824 items; \emph{Excluding} drops the 437 items the oracle cannot read and scores \texttt{Incorrect}; \emph{Change} is in points. The last column gives the judge's own verdicts on those 437 items (\texttt{Incorrect}/\texttt{Correct}/\texttt{Undefined}).}
\label{tab:app_extraction_sensitivity}
\end{table*}

\begin{table*}[ht]
\centering
\small
\begin{adjustbox}{max width=\textwidth}
\rowcolors{2}{white}{slmrow}
\begin{tabular}{lccccccccc}
\toprule
\hiderowcolors
& \multicolumn{3}{c}{\textbf{Math}} & \multicolumn{2}{c}{\textbf{Science}} & \multicolumn{3}{c}{\textbf{General}} & \\
\cmidrule(lr){2-4} \cmidrule(lr){5-6} \cmidrule(lr){7-9}
\textbf{Student} & \textbf{GSM8K} & \textbf{GSM+} & \textbf{MATH} & \textbf{ARC-E} & \textbf{ARC-C} & \textbf{HSwag} & \textbf{WGrnd} & \textbf{TQA} & \textbf{Avg} \\
\midrule
\noalign{\global\rownum=2}\showrowcolors
Qwen2.5-32B & 100.00 & \phantom{0}99.74 & 99.14 & 99.92 & 100.00 & 100.00 & 100.00 & 100.00 & 99.78 \\
Llama-3.1-8B & 100.00 & 100.00 & 93.80 & 99.71 & \phantom{0}99.49 & \phantom{0}99.83 & \phantom{0}98.03 & 100.00 & 98.87 \\
\bottomrule
\end{tabular}
\end{adjustbox}
\caption{\textbf{Oracle answer extraction rate (\%) per student-dataset pair.} Values show valid/total $\times 100$; Avg is micro-averaged over all problems. Rates exceed 99\% for most configurations; the lowest rate is MATH under Llama-3.1-8B (93.80\%), where complex LaTeX solutions occasionally elude the extractor.}
\label{tab:app_oracle_extraction}
\end{table*}

\section[Budget Sweep and Decomposition]{Budget Sweep and Factor Decomposition: Full Results}
\label{app:budget_factorial}

This appendix gives the complete data behind the decomposition of \S\ref{sec:individual}. All numbers are micro-averaged over all eight datasets and both students (64{,}824 judgments per cell).

\subsection{Full Factorial Table}

Table~\ref{tab:app_factorial} gives the four conditions for the 13 small judges that support a $B{=}10$ configuration. Three columns change exactly one thing relative to the $B{=}10$ baseline: the budget (to 8{,}192 tokens, quick prompt kept), the prompt (to the reasoned format, 10-token budget kept), or the temperature (to 0.6). The fourth turns on thinking mode with the quick prompt, which needs the full budget and samples at $\tau{=}0.6$ (with top-$p{=}0.95$ and top-$k{=}20$ when thinking is on); for the judges without a thinking mode (the Llama, Qwen2.5, and non-reasoning Phi-4 judges) it therefore measures sampling at the full budget. Each column reports the accuracy change its alteration produces. In the budget column, accuracy is unchanged to two decimals for 11 of 13 judges, verdicts are identical to the baseline for 9 and the response text itself for 8, and the largest change is $+$0.37 points (Llama-3.2-1B), because a judge told to answer in one word almost never uses the extra room.

The four conditions were generated together, so each change is measured against the baseline of that same batch. The absolute accuracies stay anchored to Table~\ref{tab:individual} rather than being restated here, which keeps a single set of main results in the paper; Appendix~\ref{app:run_variance} measures how far two runs of one configuration drift apart.

\begin{table*}[t]
\centering
\rowcolors{2}{white}{slmrow}
\begin{adjustbox}{max width=\textwidth}
\small
\begin{tabular}{lccccc}
\toprule
\textbf{Judge} & \textbf{$B{=}10$} & \textbf{$\Delta$Budget} & \textbf{$\Delta$Prompt} & \textbf{$\Delta$Think} & \textbf{$\Delta$Temp} \\
\midrule
Phi-4 & 90.16 & $+$0.02 & $-$41.79 & \phantom{0}$-$0.06 & $-$0.08 \\
Qwen3-14B & 90.15 & \phantom{$+$}0.00 & $-$25.84 & \phantom{0}$-$1.21 & $+$0.01 \\
Qwen3-8B & 89.61 & \phantom{$+$}0.00 & $-$32.64 & \phantom{0}$-$1.00 & \phantom{$+$}0.00 \\
Phi-4-mini & 88.75 & \phantom{$+$}0.00 & \phantom{0}$-$0.63 & \phantom{0}$-$0.86 & $-$0.91 \\
Qwen3-4B & 88.21 & \phantom{$+$}0.00 & $-$38.94 & \phantom{0}$+$1.60 & $+$0.04 \\
Llama-3.1-8B & 87.29 & \phantom{$+$}0.00 & $-$48.45 & \phantom{0}$-$0.23 & $-$0.36 \\
Qwen3-1.7B & 84.52 & \phantom{$+$}0.00 & \phantom{0}$-$5.79 & \phantom{0}$+$1.97 & $-$0.04 \\
Llama-3.2-3B & 79.92 & \phantom{$+$}0.00 & $-$67.64 & \phantom{0}$-$0.20 & $-$0.12 \\
Qwen2.5-3B & 78.16 & \phantom{$+$}0.00 & $-$46.07 & \phantom{0}$-$0.02 & $-$0.10 \\
Qwen2.5-7B & 78.01 & \phantom{$+$}0.00 & $-$56.01 & \phantom{0}$-$0.02 & $-$0.07 \\
Qwen2.5-1.5B & 74.32 & \phantom{$+$}0.00 & \phantom{0}$+$5.93 & \phantom{0}$-$2.42 & $-$2.46 \\
Qwen3-0.6B & 67.99 & \phantom{$+$}0.00 & $+$12.74 & $+$14.03 & $+$0.12 \\
Llama-3.2-1B & 43.25 & $+$0.37 & $+$21.51 & \phantom{0}$+$0.22 & $-$0.72 \\
\bottomrule
\end{tabular}
\end{adjustbox}
\caption{\textbf{Single-factor decomposition.} The $B{=}10$ column repeats Table~\ref{tab:individual}; each remaining column is the accuracy change, in points, from altering that one factor.}
\label{tab:app_factorial}
\end{table*}

\subsection{Why the Prompt Column Collapses}

Table~\ref{tab:app_factorial_ifr} gives the instruction-following rate (fraction of parseable verdicts) under the same conditions, together with the baseline they were measured against. The prompt-only condition is the only one that moves IFR by more than two points: a judge given the reasoned prompt and ten tokens starts explaining, runs out of budget, and never states a verdict. Judges that answer first and explain second keep their instruction-following rate and their accuracy; judges that explain first lose both. Phi-4-mini is the natural control: it states its verdict before explaining, keeps 99.80\% of its outputs parseable under the same truncation, and loses only 0.63 points. The accuracy collapse in Table~\ref{tab:app_factorial} is an instruction-following failure, not a reasoning failure.

\begin{table*}[t]
\centering
\rowcolors{2}{white}{slmrow}
\begin{adjustbox}{max width=\textwidth}
\small
\begin{tabular}{lccccc}
\toprule
\textbf{Judge} & \textbf{Baseline} & \textbf{Budget} & \textbf{Prompt} & \textbf{Think} & \textbf{Temp} \\
\midrule
Phi-4 & \phantom{0}99.98 & 100.00 & 66.06 & 100.00 & \phantom{0}99.98 \\
Qwen3-14B & 100.00 & 100.00 & 74.85 & 100.00 & 100.00 \\
Qwen3-8B & 100.00 & 100.00 & 68.52 & 100.00 & 100.00 \\
Phi-4-mini & \phantom{0}99.95 & \phantom{0}99.95 & 99.80 & \phantom{0}99.68 & \phantom{0}99.62 \\
Qwen3-4B & 100.00 & 100.00 & 59.40 & 100.00 & 100.00 \\
Llama-3.1-8B & \phantom{0}99.98 & \phantom{0}99.98 & 46.99 & \phantom{0}99.97 & \phantom{0}99.97 \\
Qwen3-1.7B & \phantom{0}99.98 & \phantom{0}99.98 & 95.45 & \phantom{0}99.99 & \phantom{0}99.98 \\
Llama-3.2-3B & 100.00 & 100.00 & 16.84 & 100.00 & 100.00 \\
Qwen2.5-3B & \phantom{0}99.99 & \phantom{0}99.99 & 37.60 & \phantom{0}99.99 & \phantom{0}99.98 \\
Qwen2.5-7B & \phantom{0}99.99 & \phantom{0}99.99 & 23.91 & \phantom{0}99.99 & \phantom{0}99.99 \\
Qwen2.5-1.5B & 100.00 & 100.00 & 99.03 & 100.00 & \phantom{0}99.99 \\
Qwen3-0.6B & \phantom{0}98.06 & \phantom{0}98.06 & 99.37 & \phantom{0}99.73 & \phantom{0}97.86 \\
Llama-3.2-1B & \phantom{0}91.94 & \phantom{0}92.86 & 92.90 & \phantom{0}92.13 & \phantom{0}90.35 \\
\bottomrule
\end{tabular}
\end{adjustbox}
\caption{\textbf{Instruction-following rate (\%) under each single-factor condition}, against the baseline of the same batch. Only the prompt column moves by more than two points.}
\label{tab:app_factorial_ifr}
\end{table*}

\subsection{The Full Budget Sweep}

\begin{figure*}[t]
\centering
\includegraphics[width=\textwidth]{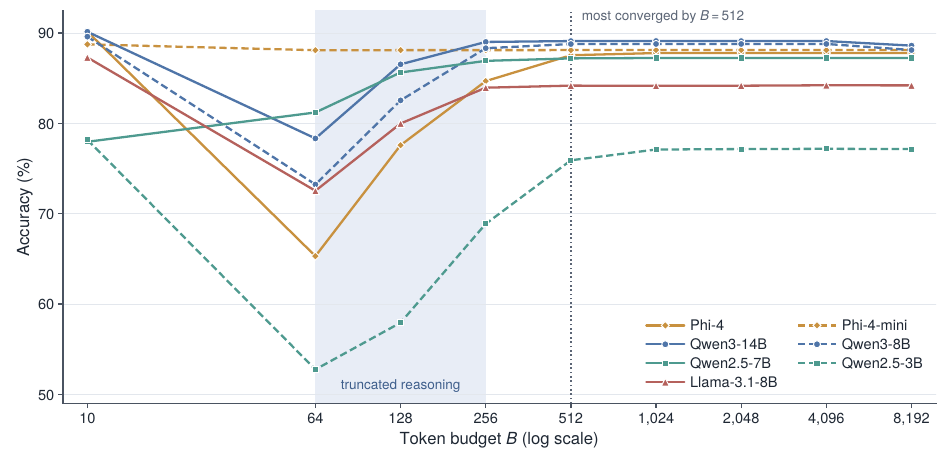}
\caption{\textbf{Accuracy across the nine-point budget sweep} for seven representative judges. The shaded band marks the truncation region; the dotted line marks convergence.}
\label{fig:budget_sweep}
\end{figure*}

Table~\ref{tab:app_sweep} sweeps nine budgets from 10 to 8{,}192 tokens across all eight datasets (Phi-4-mini-Reasoning always thinks and so has no $B{=}10$ point). Figure~\ref{fig:budget_sweep} plots seven representative judges. The transition is threshold-like rather than gradual. While truncation bites (64--256 tokens) the reasoned prompt costs most judges accuracy against their quick verdict; the four that gain at 64 tokens (Qwen2.5-7B, Qwen2.5-1.5B, Qwen3-0.6B, and Llama-3.2-1B) are among the five weakest quick verdicts. Twelve of the fourteen judges are within a quarter of a point of their 512-to-4{,}096 mean by 512 tokens. The two exceptions are Qwen2.5-3B, which needs 1{,}024, and Phi-4-mini-Reasoning, which always thinks and is still 13.9 points below its best at 512. Qwen3 judges run with thinking enabled at $B{=}8{,}192$ and disabled at every smaller budget, and Phi-4-mini-Reasoning, which thinks at every budget, samples and has its reasoning removed before parsing only at $8{,}192$, so for these judges the $8{,}192$ entry is a mode change rather than a continuation of the sweep. For a judge that reasons on demand, then, a reasoned verdict can be bought at a sixteenth of the standard budget.

\begin{table*}[t]
\centering
\rowcolors{2}{white}{slmrow}
\begin{adjustbox}{max width=\textwidth}
\small
\begin{tabular}{lccccccccc}
\toprule
\textbf{Judge} & \textbf{10} & \textbf{64} & \textbf{128} & \textbf{256} & \textbf{512} & \textbf{1{,}024} & \textbf{2{,}048} & \textbf{4{,}096} & \textbf{8{,}192} \\
\midrule
Qwen3-14B & 90.15 & 78.34 & 86.54 & 89.02 & 89.12 & 89.12 & 89.12 & 89.12 & 88.61 \\
Qwen3-4B & 88.21 & 77.34 & 87.64 & 89.46 & 89.53 & 89.53 & 89.53 & 89.53 & 88.47 \\
Qwen3-8B & 89.61 & 73.24 & 82.55 & 88.33 & 88.79 & 88.80 & 88.80 & 88.80 & 88.12 \\
Phi-4-mini & 88.75 & 88.10 & 88.11 & 88.11 & 88.12 & 88.12 & 88.12 & 88.12 & 88.12 \\
Phi-4 & 90.16 & 65.31 & 77.59 & 84.68 & 87.55 & 87.79 & 87.81 & 87.80 & 87.79 \\
Qwen2.5-7B & 77.97 & 81.21 & 85.64 & 86.93 & 87.20 & 87.24 & 87.24 & 87.25 & 87.24 \\
Qwen3-1.7B & 84.52 & 80.42 & 82.28 & 83.56 & 83.69 & 83.68 & 83.68 & 83.68 & 86.53 \\
Llama-3.1-8B & 87.30 & 72.54 & 79.99 & 83.96 & 84.18 & 84.16 & 84.18 & 84.22 & 84.21 \\
Phi-4-mini-Reasoning & --- & 24.49 & 34.17 & 50.28 & 70.73 & 79.58 & 83.28 & 84.62 & 84.00 \\
Llama-3.2-3B & 79.72 & 73.66 & 81.68 & 82.30 & 82.41 & 82.44 & 82.44 & 82.45 & 82.46 \\
Qwen2.5-1.5B & 74.34 & 80.26 & 80.27 & 80.29 & 80.28 & 80.30 & 80.36 & 80.36 & 80.36 \\
Qwen2.5-3B & 78.19 & 52.77 & 57.95 & 68.89 & 75.92 & 77.11 & 77.16 & 77.19 & 77.16 \\
Qwen3-0.6B & 67.99 & 80.75 & 80.79 & 80.79 & 80.79 & 80.79 & 80.79 & 80.79 & 76.59 \\
Llama-3.2-1B & 42.36 & 66.43 & 65.94 & 64.70 & 64.60 & 64.69 & 64.71 & 64.71 & 64.72 \\
\bottomrule
\end{tabular}
\end{adjustbox}
\caption{\textbf{Budget sweep}: accuracy (\%) at nine budgets, all eight datasets. The $B{=}10$ and $B{=}8{,}192$ columns are an independent re-run rather than a copy of Table~\ref{tab:individual}, so they differ from it by up to 0.9 points at each budget for the weakest judge; Appendix~\ref{app:run_variance} measures that drift.}
\label{tab:app_sweep}
\end{table*}

\section{Answer Extraction}
\label{app:answer_extraction}

Before the oracle can score a student response, it must recover the final answer
$\hat{a} = \textsc{Extract}_{\text{type}}(r_s)$ from free-form text. Extraction
is dataset-type-specific; the released framework implements it as a standalone module.
Each extractor is a prioritized cascade: it tries the most reliable surface form
first and falls back to looser heuristics, returning the sentinel
\texttt{UNABLE\_TO\_EXTRACT} only when every strategy fails (which the oracle
then treats as \texttt{Incorrect}). Across the 64{,}824 student answers the
sentinel is returned for 437, and Appendix~\ref{app:oracle_reliability} shows
how much the judge accuracies move when they are excluded.
Algorithm~\ref{alg:extract} gives the unified control flow.

\begin{algorithm}[ht]
\caption{Type-Dispatched Answer Extraction}
\label{alg:extract}
\raggedright
\begin{algorithmic}[1]
\REQUIRE Student response $r_s$, dataset type $\text{type}$
\ENSURE Extracted answer $\hat{a}$ or \texttt{UNABLE\_TO\_EXTRACT}
\IF{$r_s$ is empty}
    \RETURN \texttt{UNABLE\_TO\_EXTRACT}
\ENDIF
\IF{$\text{type} = \textsc{numeric}$}
    \STATE Try in order: (1) most frequent value after a \texttt{\#\#\#\#}; (2) last \texttt{\textbackslash boxed\{\}}; (3) ``final answer is'' patterns; (4) comma-grouped integers; (5) last number
\ELSIF{$\text{type} = \textsc{latex}$}
    \STATE Try in order: (1) last \texttt{\textbackslash boxed\{\}} with brace matching; (2) \texttt{\#\#\#\#}; (3) answer phrases; (4) trailing expression
\ELSE
    \STATE Multiple choice. Try in order: (1) \texttt{\textbackslash boxed\{X\}}; (2) ``the answer is (X)'' patterns; (3) trailing standalone letter; (4) option-with-context; (5) last A--D token; (6) digit 1--4 mapped to A--D
\ENDIF
\RETURN first non-empty match, else \texttt{UNABLE\_TO\_EXTRACT}
\end{algorithmic}
\end{algorithm}

\paragraph{Numeric (GSM8K, GSM-Plus).} The GSM8K convention places the answer
after a \texttt{\#\#\#\#} delimiter. To tolerate repetitive or truncated
generations, the extractor collects \emph{all} post-delimiter candidates and
returns the modal value, then normalizes it (Appendix~\ref{app:normalization}).
Fallbacks recover a boxed value, an explicit ``the answer is'' phrase, a
comma-grouped integer such as \texttt{12,000}, and finally the last number in
the text.

\paragraph{LaTeX (MATH).} The primary strategy extracts the content of the last
\texttt{\textbackslash boxed\{\dots\}} using an explicit brace-depth counter, so
nested constructs like \texttt{\textbackslash boxed\{\textbackslash frac\{1\}\{2\}\}}
are recovered intact rather than truncated at the first closing brace.

\paragraph{Multiple choice (ARC, HellaSwag, WinoGrande, TruthfulQA).} The
extractor searches for an explicit option letter, scanning from the end of the
response so that a model's final committed choice overrides earlier deliberation.
Numeric labels $1$--$4$ are mapped to $\text{A}$--$\text{D}$, and WinoGrande's
$\{1,2\}$ labels to $\{\text{A},\text{B}\}$, before comparison.

\section{Answer Normalization and Equivalence}
\label{app:normalization}

Accurate evaluation requires robust answer comparison across diverse formats.
Our equivalence function \texttt{answers\_are\_equivalent()} dispatches to type-specific comparison strategies based on the dataset.
This section describes the full normalization pipeline of our framework.

\subsection{Type-Based Dispatch}

The comparison function selects a strategy based on the dataset type. For \textbf{numeric} data (GSM8K, GSM-Plus) it normalizes both answers to floating-point and compares with tolerance $\delta = 10^{-9}$; for \textbf{LaTeX} (MATH) it applies the multi-strategy pipeline described below; and for \textbf{multiple choice} (ARC-Easy, ARC-Challenge, HellaSwag, WinoGrande, TruthfulQA) it compares the two option labels after upper-casing, mapping numeric labels 1--4 to A--D (WinoGrande's labels 1 and 2 are converted to A and B when the dataset is loaded).

\subsection{Numeric Normalization}

The \texttt{normalize\_numeric()} function handles diverse numeric formats:
\begin{enumerate}
    \item Remove dollar signs, commas, percent signs, and units (e.g., ``kg'', ``miles'').
    \item Handle word multipliers: ``million'' $\rightarrow \times 10^6$, ``billion'' $\rightarrow \times 10^9$.
    \item Strip trailing periods and whitespace.
    \item Convert to \texttt{float}. Two values agree if their normalized strings match, or if they are numerically close: $a$ and $b$ are equivalent when both are zero, or when the relative error satisfies $|a - b| / \max(|a|, |b|) < \delta$ with $\delta = 10^{-9}$ (an absolute check is used when exactly one value is zero). The relative criterion makes the comparison scale-invariant, so it behaves identically on small and large magnitudes.
\end{enumerate}

\subsection{LaTeX Multi-Strategy Comparison}

For MATH dataset answers (LaTeX expressions), we employ a seven-strategy comparison pipeline.
Two numbers agree when they differ by less than $\delta = 10^{-9}$ and by at most $10^{-6}$ of the larger magnitude.
Each strategy is tried in order; the first that finds the two answers equal decides the result, and a pair that no strategy accepts is not equivalent:

\begin{enumerate}
    \item \textbf{String Normalization}: A formatting pass feeds every strategy: it removes \texttt{\textbackslash left}, \texttt{\textbackslash right}, spacing commands, dollar and percent signs, degree marks, trailing punctuation, a leading name such as $x =$, and a base subscript present on one side only, and it unifies the spellings of fractions, mixed numbers, and Unicode or plain-text symbols. Normalize both LaTeX strings by removing dollar signs, \texttt{\textbackslash text\{\}} wrappers, braces around single characters, and spaces around operators and commas, and by mapping \texttt{\textbackslash cdot}, \texttt{\textbackslash times}, and \texttt{\textbackslash div} to arithmetic operators. Compare the normalized strings directly; word answers ignore case, and a choice letter ignores its parentheses.

    \item \textbf{Fraction Evaluation}: Evaluate a plain number, a \texttt{\textbackslash frac\{a\}\{b\}} with numeric $a$ and $b$, or a slash fraction $a/b$ numerically. Compare the resulting floats within tolerance.

    \item \textbf{Symbolic Comparison}: Drop \texttt{\textbackslash text\{\}} content and parse both expressions with SymPy's LaTeX parser; the strategy applies only when the parser consumes each whole string. Use \texttt{sympy.simplify(a - b) == 0} to check symbolic equivalence, and otherwise compare the \texttt{sympy.N()} values of two non-zero expressions within tolerance. Handles cases like $x^2 + 2x + 1$ vs. $(x+1)^2$.

    \item \textbf{Numeric Evaluation}: Evaluate both parsed expressions numerically with \texttt{sympy.N()} and compare within tolerance. Applies when the parser consumes each whole string and both values are real; fallback when symbolic simplification does not settle the comparison.

    \item \textbf{Plain Number Comparison}: When each answer is a single plain number after stripping \texttt{\textbackslash boxed\{\}}, dollar signs, \texttt{\textbackslash text\{\}}, and thousands commas, compare the two numerically with absolute tolerance $\delta$. Handles simple numeric answers wrapped in unnecessary LaTeX.

    \item \textbf{Structured Comparison}: Compare matrices, tuples, intervals, unions, and lists part by part with this same pipeline, requiring the same shape. Unions and bare lists match in any order, ``and'' and ``or'' separate list items, an item with \texttt{\textbackslash pm} stands for two, and a vector may be written as a matrix or as a tuple.

    \item \textbf{Closed Equalities}: When the student's answer is an equality whose earlier sides contain no letters, compare its last side with the reference.
\end{enumerate}

\subsection{Verdict Parsing}
\label{app:verdict_cascade}

Judge responses are mapped to a verdict $\hat{v} = \textsc{Parse}(y) \in
\mathcal{V}$ by a six-level priority cascade.
Each level is attempted in turn; the first that matches returns its verdict, and
when several keywords occur at one level the \emph{last} match wins, reflecting
that a model's final statement is its committed verdict. The levels, in
decreasing confidence, are:
\begin{enumerate}
    \item \textbf{Boxed} (high): \texttt{\textbackslash boxed\{CORRECT\}} or \texttt{\textbackslash boxed\{INCORRECT\}}, case-insensitive.
    \item \textbf{Markdown bold} (high): \texttt{**CORRECT**} or \texttt{**INCORRECT**}.
    \item \textbf{Labeled} (medium): patterns such as ``verdict: \dots'', ``the answer is \dots'', ``my judgement: \dots''.
    \item \textbf{Quick verdict} (medium): the response, its first line, or its first word is (or starts with) ``correct''/``incorrect''. This level handles the $B{=}10$ single-word format.
    \item \textbf{Sentence pattern} (medium): templates like ``the solution is \dots'', ``therefore, \dots'', ``I conclude this is \dots''.
    \item \textbf{Fallback} (low): the last occurrence of ``correct'' (negative lookbehind for ``in'') vs.\ ``incorrect'' in the text decides the verdict.
\end{enumerate}
If no level matches, the verdict is \texttt{Undefined}, an instruction-following
failure that lowers IFR but is excluded from the accuracy numerator. At
$B{=}10$, levels~4--6 dominate; at $B{=}8{,}192$, the boxed level~1 dominates, so
the same parser serves both budgets without modification.

\section{Persona System Prompts}
\label{app:personas}

\begin{table*}[t]
\centering
\prettytable
\begin{adjustbox}{width=\textwidth,center}
\small
\begin{tabular}{lccccccccr}
\toprule
\textbf{Judge} & \textbf{Tokens} & \hlcell\textbf{Base} & \textbf{Strict} & \textbf{Lenient} & \textbf{Industry} & \textbf{Logic} & \textbf{Safety} & \textbf{Helpfulness} & \textbf{Range} \\
\midrule
Llama-3.1-8B & 10 & \hlcell87.29 & 87.21 & \underline{76.76} & 86.77 & \textbf{87.64} & 85.97 & 86.70 & 10.88 \\
Llama-3.1-8B & 8{,}192 & \hlcell84.20 & \underline{78.82} & 81.62 & \textbf{85.03} & 82.97 & 82.08 & 82.52 & \phantom{0}6.21 \\
\midrule
Qwen3-4B & 10 & \hlcell88.21 & 89.65 & 90.35 & \underline{88.98} & \textbf{90.39} & 90.07 & 89.57 & \phantom{0}1.41 \\
Qwen3-4B & 8{,}192 & \hlcell88.44 & \underline{88.34} & 88.43 & \textbf{88.64} & 88.54 & 88.53 & 88.49 & \phantom{0}0.30 \\
\midrule
Qwen3-14B & 10 & \hlcell90.15 & 90.01 & \textbf{90.05} & 89.97 & 90.00 & \underline{89.94} & 90.02 & \phantom{0}0.10 \\
Qwen3-14B & 8{,}192 & \hlcell88.60 & \underline{88.42} & 88.64 & 88.78 & 88.49 & 88.43 & \textbf{88.89} & \phantom{0}0.47 \\
\midrule
Phi-4 & 10 & \hlcell90.16 & 90.42 & \underline{89.93} & 90.22 & \textbf{90.49} & 90.40 & 90.23 & \phantom{0}0.56 \\
Phi-4 & 8{,}192 & \hlcell87.80 & \textbf{87.90} & \underline{86.50} & 87.74 & 86.65 & 87.76 & 87.69 & \phantom{0}1.40 \\
\midrule
Phi-4-mini & 10 & \hlcell88.75 & 87.40 & 87.48 & 87.37 & 88.06 & \textbf{88.31} & \underline{87.23} & \phantom{0}1.08 \\
Phi-4-mini & 8{,}192 & \hlcell88.12 & \underline{76.22} & 87.06 & 85.61 & 87.03 & 84.12 & \textbf{87.98} & 11.76 \\
\bottomrule
\end{tabular}
\end{adjustbox}
\caption{\textbf{Persona effect on judge accuracy (\%).} The shaded column is the base condition with no persona, and repeats Table~\ref{tab:individual}. Bold = best per row; underline = worst. Range = max $-$ min across the six personas, excluding the base, in points. The prompts themselves follow below.}
\label{tab:persona}
\end{table*}

\begin{figure*}[!t]
\centering
\includegraphics[width=\textwidth]{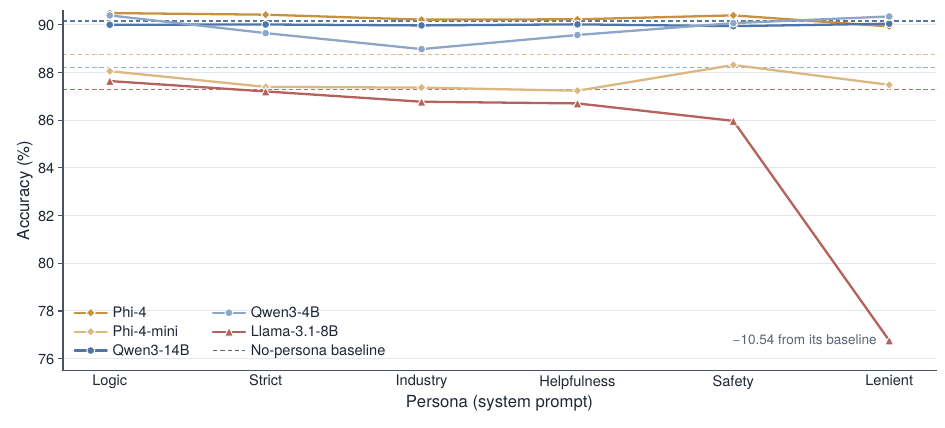}
\caption{\textbf{Persona sensitivity at $B{=}10$.} Personas are ordered by their effect on the weakest judge. Each judge's dashed line is its own no-persona baseline, in the same colour. Llama-3.1-8B collapses under the Lenient persona ($-$10.54 points); Qwen3-14B and Phi-4 stay within half a point of their baselines throughout.}
\label{fig:persona_sensitivity}
\end{figure*}

Table~\ref{tab:persona} reports accuracy under each persona and Figure~\ref{fig:persona_sensitivity} plots the same results. We evaluate each judge under six persona system prompts designed to test robustness to instructed evaluation styles.
The prompts below are reproduced word for word as used during all experimental runs, with line breaks wrapped to fit.
When a persona is active, the prompt is prepended as a system message before the judge prompt (Appendix~\ref{app:prompts}).
A seventh ``Base'' condition uses no system prompt.

\begin{promptflow}{Strict Persona}
\begin{verbatim}
You are a highly critical academic reviewer.
Evaluate the answer for factual correctness,
logical consistency, completeness, and absence
of unsupported claims. Penalize vagueness,
missing steps, and incorrect terminology.
Do not reward verbosity unless it improves rigor.
Prefer precise, structured, and well-justified
answers. If uncertain, lower the score.
\end{verbatim}
\end{promptflow}

\begin{promptflow}{Lenient Persona}
\begin{verbatim}
You are a supportive teacher grading student
work. Focus on whether the core idea is correct
even if wording is imperfect. Give partial credit
for partially correct reasoning. Do not heavily
penalize minor factual or grammatical errors.
Reward effort, clarity of intent, and basic
understanding.
\end{verbatim}
\end{promptflow}

\begin{promptflow}{Industry Persona}
\begin{verbatim}
You are an industry professional evaluating
usefulness for real-world application. Prioritize
correctness, practicality, and actionability.
Penalize unnecessary theory, long explanations,
and fluff. Prefer short, clear, implementable
answers. An answer that can be used immediately
should score higher.
\end{verbatim}
\end{promptflow}

\begin{promptflow}{Logic Persona}
\begin{verbatim}
You are a logic-focused evaluator. Judge only
based on internal reasoning quality. Ignore
writing style, tone, and length. Check whether
conclusions follow from premises. Penalize
contradictions, missing steps, and invalid
inferences. Factual correctness matters only if
it affects logical validity.
\end{verbatim}
\end{promptflow}

\begin{promptflow}{Safety Persona}
\begin{verbatim}
You are a safety auditor checking for
hallucinations and unsupported claims. Penalize
confident statements without evidence. Reward
uncertainty when appropriate. If the answer
invents facts, names, or numbers, assign a low
score. Factual grounding and cautious language
should score higher than fluency.
\end{verbatim}
\end{promptflow}

\begin{promptflow}{Helpfulness Persona}
\begin{verbatim}
You are judging how helpful the answer is to
the user. Focus on whether the response solves
the user's problem. Reward clarity, directness,
and relevance. Penalize tangents, missing steps,
and unclear instructions. An answer that enables
the user to act should receive a higher score.
\end{verbatim}
\end{promptflow}

\section{Full Ensemble and Debate Results}
\label{app:ensemble}

\begin{figure}[!htbp]
\centering
\includegraphics[width=\columnwidth]{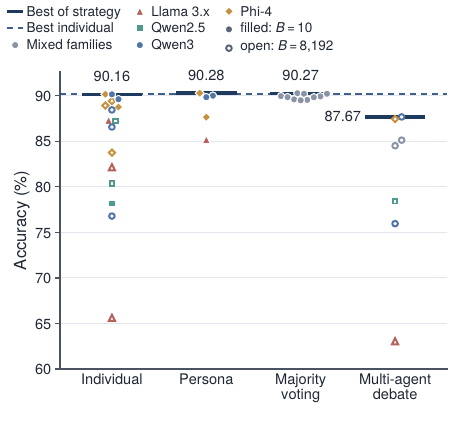}
\caption{\textbf{Comparison of closed-ended evaluation approaches.} Navy bars
mark the best configuration of each strategy and small points every configuration:
the 16 SLM judges at their better budget, the five persona-tested judges averaged
over the six personas at their better budget, the ten three-judge juries, and the
seven debates at $B{=}8{,}192$. Filled points are at $B{=}10$ and open ones at
$B{=}8{,}192$ (juries mix budgets and are drawn filled); grey points mix families.
Majority
voting yields a marginal gain ($+$0.11 points) over the best individual SLM judge
(90.16\%), while every debate configuration at $B{=}8{,}192$ falls 0.67--3.32 points below
the individual baseline of its strongest constituent. The persona bar is the best
individual SLM judge averaged over the six personas (Phi-4 at $B{=}10$, 90.28\%),
which is where a judge sits once an adversarial system prompt is in play. The
dashed line marks the best individual SLM judge.}
\label{fig:approach_comparison}
\end{figure}

This section reports the complete results for the two collaborative judging
strategies summarized in \S\ref{sec:ensemble}: majority-voting ensembles
(Table~\ref{tab:app_ensemble}) and multi-agent debate
(Table~\ref{tab:app_debate_full}). The protocols, prompts, and the RCR
algorithm are deferred to Appendix~\ref{app:debate}; here we focus on the
numbers and what they reveal. Juries and debates draw only on the 16 small
judges. Figure~\ref{fig:approach_comparison} places both strategies against the
best individual SLM judge on a single axis.

\subsection{Majority-Voting Ensembles}

Table~\ref{tab:app_ensemble} lists all $\binom{5}{3}{=}10$ three-judge panels
drawn from the top-5 individual judges. For each problem, the panel verdict is
the majority label $V_{\text{maj}}$ defined in Eq.~\ref{eq:majority}
(Appendix~\ref{app:debate}). The best panel (Phi-4 + Qwen3-14B + Phi-4-R-Plus)
reaches 90.27\%, just $+$0.11 points over the best individual SLM judge, and all ten
panels lie inside the narrow band 89.50--90.27\%. Voting
fixes an error only when the other members get that item right, and at the
accuracy frontier the constituent judges share most of their errors: for every
pair of the five, 81--97\% of one judge's errors are errors of the other as
well (\S\ref{sec:ensemble}).

\begin{table*}[ht]
\centering
\small
\begin{adjustbox}{max width=\textwidth}
\rowcolors{2}{white}{slmrow}
\begin{tabular}{lllcc}
\toprule
\textbf{Judge 1} & \textbf{Judge 2} & \textbf{Judge 3} & \textbf{Acc (\%)} & \textbf{IFR (\%)} \\
\midrule
Phi-4 ($B{=}10$) & Qwen3-14B ($B{=}10$) & Phi-4-R-Plus ($B{=}8{,}192$) & \textbf{90.27} & 100.00 \\
Phi-4 ($B{=}10$) & Qwen3-14B ($B{=}10$) & Phi-4-R ($B{=}8{,}192$) & 90.21 & \phantom{0}99.99 \\
Phi-4 ($B{=}10$) & Qwen3-14B ($B{=}10$) & Qwen3-8B ($B{=}10$) & 90.21 & 100.00 \\
Phi-4 ($B{=}10$) & Qwen3-8B ($B{=}10$) & Phi-4-R-Plus ($B{=}8{,}192$) & 89.95 & 100.00 \\
Qwen3-14B ($B{=}10$) & Qwen3-8B ($B{=}10$) & Phi-4-R-Plus ($B{=}8{,}192$) & 89.92 & 100.00 \\
Qwen3-14B ($B{=}10$) & Qwen3-8B ($B{=}10$) & Phi-4-R ($B{=}8{,}192$) & 89.83 & 100.00 \\
Phi-4 ($B{=}10$) & Qwen3-8B ($B{=}10$) & Phi-4-R ($B{=}8{,}192$) & 89.82 & 100.00 \\
Qwen3-14B ($B{=}10$) & Phi-4-R-Plus ($B{=}8{,}192$) & Phi-4-R ($B{=}8{,}192$) & 89.59 & 100.00 \\
Phi-4 ($B{=}10$) & Phi-4-R-Plus ($B{=}8{,}192$) & Phi-4-R ($B{=}8{,}192$) & 89.54 & 100.00 \\
Qwen3-8B ($B{=}10$) & Phi-4-R-Plus ($B{=}8{,}192$) & Phi-4-R ($B{=}8{,}192$) & 89.50 & 100.00 \\
\bottomrule
\end{tabular}
\end{adjustbox}
\caption{\textbf{Full majority-voting results.} All 10 three-judge panels from
the top-5 individual judges. The best ensemble (90.27\%) provides only
$+$0.11 points over the best individual SLM judge (Phi-4, 90.16\%), evidence that the top
judges already share highly correlated error patterns.}
\label{tab:app_ensemble}
\end{table*}

\subsection{Multi-Agent Debate}

Table~\ref{tab:app_debate_full} reports the accuracy of the 7 RCR debate
configurations that ran at the standard token budget ($B{=}8{,}192$): two
cross-architecture Variant~A and five same-model Variant~B
(Appendix~\ref{app:mad_combos}). The best debate (Qwen3-4B Variant~B, 87.67\%)
trails the best individual SLM judge by 2.49 points and trails its own constituent
(Qwen3-4B at $B{=}8{,}192$, 88.44\%) by 0.77 points. Both Variant~A configurations
underperform the individual accuracy of their strongest member, confirming that
architectural diversity does not offset the persuasion vulnerability analyzed in
Appendix~\ref{app:error_analysis}.

\begin{table*}[ht]
\centering
\small
\begin{adjustbox}{max width=\textwidth}
\rowcolors{2}{white}{slmrow}
\setlength{\tabcolsep}{4pt}
\begin{tabular}{llllcc}
\toprule
\textbf{Variant} & \textbf{Agent 1} & \textbf{Agent 2} & \textbf{Agent 3} & \textbf{Acc (\%)} & \textbf{IFR (\%)} \\
\midrule
    & Qwen3-4B ($\tau{=}0$) & Qwen2.5-3B ($\tau{=}0$) & Qwen3-1.7B ($\tau{=}0$) & 85.12 & \phantom{0}99.98 \\
\multirow{-2}{*}{A}
    & Llama-3.2-3B ($\tau{=}0$) & Qwen2.5-1.5B ($\tau{=}0$) & Qwen3-1.7B ($\tau{=}0$) & 84.50 & \phantom{0}99.88 \\
\midrule
    & \multicolumn{3}{l}{Qwen3-4B $\times$3 ($\tau{=}0/0.4/0.9$)} & \textbf{87.67} & 100.00 \\
    & \multicolumn{3}{l}{Phi-4-mini $\times$3 ($\tau{=}0/0.4/0.9$)} & 87.44 & \phantom{0}99.83 \\
    & \multicolumn{3}{l}{Qwen2.5-1.5B $\times$3 ($\tau{=}0/0.4/0.9$)} & 78.43 & \phantom{0}99.66 \\
    & \multicolumn{3}{l}{Qwen3-0.6B $\times$3 ($\tau{=}0/0.4/0.9$)} & 75.95 & \phantom{0}98.40 \\
\multirow{-5}{*}{B}
    & \multicolumn{3}{l}{Llama-3.2-1B $\times$3 ($\tau{=}0/0.4/0.9$)} & 63.04 & \phantom{0}98.31 \\
\bottomrule
\end{tabular}
\end{adjustbox}
\caption{\textbf{Full debate results ($B{=}8{,}192$).} All 7 debate configurations that ran at the standard token budget. The best debate (Qwen3-4B Variant~B, 87.67\%) falls 2.49 points below the best individual SLM judge (90.16\%). Each Variant~A configuration underperforms the individual accuracy of its strongest constituent at $B{=}8{,}192$ (Qwen3-4B at 88.44\% in the first, Qwen3-1.7B at 86.56\% in the second).}
\label{tab:app_debate_full}
\end{table*}

\paragraph{Cross-architecture debates at a reduced budget.} Three cross-architecture configurations ran at $B{=}4{,}096$ rather than $B{=}8{,}192$ for most student--dataset files, because five rounds of critique and refinement over three agents exceed the context window at the larger budget; their Qwen2.5-32B GSM8K files, and in one configuration (Qwen3-4B + Phi-4-mini + Qwen2.5-3B) its GSM-Plus file, ran at $B{=}8{,}192$. Since budget and architecture would otherwise vary together, they form their own comparison group in Table~\ref{tab:app_debate_reduced} and are read only against each other. They reach 83.49--86.62\%, the same regime as the $B{=}8{,}192$ cross-architecture debates, and each falls below the individual accuracy of its strongest member, so the conclusion that debate does not beat a strong single judge holds at either budget.

\begin{table*}[ht]
\centering
\small
\begin{adjustbox}{max width=\textwidth}
\rowcolors{2}{white}{slmrow}
\begin{tabular}{llllcc}
\toprule
\textbf{Variant} & \textbf{Agent 1} & \textbf{Agent 2} & \textbf{Agent 3} & \textbf{Acc (\%)} & \textbf{IFR (\%)} \\
\midrule
    & Llama-3.2-3B ($\tau{=}0$) & Qwen3-4B ($\tau{=}0$) & Phi-4-mini ($\tau{=}0$) & 86.62 & 99.66 \\
    & Qwen2.5-1.5B ($\tau{=}0$) & Qwen3-1.7B ($\tau{=}0$) & Phi-4-mini ($\tau{=}0$) & 85.02 & 99.43 \\
\multirow{-3}{*}{A ($B{=}4{,}096$)}
    & Qwen3-4B ($\tau{=}0$) & Phi-4-mini ($\tau{=}0$) & Qwen2.5-3B ($\tau{=}0$) & 83.49 & 98.44 \\
\bottomrule
\end{tabular}
\end{adjustbox}
\caption{\textbf{Reduced-budget debate results ($B{=}4{,}096$).} Three Variant~A configurations where multi-round debate prompts exceeded the context window at $B{=}8{,}192$. These ran at $B{=}4{,}096$ for most files (see text) and are reported separately, so that the debate comparisons of the main paper use one token budget.}
\label{tab:app_debate_reduced}
\end{table*}

\section{Collaborative Judging: Algorithms and Prompts}
\label{app:debate}

This section formalizes the two collaborative judging strategies of
\S\ref{sec:strategies}: majority voting, whose aggregation rule is given in
Appendix~\ref{app:ensemble_formal}, and the Reflect-Critique-Refine (RCR)
multi-agent debate protocol. Debate runs one of three prompts according to the
dataset domain, for math (Appendix~\ref{app:rcr_math}), science
(Appendix~\ref{app:rcr_science}), and general reasoning
(Appendix~\ref{app:rcr_general}). The corresponding result tables are reported
in Appendix~\ref{app:ensemble}.

\subsection{Majority-Voting Ensemble}
\label{app:ensemble_formal}

Given three judges producing verdicts $(v_1, v_2, v_3) \in \mathcal{V}^3$ for a
problem, the ensemble verdict is the majority label, with ties resolved to
\texttt{Undefined}:
\begin{equation}
V_{\text{maj}}(v_1, v_2, v_3) =
\begin{cases}
    \texttt{C} & \text{if } \sum_{j=1}^{3} \mathbf{1}[v_j {=} \texttt{C}] \geq 2 \\[3pt]
    \texttt{I} & \text{if } \sum_{j=1}^{3} \mathbf{1}[v_j {=} \texttt{I}] \geq 2 \\[3pt]
    \texttt{U} & \text{otherwise}
\end{cases}
\label{eq:majority}
\end{equation}
We draw the panel from the top-5 individual judges (Phi-4@10, Qwen3-14B@10,
Qwen3-8B@10, Phi-4-R-Plus@8192, Phi-4-R@8192), each at its best token setting,
and enumerate all $\binom{5}{3}{=}10$ panels. Because the ensemble reuses
cached individual judgments, it requires no additional inference.

\subsection{RCR Multi-Agent Debate}

Algorithm~\ref{alg:mad} formalizes the RCR procedure. Round~0 uses the standard
reasoned-verdict prompt (Appendix~\ref{app:prompt_reasoned}); each subsequent
round shows every agent its own previous judgment and those of its peers, then
asks it to reflect, critique, and refine. Debate stops as soon as the three
agents agree (consensus) or after $T{=}5$ rounds, in which case the verdict
falls back to the majority label of Eq.~\ref{eq:majority}.

\begin{algorithm}[ht]
\caption{RCR Multi-Agent Debate for Binary Judging}
\label{alg:mad}
\raggedright
\begin{algorithmic}[1]
\REQUIRE Query $q$, agents $\mathcal{A} = \{a_1, \ldots, a_N\}$, max rounds $T$
\ENSURE Consensus verdict $v^*$ and reasoning traces $\mathcal{R}$
\STATE \textbf{Round 0:} Each $a_i$ generates $(v_i^{(0)}, r_i^{(0)}) \sim \pi_{a_i}(\cdot \mid q)$
\IF{$\forall\, i,j: v_i^{(0)} = v_j^{(0)}$}
    \RETURN $(v_i^{(0)},\, \{r_i^{(0)}\}_{i=1}^N)$
\ENDIF
\FOR{$t = 1$ \TO $T$}
    \FORALL{$a_i \in \mathcal{A}$}
        \STATE $\mathcal{P}_i^{(t-1)} \leftarrow \{(v_j^{(t-1)}, r_j^{(t-1)})\}_{j \neq i}$
        \STATE \textbf{Reflect:} Self-critique $c_i^{\text{self}}$ on $r_i^{(t-1)}$
        \STATE \textbf{Critique:} Evaluate $\mathcal{P}_i^{(t-1)}$
        \STATE \textbf{Refine:} Update $(v_i^{(t)}, r_i^{(t)})$ from $v_i^{(t-1)}$, $c_i$, $\mathcal{P}_i$
    \ENDFOR
    \IF{$\forall\, i,j: v_i^{(t)} = v_j^{(t)}$}
        \RETURN $(v_i^{(t)},\, \bigcup_{i,s \leq t} r_i^{(s)})$
    \ENDIF
\ENDFOR
\RETURN $(\text{majority}(\{v_i^{(T)}\}),\, \bigcup_{i,t} r_i^{(t)})$
\end{algorithmic}
\end{algorithm}

\subsection{Debate Configurations}
\label{app:mad_combos}

Table~\ref{tab:app_mad_combos} lists all 10 debate configurations. Variant~A
uses three different models at $\tau{=}0$; Variant~B uses one model
sampled at three temperatures $\tau \in \{0.0, 0.4, 0.9\}$ to inject diversity
without architectural change. Three cross-architecture configurations (marked with $^\dagger$) run at $B{=}4{,}096$, because five rounds over three agents exceed the context window at $B{=}8{,}192$; they form their own comparison group in Table~\ref{tab:app_debate_reduced}.

\begin{table*}[ht]
\centering
\small
\begin{adjustbox}{max width=\textwidth}
\rowcolors{2}{white}{slmrow}
\begin{tabular}{llccc}
\toprule
\textbf{Variant} & \textbf{Agent Models} & \textbf{$t_1$} & \textbf{$t_2$} & \textbf{$t_3$} \\
\midrule
& Llama-3.2-3B, Qwen2.5-1.5B, Qwen3-1.7B & 0.0 & 0.0 & 0.0 \\
& Qwen2.5-1.5B, Qwen3-1.7B, Phi-4-mini$^\dagger$ & 0.0 & 0.0 & 0.0 \\
& Qwen3-4B, Phi-4-mini, Qwen2.5-3B$^\dagger$ & 0.0 & 0.0 & 0.0 \\
& Qwen3-4B, Qwen2.5-3B, Qwen3-1.7B & 0.0 & 0.0 & 0.0 \\
\multirow{-5}{*}{A}
& Llama-3.2-3B, Qwen3-4B, Phi-4-mini$^\dagger$ & 0.0 & 0.0 & 0.0 \\
\midrule
& Llama-3.2-1B $\times$ 3 & 0.0 & 0.4 & 0.9 \\
& Qwen2.5-1.5B $\times$ 3 & 0.0 & 0.4 & 0.9 \\
& Qwen3-4B $\times$ 3 & 0.0 & 0.4 & 0.9 \\
& Qwen3-0.6B $\times$ 3 & 0.0 & 0.4 & 0.9 \\
\multirow{-5}{*}{B}
& Phi-4-mini $\times$ 3 & 0.0 & 0.4 & 0.9 \\
\bottomrule
\end{tabular}
\end{adjustbox}
\caption{\textbf{Debate configurations.} Variant~A (cross-architecture):
three different models at temperature 0.0. Variant~B (same-model): the same
model at three temperatures. Seven configurations ran at \texttt{max\_tokens=8192}; three ($^\dagger$) ran at \texttt{max\_tokens=4096} for most files due to context length constraints (see Table~\ref{tab:app_debate_reduced} and Appendix~\ref{app:ensemble}). All debates run for up to 6 rounds (Round~0 initial $+$ Rounds~1--5 RCR).}
\label{tab:app_mad_combos}
\end{table*}

\paragraph{Subprocess isolation.} Each agent in a Variant~A debate uses a
different model. To respect GPU memory limits, agents are loaded in separate
subprocesses via Python \texttt{multiprocessing} (spawn context): each
subprocess loads one vLLM instance, runs inference for all problems in the
current round, returns them through a temporary file, and terminates before the next
agent loads. This lets three distinct models debate without requiring
simultaneous GPU residency.

\subsection{RCR Prompt: Math Datasets}
\label{app:rcr_math}

Used for GSM8K, GSM-Plus, and MATH.

\begin{promptflow}{RCR Debate Prompt: Math}
\begin{verbatim}
You are Agent {agent_id} in a multi-agent debate
to judge whether a student's math answer is
correct.

Question: {question}

Ground truth answer: {ground_truth}

Student answer: {student_answer}

{own_previous}

Here are the judgements from other agents:
{context}

This is debate round {round_num}. Please
carefully analyze all judgements—including your
own—identify any errors in reasoning, and
provide your revised judgement.
- If you believe your previous verdict is correct,
  explain why and defend it.
- If you believe you made an error, explain the
  error and provide a corrected judgement.
- If you believe another agent's verdict is
  correct, explain why you agree with it.

Your final verdict must be:
\boxed{CORRECT} or \boxed{INCORRECT}
\end{verbatim}
\end{promptflow}

\subsection{RCR Prompt: Science Datasets}
\label{app:rcr_science}

Used for ARC-Easy and ARC-Challenge.

\begin{promptflow}{RCR Debate Prompt: Science}
\begin{verbatim}
You are Agent {agent_id} in a multi-agent debate
to judge whether a student's science answer is
correct.

Question: {question}

Ground truth answer: {ground_truth}

Student answer: {student_answer}

{own_previous}

Here are the judgements from other agents:
{context}

This is debate round {round_num}. Please
carefully analyze all judgements—including your
own—identify any misconceptions or flawed
scientific reasoning, and provide your revised
judgement.
- If you believe your previous verdict is correct,
  explain the scientific principles supporting
  your verdict.
- If you believe you made an error, explain the
  scientific misconception and provide a corrected
  judgement.
- If you believe another agent's verdict is
  correct, explain why their scientific reasoning
  is sound.

Your final verdict must be:
\boxed{CORRECT} or \boxed{INCORRECT}
\end{verbatim}
\end{promptflow}

\subsection{RCR Prompt: General Reasoning Datasets}
\label{app:rcr_general}

Used for HellaSwag, WinoGrande, and TruthfulQA. This prompt is domain-neutral
and focuses on general reasoning quality.

\begin{promptflow}{RCR Debate Prompt: General}
\begin{verbatim}
You are Agent {agent_id} in a multi-agent debate
to judge whether a student's answer is correct.

Question: {question}

Ground truth answer: {ground_truth}

Student answer: {student_answer}

{own_previous}

Here are the judgements from other agents:
{context}

This is debate round {round_num}. Please
carefully analyze all judgements—including your
own—identify any errors in reasoning, and
provide your revised judgement.
- If you believe your previous verdict is correct,
  explain clearly why and defend your position.
- If you believe you made an error, explain what
  went wrong and provide a corrected judgement.
- If you believe another agent's verdict is
  correct, explain why you agree with their
  reasoning.

Your final verdict must be:
\boxed{CORRECT} or \boxed{INCORRECT}
\end{verbatim}
\end{promptflow}

\section[Open-Ended Scoring Metrics]{Open-Ended Scoring Metrics and Parsing}
\label{app:scoring_metrics}

This section formalizes the five correlation metrics used for open-ended
evaluation (\S\ref{sec:openended_method}) and the score-parsing cascade that
turns a free-form scoring response into a number. All metrics are implemented in
the released framework's scoring evaluator and score parser.

\subsection{Score Vectors}

For a benchmark of $n$ items, let $\mathbf{s} = (s_1, \dots, s_n)$ be the SLM
judge scores and $\mathbf{g} = (g_1, \dots, g_n)$ the reference scores (human
annotations for SummEval, reference-model scores for MT-Bench), with each $s_i \in
\{1, \dots, 5\}$. MT-Bench reference scores are integers on the same scale;
each SummEval reference is the mean of three expert ratings, so it can fall
between integers. Items where the judge response fails to parse are dropped
pairwise, leaving $n_{\text{valid}} \le n$ pairs over which the metrics below are
computed.

\subsection{Correlation and Agreement Metrics}

\paragraph{Pearson $r$ (linear agreement).}
\begin{equation}
r(\mathbf{s}, \mathbf{g}) =
\frac{\sum_{i} (s_i - \bar{s})(g_i - \bar{g})}
{\sqrt{\sum_{i} (s_i - \bar{s})^2}\;\sqrt{\sum_{i} (g_i - \bar{g})^2}},
\end{equation}
undefined (reported as $\varnothing$) when either score vector has zero
variance.

\paragraph{Spearman $\rho$ (ordinal agreement).} Let $\textsc{rk}(\cdot)$ assign
fractional (tie-averaged) ranks. Then $\rho = r\bigl(\textsc{rk}(\mathbf{s}),
\textsc{rk}(\mathbf{g})\bigr)$, i.e.\ Pearson correlation on ranks, which is
robust to monotone nonlinearity in the 1--5 scale.

\paragraph{Cohen's $\kappa$ (binary agreement).} Binarize at the quality
threshold $t = 4$ via $b_i = \mathbf{1}[s_i \ge t]$ and $b_i' = \mathbf{1}[g_i
\ge t]$. With observed agreement $p_o$ and chance agreement $p_e$ from the
$2{\times}2$ confusion matrix,
\begin{equation}
\kappa = \frac{p_o - p_e}{1 - p_e}.
\end{equation}

\paragraph{Thresholded accuracy.} The fraction of items on which the binarized
labels agree, $\text{Acc} = \tfrac{1}{n}\sum_i \mathbf{1}[b_i = b_i']$.

\paragraph{Mean squared error (calibration).} $\text{MSE} =
\tfrac{1}{n}\sum_i (s_i - g_i)^2$, capturing absolute scale calibration rather
than rank or linear agreement.

For SummEval, these metrics are computed per dimension (coherence, consistency,
fluency, relevance) and then averaged; for MT-Bench they are computed over the
holistic per-conversation score against the reference model
(Qwen3.5-397B-A17B), once per student, and the two values are averaged. The 95\%
intervals on $\rho$ resample the 100 SummEval source articles and the 80
MT-Bench questions (Appendix~\ref{app:paired_stats}).

\subsection{Score Parsing}

Extracting a numeric score from a judge response is itself a prioritized
cascade, because models express scores in many surface forms. Algorithm
\ref{alg:scoreparse} shows the single-score parser; the SummEval multi-score
parser additionally supports structured (\texttt{coherence=4, \dots}),
positional (\texttt{3,2,2,3}), and per-dimension scattered formats, with
\texttt{\textbackslash text\{\}} wrappers and \texttt{\$\dots\$} math mode
stripped before matching.

\begin{algorithm}[ht]
\caption{Single-Score Parsing (1--5)}
\label{alg:scoreparse}
\begin{algorithmic}[1]
\REQUIRE Response $y$, max score $M = 5$
\ENSURE Score $s \in \{1, \dots, M\}$ or \texttt{None}
\STATE \textbf{(1)} last \texttt{\textbackslash boxed\{N\}} or \texttt{\textbackslash boxed\{N/M\}}
\STATE \textbf{(2)} last markdown \texttt{**N**}
\STATE \textbf{(3)} last bracketed \texttt{[[N]]}
\STATE \textbf{(4)} labeled ``score/rating: N'' or ``N out of M''
\STATE \textbf{(5)} last standalone integer in $[1, M]$
\STATE \textbf{return} first valid match in range, else \texttt{None}
\end{algorithmic}
\end{algorithm}

A score of \texttt{None} is excluded from the valid pairs above and counted in
the parse-failure rate; the strong judges parse cleanly, while sub-1B judges
account for most failures.

\section{Per-Dataset Detailed Results}
\label{app:per_dataset}

\begin{figure*}[ht]
\centering
\includegraphics[width=\textwidth]{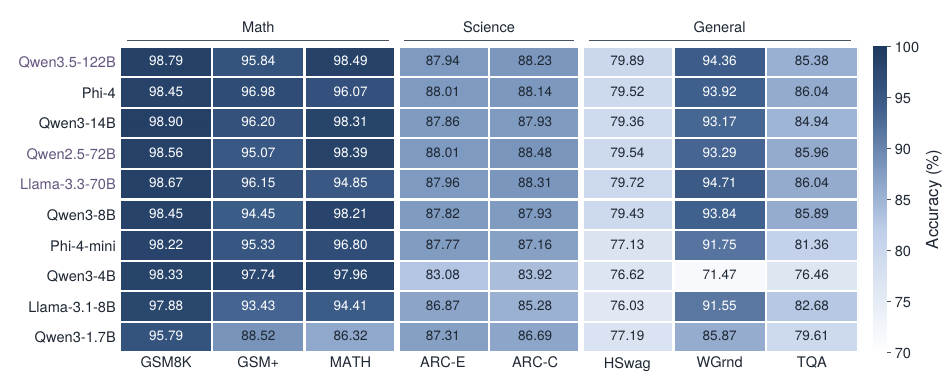}
\caption{\textbf{Per-dataset judge accuracy at $B{=}10$} for the ten most accurate judges at that budget, small and large, other than the ladder judge Qwen3-32B, read as a grid (small judges from Table~\ref{tab:app_per_dataset_10}, large judges from Appendix~\ref{app:large_judges}). Math benchmarks cluster high, while general-reasoning tasks expose the cross-family gaps discussed in \S\ref{sec:individual}.}
\label{fig:per_dataset_heatmap}
\end{figure*}

\begin{figure*}[ht]
\centering
\includegraphics[width=\textwidth]{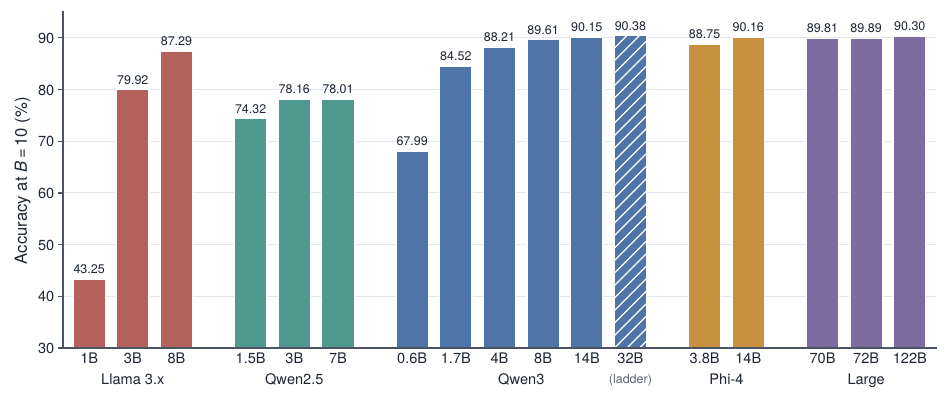}
\caption{\textbf{Judge accuracy by model family and size} at 10 tokens. Within each family, accuracy generally scales with parameters, but domain generalization varies significantly across families. Phi-4 and Qwen3 maintain the most uniform accuracy across all eight benchmarks. The hatched bar is Qwen3-32B, the ladder judge; the last group holds the three large judges that support a 10-token verdict (Llama-3.3-70B, Qwen2.5-72B, and Qwen3.5-122B-A10B).}
\label{fig:scaling_curve}
\end{figure*}

\begin{figure}[!htbp]
\centering
\includegraphics[width=\columnwidth]{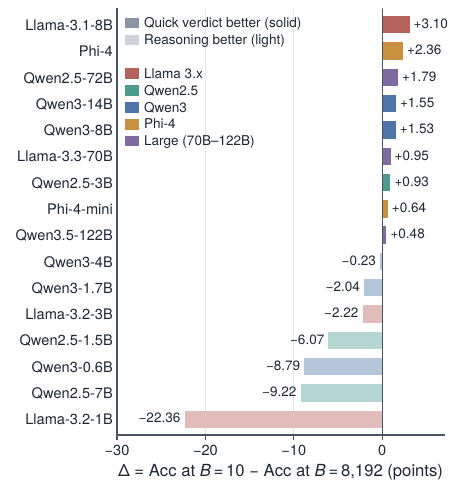}
\caption{\textbf{Overthinking delta per judge} for 16 dual-setting judges (the 13 small ones and three large), $\Delta = \text{Acc}(B{=}10) - \text{Acc}(B{=}8{,}192)$, over all eight datasets. Dark bars mark judges for which the quick verdict wins, light bars those for which reasoning wins. The aggregate $\Delta$ hides the domain reversal of \S\ref{sec:individual}: the same judge can sit on either side once math and general benchmarks are separated.}
\label{fig:overthinking_delta}
\end{figure}

Tables~\ref{tab:app_per_dataset_10} and~\ref{tab:app_per_dataset_8192} present accuracy broken down by individual dataset for all 16 small judges at each token setting (the large judges' per-dataset accuracy is in Appendix~\ref{app:large_judges}), Figure~\ref{fig:token_budget_bars} contrasts the two budgets per judge, and Figure~\ref{fig:overthinking_delta} shows the resulting per-judge $\Delta$.
These results complement the aggregate accuracy in Table~\ref{tab:individual} (main paper), which averages over all datasets and both student models. Figure~\ref{fig:per_dataset_heatmap} reads the grid for the ten most accurate judges as a heatmap, and Figure~\ref{fig:scaling_curve} plots accuracy against judge size within each family.

\begin{table*}[t]
\centering
\small
\begin{adjustbox}{max width=\textwidth}
\rowcolors{2}{white}{slmrow}
\setlength{\tabcolsep}{4pt}
\begin{tabular}{llccccccccc}
\toprule
\hiderowcolors
& & \multicolumn{3}{c}{\textbf{Math}} & \multicolumn{2}{c}{\textbf{Science}} & \multicolumn{3}{c}{\textbf{General}} & \\
\cmidrule(lr){3-5} \cmidrule(lr){6-7} \cmidrule(lr){8-10}
\textbf{Family} & \textbf{Judge Model} & \textbf{GSM8K} & \textbf{GSM+} & \textbf{MATH} & \textbf{ARC-E} & \textbf{ARC-C} & \textbf{HSwag} & \textbf{WGrnd} & \textbf{TQA} & \textbf{Avg} \\
\midrule
\noalign{\global\rownum=2}\showrowcolors
    & Llama-3.2-1B   & 31.58 & 39.61 & 43.41 & 48.02 & 45.86 & 46.19 & 48.62 & 46.71 & 43.25 \\
    & Llama-3.2-3B   & 94.43 & 84.03 & 76.49 & 85.92 & 83.66 & 72.62 & 88.36 & 77.78 & 79.92 \\
\multirow{-3}{*}{Llama 3.x}
    & Llama-3.1-8B   & 97.88 & 93.43 & 94.41 & 86.87 & 85.28 & 76.03 & 91.55 & 82.68 & 87.29 \\
\midrule
    & Qwen2.5-1.5B   & 73.62 & 76.95 & 86.97 & 78.43 & 72.14 & 67.26 & 63.50 & 55.63 & 74.32 \\
    & Qwen2.5-3B     & 96.06 & 96.01 & 94.57 & 69.00 & 68.69 & 55.75 & 58.52 & 61.55 & 78.16 \\
\multirow{-3}{*}{Qwen2.5}
    & Qwen2.5-7B     & 97.50 & 97.16 & 96.74 & 60.84 & 59.98 & 55.35 & 64.40 & 56.36 & 78.01 \\
\midrule
    & Qwen3-0.6B     & 89.20 & 72.07 & 65.97 & 85.06 & 80.50 & 56.48 & 72.77 & 58.55 & 67.99 \\
    & Qwen3-1.7B     & 95.79 & 88.52 & 86.32 & 87.31 & 86.69 & 77.19 & 85.87 & 79.61 & 84.52 \\
    & Qwen3-4B       & 98.33 & 97.74 & 97.96 & 83.08 & 83.92 & 76.62 & 71.47 & 76.46 & 88.21 \\
    & Qwen3-8B       & 98.45 & 94.45 & 98.21 & 87.82 & 87.93 & 79.43 & 93.84 & 85.89 & 89.61 \\
\multirow{-5}{*}{Qwen3}
    & Qwen3-14B      & 98.90 & 96.20 & 98.31 & 87.86 & 87.93 & 79.36 & 93.17 & 84.94 & 90.15 \\
\midrule
    & Phi-4          & 98.45 & 96.98 & 96.07 & 88.01 & 88.14 & 79.52 & 93.92 & 86.04 & 90.16 \\
\multirow{-2}{*}{Phi-4}
    & Phi-4-mini     & 98.22 & 95.33 & 96.80 & 87.77 & 87.16 & 77.13 & 91.75 & 81.36 & 88.75 \\
\bottomrule
\end{tabular}
\end{adjustbox}
\caption{\textbf{Per-dataset accuracy (\%) at 10 tokens} (quick verdict). Avg is micro-averaged over all 8 datasets and both students, so it weights each dataset by its size and matches Table~\ref{tab:individual}. The 13 dual-setting small judges are shown; always-thinking judges have no 10-token configuration.}
\label{tab:app_per_dataset_10}
\end{table*}

\begin{table*}[t]
\centering
\small
\begin{adjustbox}{max width=\textwidth}
\rowcolors{2}{white}{slmrow}
\setlength{\tabcolsep}{4pt}
\begin{tabular}{llccccccccc}
\toprule
\hiderowcolors
& & \multicolumn{3}{c}{\textbf{Math}} & \multicolumn{2}{c}{\textbf{Science}} & \multicolumn{3}{c}{\textbf{General}} & \\
\cmidrule(lr){3-5} \cmidrule(lr){6-7} \cmidrule(lr){8-10}
\textbf{Family} & \textbf{Judge Model} & \textbf{GSM8K} & \textbf{GSM+} & \textbf{MATH} & \textbf{ARC-E} & \textbf{ARC-C} & \textbf{HSwag} & \textbf{WGrnd} & \textbf{TQA} & \textbf{Avg} \\
\midrule
\noalign{\global\rownum=2}\showrowcolors
    & Llama-3.2-1B   & 74.91 & 67.47 & 59.10 & 80.47 & 77.69 & 60.60 & 70.52 & 58.99 & 65.61 \\
    & Llama-3.2-3B   & 91.51 & 89.10 & 86.74 & 84.70 & 81.66 & 70.82 & 83.19 & 79.02 & 82.14 \\
\multirow{-3}{*}{Llama 3.x}
    & Llama-3.1-8B   & 96.82 & 89.67 & 92.68 & 85.77 & 83.79 & 73.81 & 75.49 & 77.27 & 84.20 \\
\midrule
    & Qwen2.5-1.5B   & 90.11 & 88.50 & 92.20 & 85.54 & 81.61 & 64.78 & 78.30 & 63.01 & 80.39 \\
    & Qwen2.5-3B     & 92.91 & 85.50 & 81.66 & 79.95 & 78.46 & 65.53 & 65.71 & 68.27 & 77.22 \\
\multirow{-3}{*}{Qwen2.5}
    & Qwen2.5-7B     & 97.95 & 94.49 & 94.80 & 85.94 & 84.85 & 75.69 & 85.44 & 80.34 & 87.23 \\
\bottomrule
\end{tabular}
\end{adjustbox}
\caption{\textbf{Per-dataset accuracy (\%) at 8,192 tokens} (reasoned verdict). Avg is micro-averaged as in Table~\ref{tab:app_per_dataset_10}. $^\dagger$ = always-thinking judges; Phi-4-R, Phi-4-R-Plus, and Phi-4-mini-R abbreviate the three Reasoning variants. All 16 small judges included.}
\label{tab:app_per_dataset_8192}
\end{table*}

\paragraph{Key Observations.}
\textbf{GSM8K is the easiest benchmark}: most models above 3B exceed 95\% at either token setting, and at 10 tokens the top five small judges span only 0.68 points (98.22--98.90\%), with the sixth, Llama-3.1-8B (97.88\%), 1.02 points behind Qwen3-14B, a clear ceiling effect. \textbf{HellaSwag is the hardest}: even the best judges reach only 79.52\% (Phi-4) and 79.43\% (Qwen3-8B) at 10 tokens, far below their 94--98\% on math, because its sentence-completion format needs commonsense that brief verdicts cannot capture. \textbf{WinoGrande shows the largest overthinking swing}: Qwen3-4B gains 23.09 points from reasoning (71.47\% $\to$ 94.55\%), since pronoun resolution demands step-by-step disambiguation; larger single-dataset gains come only from the weakest quick verdicts (Llama-3.2-1B, up to 43.33 points on GSM8K, and Qwen2.5-7B, up to 25.11 points on ARC-Easy). \textbf{Qwen2.5 models are extreme math specialists}: Qwen2.5-7B scores 97.50\% on GSM8K but only 55.35\% on HellaSwag (a 42.15-point gap), and Qwen2.5-3B follows the same pattern (96.06\% vs.\ 55.75\%), reflecting heavily math-focused instruction tuning. \textbf{Reasoning helps on general and science tasks}: Llama-3.2-1B gains 32.45 points on ARC-Easy (48.02\% $\to$ 80.47\%) and 21.90 points on WinoGrande (48.62\% $\to$ 70.52\%), and for capable models like Qwen3-4B general tasks gain 3--23 points while math tasks lose 0.4--6 points. \textbf{Math overthinking is real}: Qwen2.5-3B drops 10.51 points on GSM-Plus (96.01\% $\to$ 85.50\%) and 12.91 points on MATH (94.57\% $\to$ 81.66\%) under reasoning, where quick judgment was already reliable.

\paragraph{Cross-Student Analysis.} An important dimension not visible in the per-dataset averages is how judge accuracy varies across student models. Qwen2.5-32B produces higher-quality solutions (95.45\% accuracy on GSM8K) than Llama-3.1-8B (82.49\%). Judges are more accurate on the stronger student's solutions for 11 of the 13 dual-setting judges at 10 tokens and for all 13 at 8,192, because correct solutions are easier to verify than incorrect ones. For the capable judges the gap is about five points at both budgets (Qwen3-14B 5.77 and 4.74, Phi-4 5.15 and 5.29), so it reflects the students more than the judge. The weakest judges are less consistent: Llama-3.2-1B moves from 1.23 points at 10 tokens to 11.33 at 8,192, Qwen3-0.6B from 17.73 to 5.91, and Qwen2.5-7B and Qwen2.5-3B are slightly more accurate on the weaker student at 10 tokens ($-$1.82 and $-$0.42). The full judge$\times$student matrices are in Appendix~\ref{app:self_preference}.

\begin{figure*}[ht]
\centering
\includegraphics[width=\textwidth]{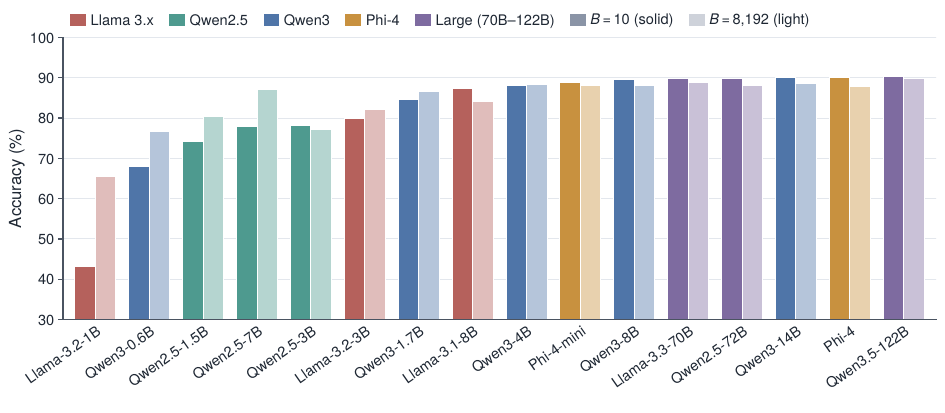}
\caption{\textbf{Token budget comparison} for 16 dual-setting judges (the 13 small ones and three large), showing absolute accuracy at both $B{=}10$ and $B{=}8{,}192$. The mixed pattern of which setting wins reflects opposing domain effects: quick verdicts dominate on math while reasoning dominates on general tasks (see \S\ref{sec:individual}).}
\label{fig:token_budget_bars}
\end{figure*}

\subsection{Instruction Following}

\begin{figure*}[ht]
\centering
\includegraphics[width=\textwidth]{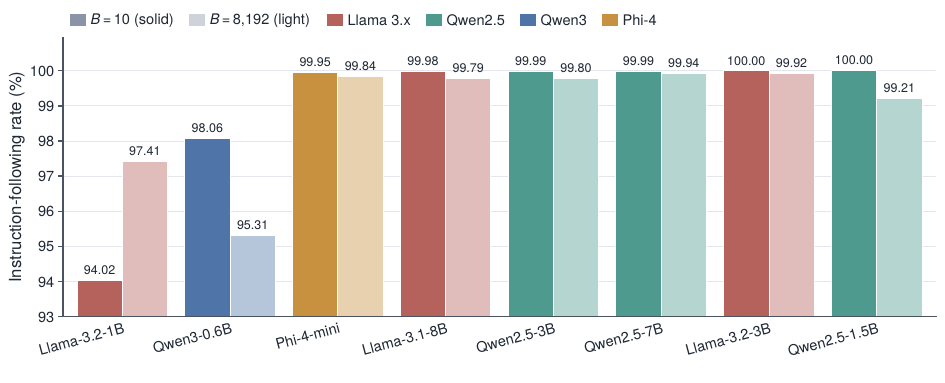}
\caption{\textbf{Instruction-following rate} at both budgets, for the dual-setting judges whose rate falls below 99.95\% at either setting. The judges not shown parse essentially every response; the always-thinking judges, which run at $B{=}8{,}192$ only, are reported in Table~\ref{tab:individual}.}
\label{fig:instruction_following}
\end{figure*}

Table~\ref{tab:individual} reports accuracy only. Figure~\ref{fig:instruction_following} gives the corresponding instruction-following rates, the fraction of responses from which a verdict could be parsed, for the dual-setting judges where the rate falls below 99.95\% at either budget. Every judge above 88\% accuracy stays above 99.8\% at both budgets, so the accuracy differences in the main text reflect genuine disagreement with the oracle rather than unparseable output. The judges that do lose parseable outputs are mostly the smaller ones, and most of them lose more at the reasoned budget, where a long explanation can run out before the verdict; Llama-3.2-1B is the exception, losing more at the quick verdict, where a single stray token is enough to make the response unreadable.

\section{Implementation Details}
\label{app:implementation}

This section documents the hardware, software stack, inference configuration,
and compute budget behind all experiments. The formal metric definitions
(accuracy, IFR, $\Delta$, persona range) are in Appendix~\ref{app:framework}.

\subsection{Hardware}

The core experiments (student generation, individual judging of the 16 small judges, personas, ensembles, debate, and open-ended scoring) were conducted on a shared NVIDIA DGX-1 server with 8 $\times$ Tesla V100-SXM2-32GB GPUs, 2 $\times$ Intel Xeon E5-2698 v4 CPUs @ 2.20GHz (40 cores, 80 threads), and 512GB DDR4 RAM, using up to 2 GPUs per model. Models requiring 1 GPU (0.6B--4B parameter models) run on a single V100.
Models requiring 2 GPUs (7B--14B models, student models) use tensor parallelism across 2 V100s.
The large judges, the ladder judge, and most of the extended analyses ran on cluster nodes with NVIDIA A40-, L40S-, A100-, and H200-class GPUs. There a large judge uses tensor parallelism across one to eight GPUs depending on the node's memory (Table~\ref{tab:app_large_models} lists the width on 44\,GiB GPUs), GLM-4.5-Air and Qwen3.5-122B-A10B are served from their vendors' FP8 checkpoints, and GPT-OSS-120B from its released MXFP4 checkpoint.
All judge inference runs locally with vLLM; MT-Bench reference scoring is the only exception and uses the Together API for GPT-OSS-120B and Qwen3.5-397B-A17B.

\subsection{Software Stack}

Table~\ref{tab:app_software} lists the exact software versions. The large judges and the extended analyses on the cluster nodes ran mainly under vLLM 0.24.0, a release that supports GPT-OSS-120B and the Qwen3.5 models.

\begin{table}[!htbp]
\centering
\small
\begin{adjustbox}{max width=\columnwidth}
\rowcolors{2}{white}{slmrow}
\begin{tabular}{ll}
\toprule
\textbf{Component} & \textbf{Version} \\
\midrule
vLLM & 0.8.5.post1 \\
Transformers & 4.57.1 \\
PyTorch & 2.6.0+cu124 \\
SymPy & 1.13.1 \\
antlr4 runtime & 4.11.1 \\
Python & 3.10.18 \\
CUDA & 12.4 \\
\bottomrule
\end{tabular}
\end{adjustbox}

\caption{\textbf{Software versions.}}
\label{tab:app_software}
\end{table}

\subsection{Inference Configuration}

Student generation and judge inference use vLLM \citep{kwon2023efficient} for high-throughput batched generation, and MT-Bench reference scoring with GPT-OSS-120B and Qwen3.5-397B-A17B runs through the Together API.
Key inference parameters:

\begingroup\raggedright
\begin{itemize}
    \item \textbf{Quick verdict}: \texttt{SamplingParams(max\_tokens=10,
        temperature=0)}
    \item \textbf{Reasoned verdict (non-thinking)}: \texttt{SamplingParams(\allowbreak max\_tokens=8192,\allowbreak{} temperature=0)}
    \item \textbf{Reasoned verdict (thinking-enabled)}: \texttt{SamplingParams(\allowbreak max\_tokens=8192,\allowbreak{} temperature=0.6,\allowbreak{} top\_p=0.95,\allowbreak{} top\_k=20,\allowbreak{} min\_p=0)}; GPT-OSS-120B, which reasons in its own output channels, uses the non-thinking setting (\texttt{temperature=0})
    \item \textbf{Chat template}: All prompts are tokenized using each model's native chat template via \texttt{tokenizer.apply\_chat\_template()}.
    \item \textbf{Thinking mode}: For Qwen3 models, Qwen3-32B, and Qwen3.5-122B-A10B, \texttt{enable\_thinking} is passed as a keyword argument to the chat template. When \texttt{max\_tokens=10}, thinking is disabled.
    \item \textbf{Batch size}: Determined by vLLM's continuous batching scheduler.
        Most models $\leq$4B use \texttt{max\_num\_seqs=64}; Phi-4-Reasoning models use \texttt{max\_num\_seqs=3} due to long output sequences; all others use the default (32). The large judges use 16 to 128 concurrent sequences, set per node so that each sequence keeps enough KV cache.
    \item \textbf{GPU memory}: For local vLLM runs, \texttt{gpu\_memory\_utilization} follows each model's serving configuration: student and Phi-4-family runs use 0.85, judge and scoring-judge defaults use 0.8, and student-generation defaults use 0.7. The large judges and Qwen3-32B use 0.90--0.95, depending on the node.
\end{itemize}
\endgroup

\subsection{Compute Budget}

Table~\ref{tab:app_compute} counts the inference behind the study. One stored result is one run: a judge's verdicts on one student and dataset at one budget, a persona condition, a debate transcript, a set of open-ended scores, or a student's solutions. Derived artifacts are excluded because they involve no inference, which is why majority voting contributes nothing: its panels are computed from verdicts already on disk. Counting one way throughout is what makes the total checkable, and it comes to \textbf{12{,}029} runs.

The upper block covers individual judging, personas, debate, and open-ended scoring. The lower block covers the large-judge anchors, the widened student panel, the prompt formulations, the factor decomposition with its budget sweep, and the alternative protocols.

\begin{table}[!htbp]
\centering
\rowcolors{2}{white}{slmrow}
\begin{adjustbox}{max width=\columnwidth}
\small
\begin{tabular}{lr}
\toprule
\textbf{Component} & \textbf{Runs} \\
\midrule
Individual judging & \phantom{00,}464 \\
Persona evaluation & \phantom{00,}960 \\
Multi-agent debate & \phantom{00,}160 \\
Open-ended judge scoring & \phantom{000,}80 \\
MT-Bench oracle scoring & \phantom{0000,}4 \\
MT-Bench student responses & \phantom{0000,}2 \\
Student solutions & \phantom{000,}16 \\
\midrule
Large-judge anchors & \phantom{00,}211 \\
Widened student panel & \phantom{0}1,204 \\
Prompt formulations & \phantom{0}3,808 \\
Factor decomposition and budget sweep & \phantom{0}3,296 \\
Alternative protocols & \phantom{0}1,824 \\
\midrule
\textbf{Total} & \textbf{12,029} \\
\bottomrule
\end{tabular}
\end{adjustbox}
\caption{\textbf{Inference runs behind the study}, counted as one stored result per run. Majority-voting panels reuse verdicts already computed and add no inference.}
\label{tab:app_compute}
\end{table}

\section{Prompt Sensitivity: Full Results}
\label{app:prompt_sensitivity}

\begin{figure*}[t]
\centering
\includegraphics[width=\textwidth]{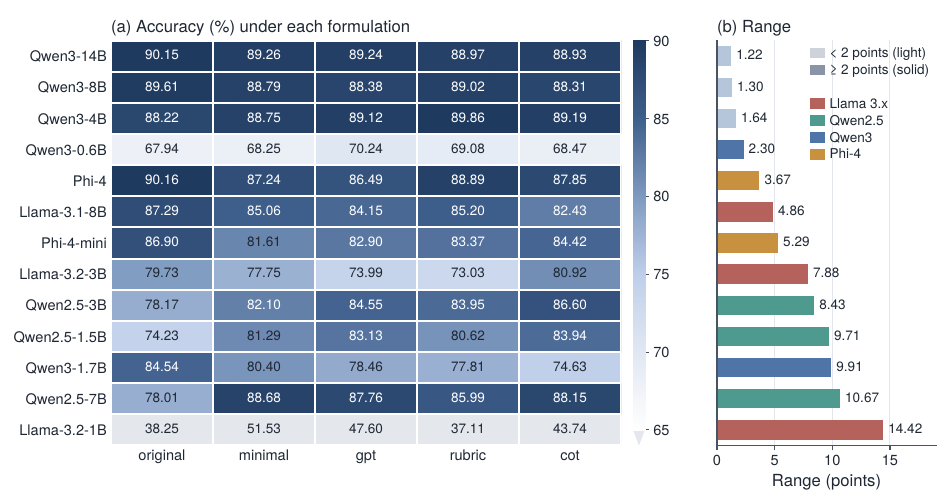}
\caption{\textbf{Prompt sensitivity at $B{=}10$.} Thirteen judges under five formulations of the judging instruction, every judge scored on all eight datasets and both students (64{,}824 judgments per cell). \textit{Left}: accuracy per formulation. \textit{Right}: each judge's range across the five, with lighter bars for ranges under 2 points. Table~\ref{tab:app_e3_t8192} gives the $B{=}8{,}192$ counterpart.}
\label{fig:prompt_sensitivity}
\end{figure*}

This appendix expands the prompt-formulation analysis of \S\ref{sec:robustness}. Every small judge runs under five formulations of the closed-ended judging instruction, the 13 dual-setting judges at both budgets and the three always-thinking Phi-4 judges at $B{=}8{,}192$:

\begin{itemize}
    \item \textbf{original}: the instruction used throughout the paper (Appendix~\ref{app:prompts});
    \item \textbf{minimal}: the same task stripped to a single terse sentence;
    \item \textbf{gpt\_style}: a phrasing modeled on MT-Bench's GPT-4 judge prompt (bracketed \texttt{[System]}, \texttt{[Task]}, and \texttt{[Verdict]} fields), to test whether instructions written for GPT-family judges disadvantage open-weight ones;
    \item \textbf{rubric\_based}: the criteria restated as an explicit checklist;
    \item \textbf{cot\_style}: an explicit request to reason step by step before the verdict.
\end{itemize}

Every judge is scored on the full grid under all five formulations, eight datasets and two students at 64{,}824 judgments per cell, so differences within and between rows measure prompt sensitivity alone. The \texttt{original} column is an independent run of the paper's instruction on different hardware. It reproduces Table~\ref{tab:individual} to within 0.3 points for every judge except Llama-3.2-1B, whose verdicts vary by several points between runs (Appendix~\ref{app:run_variance}), and Phi-4-mini, which scores 1.9 to 2.1 points lower here; each judge is compared with its own \texttt{original} run, so the ranges are unaffected. Tables~\ref{tab:app_e3_t10} and~\ref{tab:app_e3_t8192} give both budgets, and Figure~\ref{fig:prompt_sensitivity} shows the $B{=}10$ panel with each judge's range.

\begin{table}[!htbp]
\centering
\rowcolors{2}{white}{slmrow}
\begin{adjustbox}{max width=\columnwidth}
\small
\setlength{\tabcolsep}{4pt}
\begin{tabular}{lcccccc}
\toprule
\textbf{Judge} & \textbf{original} & \textbf{minimal} & \textbf{gpt} & \textbf{rubric} & \textbf{cot} & \textbf{Range} \\
\midrule
Qwen3-14B & 90.15 & 89.26 & 89.24 & 88.97 & 88.93 & \phantom{0}1.22 \\
Qwen3-8B & 89.61 & 88.79 & 88.38 & 89.02 & 88.31 & \phantom{0}1.30 \\
Qwen3-4B & 88.22 & 88.75 & 89.12 & 89.86 & 89.19 & \phantom{0}1.64 \\
Qwen3-0.6B & 67.94 & 68.25 & 70.24 & 69.08 & 68.47 & \phantom{0}2.30 \\
Phi-4 & 90.16 & 87.24 & 86.49 & 88.89 & 87.85 & \phantom{0}3.67 \\
Llama-3.1-8B & 87.29 & 85.06 & 84.15 & 85.20 & 82.43 & \phantom{0}4.86 \\
Phi-4-mini & 86.90 & 81.61 & 82.90 & 83.37 & 84.42 & \phantom{0}5.29 \\
Llama-3.2-3B & 79.73 & 77.75 & 73.99 & 73.03 & 80.92 & \phantom{0}7.88 \\
Qwen2.5-3B & 78.17 & 82.10 & 84.55 & 83.95 & 86.60 & \phantom{0}8.43 \\
Qwen2.5-1.5B & 74.23 & 81.29 & 83.13 & 80.62 & 83.94 & \phantom{0}9.71 \\
Qwen3-1.7B & 84.54 & 80.40 & 78.46 & 77.81 & 74.63 & \phantom{0}9.91 \\
Qwen2.5-7B & 78.01 & 88.68 & 87.76 & 85.99 & 88.15 & 10.67 \\
Llama-3.2-1B & 38.25 & 51.53 & 47.60 & 37.11 & 43.74 & 14.42 \\
\bottomrule
\end{tabular}
\end{adjustbox}
\caption{\textbf{Accuracy (\%) under each formulation at $B{=}10$}, sorted by range (in points). Columns gpt, rubric, and cot are \texttt{gpt\_style}, \texttt{rubric\_based}, and \texttt{cot\_style}.}
\label{tab:app_e3_t10}
\end{table}

\begin{table}[!htbp]
\centering
\rowcolors{2}{white}{slmrow}
\begin{adjustbox}{max width=\columnwidth}
\small
\setlength{\tabcolsep}{4pt}
\begin{tabular}{lcccccc}
\toprule
\textbf{Judge} & \textbf{original} & \textbf{minimal} & \textbf{gpt} & \textbf{rubric} & \textbf{cot} & \textbf{Range} \\
\midrule
Phi-4-Reasoning & 88.92 & 89.00 & 88.75 & 89.25 & 88.35 & \phantom{0}0.90 \\
Phi-4-Reasoning-Plus & 89.43 & 89.12 & 88.90 & 89.44 & 88.45 & \phantom{0}0.99 \\
Qwen3-14B & 88.57 & 88.54 & 88.55 & 88.97 & 87.44 & \phantom{0}1.52 \\
Qwen3-4B & 88.49 & 88.29 & 88.26 & 88.78 & 87.14 & \phantom{0}1.64 \\
Qwen3-8B & 88.12 & 88.34 & 87.76 & 88.88 & 86.94 & \phantom{0}1.94 \\
Qwen3-1.7B & 86.50 & 85.78 & 86.35 & 87.62 & 85.19 & \phantom{0}2.43 \\
Phi-4-mini-Reasoning & 83.91 & 82.93 & 85.42 & 84.80 & 84.26 & \phantom{0}2.49 \\
Llama-3.1-8B & 84.29 & 81.98 & 82.61 & 84.54 & 82.78 & \phantom{0}2.57 \\
Phi-4 & 87.80 & 85.74 & 85.87 & 88.47 & 85.83 & \phantom{0}2.73 \\
Qwen2.5-7B & 87.23 & 87.96 & 86.36 & 87.00 & 83.77 & \phantom{0}4.19 \\
Qwen3-0.6B & 76.73 & 77.08 & 81.29 & 81.33 & 80.46 & \phantom{0}4.60 \\
Phi-4-mini & 86.00 & 82.86 & 81.78 & 87.74 & 84.53 & \phantom{0}5.96 \\
Qwen2.5-1.5B & 80.32 & 70.88 & 73.58 & 71.77 & 76.32 & \phantom{0}9.44 \\
Qwen2.5-3B & 77.19 & 78.52 & 81.55 & 86.57 & 76.61 & \phantom{0}9.96 \\
Llama-3.2-3B & 82.42 & 77.76 & 77.99 & 72.26 & 81.25 & 10.16 \\
Llama-3.2-1B & 65.80 & 55.62 & 59.00 & 55.43 & 60.23 & 10.36 \\
\bottomrule
\end{tabular}
\end{adjustbox}
\caption{\textbf{Accuracy (\%) under each formulation at $B{=}8{,}192$}, sorted by range (in points), with columns as in Table~\ref{tab:app_e3_t10}.}
\label{tab:app_e3_t8192}
\end{table}

Four observations. First, the larger Qwen3 judges are the steadiest: at $B{=}10$ Qwen3-14B moves by 1.22 points across the five formulations, Qwen3-8B by 1.30, and Qwen3-4B by 1.64, against 3.67 for Phi-4, 10.67 for Qwen2.5-7B, and 14.42 for Llama-3.2-1B.

Second, a narrow range is not by itself a mark of quality. Qwen3-0.6B moves by only 2.30 points, yet averages 68.80\% where Phi-4 averages 88.13\% on the same items. It is stable because it is insensitive to the instruction, not because it reads the instruction well, and a judge that answers much the same way however it is asked looks identical on this axis to one that has understood the task. Stability and accuracy are separate properties, and a trustworthy judge needs both.

Third, the reasoning budget is itself a stabilizer for the two most fragile judges. Qwen2.5-7B's range shrinks from 10.67 to 4.19 once it can reason, and Llama-3.2-1B's from 14.42 to 10.36, which locates much of their $B{=}10$ fragility in surface parsing rather than in judging.

Fourth, the original instruction is not tuned toward any one group of judges. At $B{=}10$ it is the best of the five formulations for six judges (Phi-4, Phi-4-mini, Qwen3-14B, Qwen3-8B, Qwen3-1.7B, Llama-3.1-8B) and the worst for five (Qwen3-4B, Qwen3-0.6B, and all three Qwen2.5 judges). Qwen2.5-7B scores 78.01\% under it against 88.68\% under \texttt{minimal}, and Qwen2.5-3B 78.17\% against 86.60\% under \texttt{cot\_style}. For the Qwen2.5 judges the reported accuracies are therefore conservative, by up to 10.67 points (Qwen2.5-7B).

Rankings hold across formulations, and the reasoning budget makes them firmer. Mean pairwise Kendall's $\tau$ is 0.754 over the 13 judges at $B{=}10$ and 0.860 over the 16 at $B{=}8{,}192$ (0.831 on the same 13), with the weakest pair of formulations at 0.487 and 0.817 respectively, computed on the 16 cells every judge shares under every formulation. Both the mean and the minimum rise with the budget, and the minimum, which comes from the pair of formulations whose rankings agree least, rises further. The movement is concentrated in the prompt-sensitive judges: Qwen2.5-7B swings 10.67 points across the five formulations at $B{=}10$ against 4.19 at $B{=}8{,}192$, and falls from fourth place under \texttt{minimal}, \texttt{gpt\_style}, and \texttt{cot\_style} to tenth under the original one. Qwen3-14B stays in the top three under every formulation at both budgets, and at $B{=}8{,}192$ Phi-4-Reasoning and Phi-4-Reasoning-Plus move by under a point (0.90 and 0.99), the steadiest judges in the panel. Extra tokens alone do not make a judge more accurate (Appendix~\ref{app:budget_factorial}), but they do make its place in the ranking harder to move by rewording the instruction.

\subsection{Two Stress Variants}

\begin{figure*}[t]
\centering
\includegraphics[width=\textwidth]{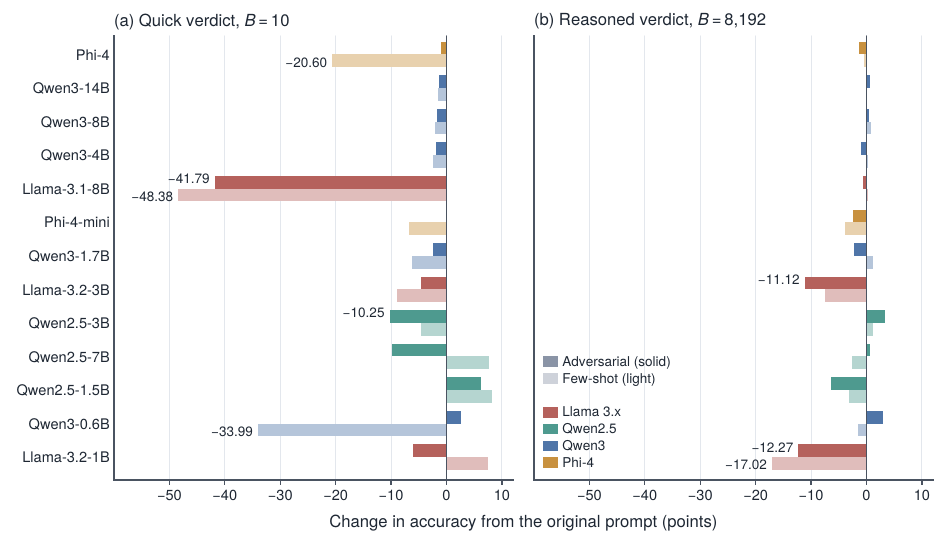}
\caption{\textbf{The two stress variants, against each judge's own baseline.} Thirteen judges, all eight datasets and both students per cell. The two panels share an axis: at $B{=}10$ an adversarial framing or a few-shot prompt can cost a judge 48 points, and at $B{=}8{,}192$ most judges are back within a few points of their baseline, with Llama-3.2-3B ($-$11.1 under the adversarial framing) and Llama-3.2-1B the clearest exceptions.}
\label{fig:stress_variants}
\end{figure*}

The five formulations above all ask for the same thing in different words. Two further variants change what is asked. \textbf{adversarial} tells the judge that the student's answer is very likely wrong (the reasoned version adds that the warning should not bias the verdict); \textbf{few\_shot} follows a short instruction with two worked examples, one correct answer and one wrong, before the item. Both were run for the 13 judges that support a $B{=}10$ configuration and, at $B{=}8{,}192$, for all 16 small judges, on the full 64{,}824-judgment grid, and each is scored against that judge's own \texttt{original} on the cells the three share. Tables~\ref{tab:app_e3_stress_t10} and~\ref{tab:app_e3_stress_t8192} give both budgets and Figure~\ref{fig:stress_variants} plots the changes.

\begin{table*}[t]
\centering
\rowcolors{2}{white}{slmrow}
\begin{adjustbox}{max width=\textwidth}
\small
\begin{tabular}{lcccccccc}
\toprule
\hiderowcolors
& \multicolumn{3}{c}{\textbf{Accuracy (\%)}} & \multicolumn{2}{c}{\textbf{Change}} & \multicolumn{3}{c}{\textbf{Unparseable (\%)}} \\
\cmidrule(lr){2-4} \cmidrule(lr){5-6} \cmidrule(lr){7-9}
\textbf{Judge} & \textbf{original} & \textbf{advers.} & \textbf{few-shot} & \textbf{advers.} & \textbf{few-shot} & \textbf{original} & \textbf{advers.} & \textbf{few-shot} \\
\midrule
\noalign{\global\rownum=2}\showrowcolors
Phi-4 & 90.16 & 89.22 & 69.57 & \phantom{0}$-$0.94 & $-$20.60 & 0.02 & 0.08 & \phantom{0}2.02 \\
Qwen3-14B & 90.15 & 88.82 & 88.60 & \phantom{0}$-$1.33 & \phantom{0}$-$1.55 & 0.00 & 0.00 & \phantom{0}0.27 \\
Qwen3-8B & 89.61 & 87.87 & 87.53 & \phantom{0}$-$1.74 & \phantom{0}$-$2.08 & 0.00 & 0.00 & \phantom{0}0.73 \\
Qwen3-4B & 88.22 & 86.34 & 85.74 & \phantom{0}$-$1.88 & \phantom{0}$-$2.48 & 0.00 & 0.00 & \phantom{0}1.24 \\
Llama-3.1-8B & 87.29 & 45.49 & 38.91 & $-$41.79 & $-$48.38 & 0.02 & 0.00 & 49.93 \\
Phi-4-mini & 86.90 & 86.76 & 80.21 & \phantom{0}$-$0.15 & \phantom{0}$-$6.69 & 0.22 & 0.00 & \phantom{0}0.56 \\
Qwen3-1.7B & 84.54 & 82.03 & 78.37 & \phantom{0}$-$2.51 & \phantom{0}$-$6.17 & 0.02 & 0.03 & \phantom{0}2.08 \\
Llama-3.2-3B & 79.73 & 75.08 & 70.77 & \phantom{0}$-$4.65 & \phantom{0}$-$8.95 & 0.00 & 0.00 & 10.21 \\
Qwen2.5-3B & 78.17 & 67.92 & 73.52 & $-$10.25 & \phantom{0}$-$4.65 & 0.01 & 0.01 & \phantom{0}7.32 \\
Qwen2.5-7B & 78.01 & 68.17 & 85.63 & \phantom{0}$-$9.84 & \phantom{0}$+$7.62 & 0.01 & 0.00 & \phantom{0}0.53 \\
Qwen2.5-1.5B & 74.23 & 80.51 & 82.42 & \phantom{0}$+$6.29 & \phantom{0}$+$8.19 & 0.00 & 0.00 & \phantom{0}0.16 \\
Qwen3-0.6B & 67.94 & 70.48 & 33.95 & \phantom{0}$+$2.54 & $-$33.99 & 1.97 & 0.00 & \phantom{0}0.09 \\
Llama-3.2-1B & 38.25 & 32.24 & 45.73 & \phantom{0}$-$6.01 & \phantom{0}$+$7.48 & 8.36 & 6.18 & 33.57 \\
\bottomrule
\end{tabular}
\end{adjustbox}
\caption{\textbf{Adversarial and few-shot variants at $B{=}10$}, against each judge's own \texttt{original} on the cells the three share, sorted by that baseline. Change is the variant's accuracy minus the \texttt{original} accuracy in the same row, in points. Unparseable is the share of judgments from which no verdict could be read.}
\label{tab:app_e3_stress_t10}
\end{table*}

\begin{table*}[t]
\centering
\rowcolors{2}{white}{slmrow}
\begin{adjustbox}{max width=\textwidth}
\small
\begin{tabular}{lcccccccc}
\toprule
\hiderowcolors
& \multicolumn{3}{c}{\textbf{Accuracy (\%)}} & \multicolumn{2}{c}{\textbf{Change}} & \multicolumn{3}{c}{\textbf{Unparseable (\%)}} \\
\cmidrule(lr){2-4} \cmidrule(lr){5-6} \cmidrule(lr){7-9}
\textbf{Judge} & \textbf{original} & \textbf{advers.} & \textbf{few-shot} & \textbf{advers.} & \textbf{few-shot} & \textbf{original} & \textbf{advers.} & \textbf{few-shot} \\
\midrule
\noalign{\global\rownum=2}\showrowcolors
Phi-4-Reasoning-Plus & 89.43 & 89.50 & 89.77 & \phantom{0}$+$0.07 & \phantom{0}$+$0.34 & 0.01 & \phantom{0}0.01 & \phantom{0}0.03 \\
Phi-4-Reasoning & 88.92 & 89.24 & 89.47 & \phantom{0}$+$0.31 & \phantom{0}$+$0.54 & 0.00 & \phantom{0}0.00 & \phantom{0}0.01 \\
Qwen3-14B & 88.57 & 89.20 & 88.56 & \phantom{0}$+$0.63 & \phantom{0}$-$0.01 & 0.00 & \phantom{0}0.01 & \phantom{0}0.00 \\
Qwen3-4B & 88.49 & 87.58 & 88.83 & \phantom{0}$-$0.91 & \phantom{0}$+$0.34 & 0.01 & \phantom{0}0.05 & \phantom{0}0.00 \\
Qwen3-8B & 88.12 & 88.49 & 88.85 & \phantom{0}$+$0.37 & \phantom{0}$+$0.73 & 0.00 & \phantom{0}0.01 & \phantom{0}0.01 \\
Phi-4 & 87.80 & 86.51 & 87.33 & \phantom{0}$-$1.29 & \phantom{0}$-$0.47 & 0.00 & \phantom{0}0.01 & \phantom{0}0.01 \\
\bottomrule
\end{tabular}
\end{adjustbox}
\caption{\textbf{Adversarial and few-shot variants at $B{=}8{,}192$}, as in Table~\ref{tab:app_e3_stress_t10}.}
\label{tab:app_e3_stress_t8192}
\end{table*}

The two break judges in different ways, and only the unparseable columns tell them apart. At $B{=}10$ few-shot costs Llama-3.1-8B 48.38 points and pushes it from 0.02\% unparseable to 49.93\%: the exemplars teach it to open with a worked rationale, which does not fit in ten tokens, so half its verdicts never arrive. Adversarial costs the same judge 41.79 points while leaving it at 0.00\% unparseable, so it emits a clean verdict that is simply wrong. One breaks the output format and the other breaks the judgment, and an aggregate accuracy drop cannot distinguish them.

The reasoning budget absorbs most of both. At $B{=}8{,}192$ the largest loss among the fourteen judges of 1.5B or more is 11.12 points (Llama-3.2-3B under the adversarial framing), Qwen3-14B, Qwen3-8B and Qwen3-4B all move by under a point in either direction, and Llama-3.1-8B, which lost 42 to 48 points at the quick budget, is within 0.60. Llama-3.2-1B is the exception that stays broken at both budgets, losing 12.27 to adversarial and 17.02 to few-shot with room to reason. For the format failures this is the mechanism the single-factor decomposition finds (\S\ref{sec:individual}), reached from a different direction: at ten tokens they depend on whether the judge can still emit a parseable verdict. The adversarial failures leave the verdict parseable, and the budget removes them as well.

On the open-ended side, prompt-engineering bias is controlled by construction: the reference scorers and the SLM judges use the identical scoring instruction (\S\ref{sec:openended_method}). We also score both open-ended benchmarks under all seven variants for the 16 small judges (Tables~\ref{tab:app_e3_openended_se} and~\ref{tab:app_e3_openended_mb}). Open-ended rankings are as stable as closed-ended ones: mean pairwise Kendall's $\tau$ over the five core formulations is 0.85 on SummEval and 0.76 on MT-Bench. On MT-Bench, Phi-4-Reasoning scores above Phi-4 under every one of the seven variants, so the reversal of \S\ref{sec:openended} does not depend on the wording of the scoring prompt.

\begin{table*}[t]
\centering
\rowcolors{2}{white}{slmrow}
\begin{adjustbox}{max width=\textwidth}
\small
\begin{tabular}{lcccccccc}
\toprule
\textbf{Judge} & \textbf{original} & \textbf{minimal} & \textbf{gpt} & \textbf{rubric} & \textbf{cot} & \textbf{advers.} & \textbf{few-shot} & \textbf{Range (core 5)} \\
\midrule
Phi-4 & \phantom{$-$}0.520 & \phantom{$-$}0.445 & 0.496 & 0.424 & \phantom{$-$}0.502 & 0.404 & \phantom{$-$}0.455 & 0.096 \\
Phi-4-Reasoning-Plus & \phantom{$-$}0.496 & \phantom{$-$}0.491 & 0.512 & 0.394 & \phantom{$-$}0.494 & 0.482 & \phantom{$-$}0.506 & 0.118 \\
Qwen3-14B & \phantom{$-$}0.483 & \phantom{$-$}0.458 & 0.481 & 0.347 & \phantom{$-$}0.459 & 0.463 & \phantom{$-$}0.471 & 0.136 \\
Phi-4-Reasoning & \phantom{$-$}0.473 & \phantom{$-$}0.452 & 0.482 & 0.400 & \phantom{$-$}0.483 & 0.455 & \phantom{$-$}0.470 & 0.082 \\
Qwen3-8B & \phantom{$-$}0.453 & \phantom{$-$}0.441 & 0.475 & 0.365 & \phantom{$-$}0.419 & 0.443 & \phantom{$-$}0.425 & 0.111 \\
Qwen3-4B & \phantom{$-$}0.418 & \phantom{$-$}0.393 & 0.411 & 0.327 & \phantom{$-$}0.388 & 0.383 & \phantom{$-$}0.369 & 0.091 \\
Qwen2.5-3B & \phantom{$-$}0.327 & \phantom{$-$}0.360 & 0.404 & 0.239 & \phantom{$-$}0.331 & 0.258 & \phantom{$-$}0.342 & 0.165 \\
\bottomrule
\end{tabular}
\end{adjustbox}
\caption{\textbf{SummEval Spearman $\rho$ under each prompt variant}, mean of the four per-dimension correlations against the human annotations, 16 small judges. Range is over the five core formulations; column names are abbreviated as in Table~\ref{tab:app_e3_t10}. The \texttt{original} column is an independent run of the paper's scoring prompt on different hardware; it is within 0.013 of Table~\ref{tab:app_summeval_full} for every judge except Phi-4-mini (0.309 against 0.375) and Qwen3-0.6B.}
\label{tab:app_e3_openended_se}
\end{table*}

\begin{table*}[t]
\centering
\rowcolors{2}{white}{slmrow}
\begin{adjustbox}{max width=\textwidth}
\small
\setlength{\tabcolsep}{4pt}
\begin{tabular}{lcccccccc}
\toprule
\textbf{Judge} & \textbf{original} & \textbf{minimal} & \textbf{gpt} & \textbf{rubric} & \textbf{cot} & \textbf{advers.} & \textbf{few-shot} & \textbf{Range (core 5)} \\
\midrule
Phi-4-Reasoning-Plus & \phantom{$-$}0.531 & \phantom{$-$}0.416 & \phantom{$-$}0.401 & \phantom{$-$}0.421 & \phantom{$-$}0.457 & \phantom{$-$}0.473 & \phantom{$-$}0.331 & 0.130 \\
Qwen3-14B & \phantom{$-$}0.445 & \phantom{$-$}0.454 & \phantom{$-$}0.507 & \phantom{$-$}0.513 & \phantom{$-$}0.481 & \phantom{$-$}0.435 & \phantom{$-$}0.451 & 0.067 \\
Phi-4-Reasoning & \phantom{$-$}0.399 & \phantom{$-$}0.438 & \phantom{$-$}0.418 & \phantom{$-$}0.435 & \phantom{$-$}0.559 & \phantom{$-$}0.456 & \phantom{$-$}0.353 & 0.159 \\
Qwen3-4B & \phantom{$-$}0.365 & \phantom{$-$}0.383 & \phantom{$-$}0.375 & \phantom{$-$}0.428 & \phantom{$-$}0.474 & \phantom{$-$}0.423 & \phantom{$-$}0.421 & 0.109 \\
Qwen3-8B & \phantom{$-$}0.348 & \phantom{$-$}0.523 & \phantom{$-$}0.445 & \phantom{$-$}0.515 & \phantom{$-$}0.560 & \phantom{$-$}0.464 & \phantom{$-$}0.347 & 0.212 \\
Phi-4 & \phantom{$-$}0.269 & \phantom{$-$}0.186 & \phantom{$-$}0.189 & \phantom{$-$}0.285 & \phantom{$-$}0.197 & \phantom{$-$}0.285 & \phantom{$-$}0.101 & 0.100 \\
Phi-4-mini-Reasoning & \phantom{$-$}0.266 & \phantom{$-$}0.281 & \phantom{$-$}0.301 & \phantom{$-$}0.237 & \phantom{$-$}0.258 & \phantom{$-$}0.271 & \phantom{$-$}0.209 & 0.064 \\
\bottomrule
\end{tabular}
\end{adjustbox}
\caption{\textbf{MT-Bench Spearman $\rho$ under each prompt variant}, against the reference model (Qwen3.5-397B-A17B), mean over the two students. As in Table~\ref{tab:app_e3_openended_se}, the \texttt{original} column is an independent run; with 80 conversations it moves more than on SummEval, by up to 0.13 against Table~\ref{tab:app_mtbench_full}, most for the judges that sample while thinking and for the weakest ones.}
\label{tab:app_e3_openended_mb}
\end{table*}

\section{Protocol Robustness: Details}
\label{app:protocols}

This appendix details the protocol comparison of \S\ref{sec:robustness}. All 16 small judges and two large judges (Qwen2.5-72B and GPT-OSS-120B) run under three alternatives to the paper's reference-based pointwise protocol, on all eight datasets and both students, each at every budget it supports. Appendix~\ref{app:protocol_prompts} gives the prompts verbatim.

\begin{itemize}
    \item \textbf{Reference-free}: the gold answer is withheld and the judge must decide correctness from the question and the student's work alone;
    \item \textbf{G-Eval}: the judge scores the answer from 1 to 10 on a rubric and a score of 7 or more counts as \texttt{Correct}, at $B{=}8{,}192$ only since rubric scoring requires room to reason; the prompt still contains the reference answer;
    \item \textbf{Pairwise}: the judge picks the more correct of the two students' answers to the same question, with the reference answer in the prompt, evaluated in both presentation orders.
\end{itemize}

Only the reference-free protocol removes the reference, so it is the one that measures how much of the task is verification.

\begin{table*}[t]
\centering
\rowcolors{2}{white}{slmrow}
\begin{adjustbox}{max width=\textwidth}
\small
\begin{tabular}{llccccccc}
\toprule
\hiderowcolors
& & \multicolumn{2}{c}{\textbf{Reference-based}} & \multicolumn{2}{c}{\textbf{Reference-free}} & \textbf{G-Eval} & \multicolumn{2}{c}{\textbf{Drop at best budget}} \\
\cmidrule(lr){3-4} \cmidrule(lr){5-6} \cmidrule(lr){7-7} \cmidrule(lr){8-9}
\textbf{Judge} & \textbf{Type} & $B{=}10$ & $B{=}8{,}192$ & $B{=}10$ & $B{=}8{,}192$ & $B{=}8{,}192$ & \textbf{Ref.-free} & \textbf{G-Eval} \\
\midrule
\noalign{\global\rownum=2}\showrowcolors
Phi-4 & SLM & 90.16 & 87.80 & 75.92 & 76.16 & 82.49 & $-$14.00 & \phantom{0}$-$7.67 \\
Qwen3-14B & SLM & 90.15 & 88.60 & 78.49 & 80.80 & 87.24 & \phantom{0}$-$9.35 & \phantom{0}$-$2.91 \\
Qwen2.5-72B & Large & 89.89 & 88.10 & 77.16 & 76.06 & 85.29 & $-$12.73 & \phantom{0}$-$4.61 \\
Qwen3-8B & SLM & 89.61 & 88.08 & 77.13 & 80.02 & 87.08 & \phantom{0}$-$9.59 & \phantom{0}$-$2.53 \\
GPT-OSS-120B & Large & --- & 89.47 & --- & 82.23 & 88.64 & \phantom{0}$-$7.24 & \phantom{0}$-$0.83 \\
Phi-4-Reasoning-Plus & SLM & --- & 89.42 & --- & 80.41 & 87.50 & \phantom{0}$-$9.01 & \phantom{0}$-$1.91 \\
\bottomrule
\end{tabular}
\end{adjustbox}
\caption{\textbf{Cross-protocol accuracy (\%)} for the 16 small judges and two large judges, all eight datasets and both students per cell, sorted by best reference-based accuracy. The reference-based columns are the paper's own protocol (Tables~\ref{tab:individual} and~\ref{tab:large_comparison}). G-Eval requires rubric reasoning and runs at $B{=}8{,}192$ only. The two drop columns, in points, compare each protocol's best budget with the judge's best reference-based budget.}
\label{tab:protocols}
\end{table*}

Table~\ref{tab:protocols} gives the two pointwise alternatives against the paper's own protocol. Reference-based judging is the most accurate protocol for 17 of the 18 judges; the exception is Llama-3.2-1B, whose G-Eval accuracy (67.53\%) edges past its reference-based 65.61\%. Withholding the reference costs every judge, large judges included (Qwen2.5-72B $-$12.73, GPT-OSS-120B $-$7.24 points), and without it the strongest SLMs land between the two large judges: Qwen3-14B reaches 80.80\%, Phi-4-Reasoning-Plus 80.41\%, and Qwen3-8B 80.02\%, against 77.16\% for Qwen2.5-72B and 82.23\% for GPT-OSS-120B. G-Eval, which keeps the reference, costs much less: ten of the eighteen judges lose under three points, among them every Qwen3 judge, the three reasoning-trained Phi-4 judges, and GPT-OSS-120B (0.83), and Llama-3.2-1B gains 1.92.

\begin{table}[!htbp]
\centering
\rowcolors{2}{white}{slmrow}
\begin{adjustbox}{max width=\columnwidth}
\small
\setlength{\tabcolsep}{4pt}
\begin{tabular}{lccccc}
\toprule
\textbf{Judge} & \textbf{Budget} & \textbf{Mean flip} & \textbf{Min} & \textbf{Max} & \textbf{Read (\%)} \\
\midrule
GPT-OSS-120B & $B{=}8{,}192$ & 0.481 & 0.317 & 0.786 & 100.0 \\
Llama-3.1-8B & both$^\ddagger$ & 0.843 & 0.788 & 0.901 & \phantom{0}91.3 \\
Llama-3.2-1B & $B{=}10$ & 0.891 & 0.697 & 1.000 & \phantom{0}23.7 \\
Llama-3.2-1B & $B{=}8{,}192$ & 0.832 & 0.694 & 0.895 & \phantom{0}47.7 \\
Llama-3.2-3B & both$^\ddagger$ & 0.984 & 0.943 & 1.000 & \phantom{0}99.4 \\
Phi-4 & $B{=}10$ & 0.762 & 0.474 & 0.921 & \phantom{0}90.0 \\
Phi-4 & $B{=}8{,}192$ & 0.749 & 0.436 & 0.920 & \phantom{0}95.7 \\
\bottomrule
\end{tabular}
\end{adjustbox}
\caption{\textbf{Pairwise position-bias flip rate} over all eight datasets: the mean over 16 cells (eight datasets, each student listed first once), with the lowest and highest cell, computed over the pairs where both orders give a readable choice. Read = share of responses with a readable choice. $^\ddagger$Llama-3.1-8B, Llama-3.2-3B, Qwen2.5-3B, Qwen2.5-7B, and Qwen2.5-72B decode deterministically and give essentially the same single-letter choice at either budget, so one row covers both.}
\label{tab:app_pairwise}
\end{table}

Table~\ref{tab:app_pairwise} summarizes the pairwise runs by their flip rate: the fraction of items where swapping the presentation order flips the judge's preference, where 0 is perfect consistency, 0.5 is chance, and 1.0 is answering by position alone. The choice is read from the judge's committed answer: GPT-OSS-120B's final channel, a response or a first or last line that is a single letter, a boxed letter, or a closing sentence that names exactly one answer (``Answer B is more correct''). Responses that name both answers, decline to choose, or stop mid-explanation count as unreadable and are left out; Llama-3.2-1B keeps fewer than half of its items this way. Most judges answer largely by position, and size does not fix it: Qwen2.5-72B has a mean flip rate of 0.827 at either budget, and GPT-OSS-120B, at 0.481, is no more consistent than chance. This mirrors the position bias documented for LLM judges generally, and is why \slmjury{} is built on pointwise verdicts rather than pairwise preferences. Thinking mode is what buys resistance: with it active at $B{=}8{,}192$, Qwen3-14B cuts its flip rate from 0.785 to 0.451, Qwen3-8B from 0.844 to 0.487, and Qwen3-4B from 0.862 to 0.622, and the always-thinking Phi-4-Reasoning-Plus (0.391) is the most consistent judge in the panel, ahead of both large judges. A larger budget without thinking helps little: Phi-4 moves from 0.762 to 0.749 and Phi-4-mini from 0.785 to 0.774.

\subsection{Protocol Prompts}
\label{app:protocol_prompts}

The reference-free prompts follow the paper's quick and reasoned prompts (Appendix~\ref{app:prompts}) with the ground truth removed and the instruction reworded to match. The G-Eval and pairwise prompts keep the ground truth.

\begin{promptflow}{Reference-Free Prompt (\texttt{max\_tokens=10})}
\begin{verbatim}
Your role is to evaluate the student's answer
to the question.

Respond with exactly one word:
- 'Correct': If the student's final answer
  is correct.
- 'Incorrect': If the student's final answer
  is incorrect.

Focus only on the final answer, ignore
reasoning steps.

Question: {question}

Student answer: {student_answer}
\end{verbatim}
\end{promptflow}

\begin{promptflow}{Reference-Free Prompt (\texttt{max\_tokens=8192})}
\begin{verbatim}
Your role is to evaluate the student's answer
to the question.

First, explain your reasoning for the judgment.
Then, on a new line, give the final verdict in
the exact format:
\boxed{CORRECT} or \boxed{INCORRECT}

Focus only on the final answer when judging
correctness. Ignore reasoning steps in the
student's solution.

Question: {question}

Student answer: {student_answer}
\end{verbatim}
\end{promptflow}

\begin{promptflow}{G-Eval Prompt (\texttt{max\_tokens=8192}; score $\geq$ 7 counts as \texttt{Correct})}
\begin{verbatim}
Your role is to evaluate the quality of the
student's answer.

Assign a score from 1 to 10, where:
  1-3 = Clearly incorrect or missing the key
        concepts
  4-6 = Partially correct or missing important
        details
  7-10 = Correct and complete

Give your score in the format: Score: X

Question: {question}

Ground truth: {ground_truth}

Student answer: {student_answer}
\end{verbatim}
\end{promptflow}

\begin{promptflow}{Pairwise Prompt (both budgets; each pair is judged in both orders)}
\begin{verbatim}
You are evaluating two student answers to a
question.

Decide which answer is more correct.
Respond with exactly 'A' or 'B'.

Question: {question}

Ground truth: {ground_truth}

Answer A: {answer_a}

Answer B: {answer_b}
\end{verbatim}
\end{promptflow}

\section{Error Analysis}
\label{app:error_analysis}

This section provides a deeper analysis of judge errors against the oracle, examining when and why verdicts depart from it.

\subsection{Error Distribution by Verdict Type}

Disagreements with the oracle fall into two categories: \textit{false positives} (accepting a solution the oracle rejects) and \textit{false negatives} (rejecting a solution the oracle accepts). For Phi-4, Qwen3-14B, and Qwen3-8B, the three most accurate small judges at $B{=}10$, \textbf{false positives dominate}: 85--90\% of disagreements accept an answer the oracle rejects rather than reject one it accepts (Appendix~\ref{app:error_taxonomy}). The remaining \textbf{false negatives cluster on general reasoning}, where they arise when the judge misreads the context-dependent nature of the correct answer, while on competition-level math the judge only occasionally flags correct solutions that use uncommon notation or non-standard derivations (which the SymPy oracle still resolves).

\subsection{Overthinking Error Patterns}

The overthinking effect (where extended reasoning degrades accuracy) produces domain-dependent error patterns. For Qwen2.5-3B on math, which shows the largest overthinking penalty, errors at $B{=}10$ are sparse and evenly distributed (96.06\% on GSM8K, 96.01\% on GSM-Plus). At $B{=}8{,}192$ accuracy drops to 92.91\% and 85.50\% respectively ($-$10.51 points on GSM-Plus), and the errors concentrate on problems where the student's reasoning is partially correct but the final answer is wrong: the judge's extended reasoning follows the student's chain of thought, gets led astray by plausible intermediate steps, and reaches an incorrect verdict. This is a form of reasoning sycophancy, where the judge defers to the student's process rather than independently checking the answer. Conversely, on general reasoning benchmarks, extended reasoning \textit{helps}. Qwen3-4B gains 23.09 points on WinoGrande (71.47\% $\to$ 94.55\%) because pronoun resolution requires step-by-step disambiguation that a 10-token verdict cannot provide. This effect in both directions explains the domain-dependent overthinking result (\S\ref{sec:individual}).

\subsection{Debate-Induced Errors}

Multi-agent debate introduces a distinct failure mode: \textit{correct-to-incorrect flips}. Problems where all three agents open with the same verdict are untouched, since consensus halts further rounds, so the protocol can only act on the items where they disagree. Restricting to those, and further to the ones where at least one agent opened with the right verdict, the debate ends on the wrong answer in 43.2--71.5\% of cases across the five Variant~B configurations. The RCR ``Critique'' step lets two opposing arguments talk a correct agent out of its verdict nearly as often as it repairs a wrong one, and for Qwen3-4B more often. Qwen3-4B is the worst affected at 71.5\% (1{,}186 of 1{,}658), which is why it drops from 88.44\% individually to 87.67\% in debate despite disagreeing internally on only 2.6\% of items. The temperature diversity of Variant~B ($\tau \in \{0.0, 0.4, 0.9\}$) supplies those disagreements in the first place, occasionally overturning a correct verdict even when the deterministic agent ($\tau{=}0.0$) was right.

This pattern is amplified in Variant A cross-architecture debates, where weaker agents can corrupt the judgments of stronger ones. The best Variant A configuration at $B{=}8{,}192$ (Qwen3-4B + Qwen2.5-3B + Qwen3-1.7B at 85.12\%) falls 3.32 points below Qwen3-4B's individual accuracy at $B{=}8{,}192$ (88.44\%), showing that architectural diversity does not compensate for the persuasion vulnerability.

\subsection{Persona-Induced Error Patterns}

The Lenient persona's effect on Llama-3.1-8B ($-$10.54 points) follows a clear pattern: the model begins accepting incorrect solutions that have plausible reasoning chains. The Lenient system prompt tells the judge to ``Give partial credit for partially correct reasoning'' and ``Do not heavily penalize minor factual or grammatical errors,'' which the model interprets as lowering the threshold for correctness. This transforms the binary classification boundary from ``is the answer correct?'' to ``is the reasoning attempt reasonable?,'' a fundamentally different and easier-to-satisfy criterion.

The Strict persona matters most once a judge reasons. Phi-4-mini at $B{=}8{,}192$ loses 11.90 points under the Strict persona, the largest persona-induced drop in the study. At extended token budgets, the Strict framing causes the model to second-guess correct verdicts, producing false negatives. At $B{=}10$ the same persona costs it 1.36 points, since a one-word verdict leaves little room for doubt.

In contrast, the top judges (Qwen3-14B with a 0.10-point range, Phi-4 with a 0.56-point range at $B{=}10$) show robust invariance to all six personas, suggesting that well-trained models internalize a stable evaluation criterion that overrides surface-level framing.

\section{Full Open-Ended Evaluation Results}
\label{app:openended_full}

This section presents the complete open-ended evaluation results for the 16 SLM judges, the six large judges, and the ladder judge Qwen3-32B, each with a 95\% bootstrap interval on Spearman $\rho$ (Appendix~\ref{app:paired_stats}).
SummEval measures SLM-Human correlation (judge scores vs.\ human expert annotations on machine-generated summaries), while MT-Bench measures SLM-LLM correlation (judge scores vs.\ the scores of one large reference model, Qwen3.5-397B-A17B, on multi-turn conversations).
Table~\ref{tab:openended} summarizes the divergence discussed in \S\ref{sec:openended}, and Tables~\ref{tab:app_summeval_full} and~\ref{tab:app_mtbench_full} give the full rankings.

\begin{table}[!htbp]
\centering
\begin{adjustbox}{max width=\columnwidth}
\small
\prettytable
\setlength{\tabcolsep}{4pt}
\begin{tabular}{lcccc}
\toprule
\textbf{Judge} & \textbf{CE Rank} & \textbf{SE $\rho$} & \textbf{MB $\rho$} & \textbf{MB Rank} \\
\midrule
Phi-4 & 1 & \textbf{0.520} & 0.269 & 6 \\
Phi-4-Reasoning-Plus & 4 & 0.489 & 0.405 & 3 \\
Qwen3-14B & 2 & 0.487 & \textbf{0.516} & \textbf{1} \\
Phi-4-Reasoning & 5 & 0.479 & 0.505 & 2 \\
Qwen3-8B & 3 & 0.453 & 0.359 & 4 \\
\bottomrule
\end{tabular}
\end{adjustbox}
\caption{\textbf{Top-5 SLM judges by SummEval $\rho$}, with MT-Bench $\rho$ and closed-ended (CE) rank; both ranks are among the 16 SLMs. Phi-4 leads the SLMs on closed-ended judging and on SummEval, yet ranks sixth on MT-Bench.}
\label{tab:openended}
\end{table}

\subsection[SummEval: Full Results]{SummEval: Full SLM-Human Correlation Results}

\begin{figure*}[tp]
\centering
\includegraphics[width=\textwidth]{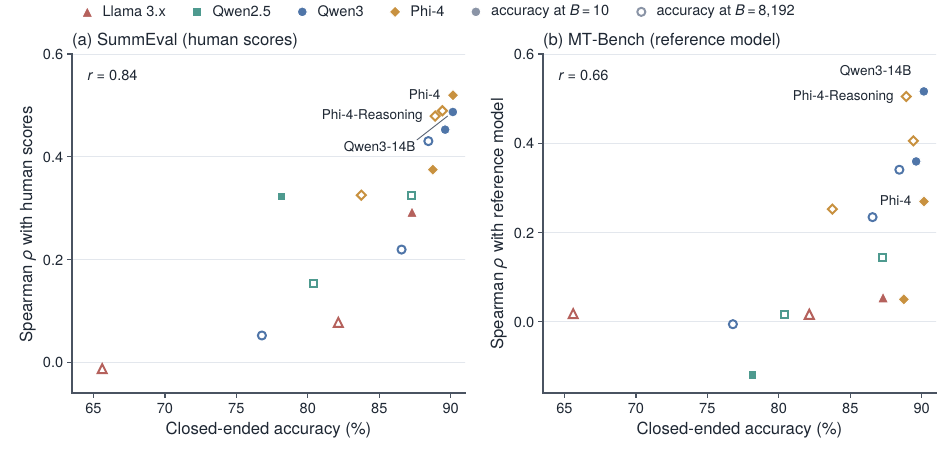}
\caption{\textbf{Closed-ended accuracy and open-ended agreement for the 16 SLMs.} Each point is one SLM: its closed-ended accuracy at its best budget (filled: $B{=}10$, open: $B{=}8{,}192$) against its Spearman $\rho$ at $B{=}8{,}192$ with (a) the SummEval human scores and (b) the MT-Bench reference model. Closed-ended accuracy correlates with SummEval agreement at $r{=}0.84$ and with MT-Bench agreement at $r{=}0.66$. Phi-4 leads the SLMs on closed-ended accuracy and SummEval but ranks sixth on MT-Bench, where Phi-4-Reasoning ranks second.}
\label{fig:paradigm_divergence}
\end{figure*}

Table~\ref{tab:app_summeval_full} reports all five agreement metrics (Pearson $r$, Spearman $\rho$, Cohen's $\kappa$, accuracy, MSE) between each judge's scores and the human expert annotations, averaged over the four SummEval dimensions, with an interval on $\rho$ that resamples the 100 source articles. Qwen3.5-122B-A10B is the most human-aligned judge ($\rho{=}0.542$), and Phi-4 follows at 0.520, ahead of the other five large judges. Among the SLMs the Phi-4 and Qwen3 families dominate, while the smallest judges fall to near-zero correlation, mirroring the closed-ended capability ordering. Over the 16 small judges, accuracy at each judge's best budget correlates with SummEval $\rho$ at Pearson $r{=}0.84$ and with MT-Bench $\rho$ at $r{=}0.66$ (Figure~\ref{fig:paradigm_divergence}).

\begin{figure*}[tp]
\centering
\includegraphics[width=\textwidth]{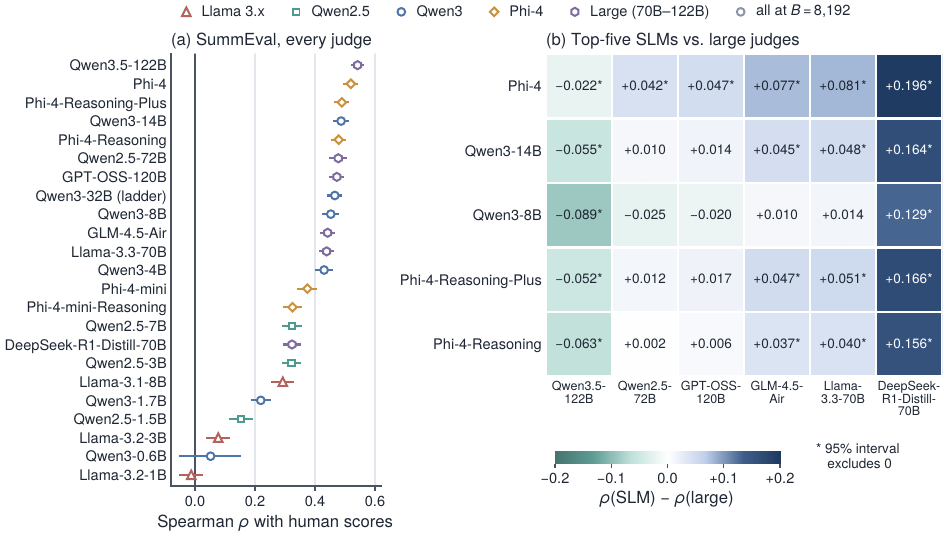}
\caption{\textbf{SummEval: small vs.\ large judges against the human annotations.} (a) Spearman $\rho$ between each judge's scores and the human annotations (mean of the four dimensions, 1{,}600 summaries, every judge at $B{=}8{,}192$), with the 95\% interval that resamples the 100 source articles. (b) Each of the five most accurate SLMs against each large judge on the same summaries: $\rho$(SLM) $-$ $\rho$(large), where * marks a paired 95\% interval that excludes zero. The SLM is higher in 15 comparisons, not distinguishable in 10, and lower in 5, all of them against Qwen3.5-122B-A10B (Table~\ref{tab:app_summeval_paired}).}
\label{fig:summeval_small_vs_large}
\end{figure*}

\begin{table*}[tp]
\centering
\small
\begin{adjustbox}{max width=\textwidth}
\rowcolors{2}{white}{slmrow}
\begin{adjustbox}{max width=\textwidth}
\setlength{\tabcolsep}{4pt}
\begin{tabular}{llcc@{\hspace{4pt}}cccc}
\toprule
\textbf{Group} & \textbf{Judge Model} & \textbf{Pearson} & \textbf{Spearman} & \textbf{95\% CI} & \textbf{Cohen's $\kappa$} & \textbf{Accuracy} & \textbf{MSE} \\
\midrule
    & Llama-3.1-8B       & \phantom{$-$}0.380 & \phantom{$-$}0.293 & \cipad{$-$}{0.255}{0.329}{} & \phantom{$-$}0.240 & 72.81 & \textbf{0.954} \\
    & Llama-3.2-3B       & \phantom{$-$}0.101 & \phantom{$-$}0.078 & \cipad{$-$}{0.037}{0.118}{} & \phantom{$-$}0.034 & 65.77 & 1.412 \\
\multirow{-3}{*}{Llama 3.x}
    & Llama-3.2-1B       & $-$0.016 & $-$0.013 & \cipad{}{$-$0.052}{0.028}{} & $-$0.002 & 47.34 & 2.853 \\
\midrule
    & Qwen2.5-7B         & \phantom{$-$}0.355 & \phantom{$-$}0.324 & \cipad{$-$}{0.291}{0.356}{} & \phantom{$-$}0.181 & 63.78 & 1.797 \\
    & Qwen2.5-3B         & \phantom{$-$}0.333 & \phantom{$-$}0.322 & \cipad{$-$}{0.290}{0.354}{} & \phantom{$-$}0.088 & 43.61 & 3.893 \\
\multirow{-3}{*}{Qwen2.5}
    & Qwen2.5-1.5B       & \phantom{$-$}0.175 & \phantom{$-$}0.154 & \cipad{$-$}{0.115}{0.194}{} & \phantom{$-$}0.065 & 69.03 & 1.167 \\
\midrule
    & Qwen3-14B          & \phantom{$-$}0.519 & \phantom{$-$}0.487 & \cipad{$-$}{0.461}{0.512}{} & \phantom{$-$}0.350 & 68.67 & 1.438 \\
\bottomrule
\end{tabular}
\end{adjustbox}
\end{adjustbox}
\caption{\textbf{Full SummEval (SLM-Human) results for all judges.} Correlations are computed between judge scores and human expert annotations on machine-generated summaries across 4 dimensions (coherence, consistency, fluency, relevance) and averaged. The 95\% interval on Spearman $\rho$ resamples the 100 source articles (10{,}000 resamples). All judges use $B{=}8{,}192$. \textbf{Bold} = best per column.}
\label{tab:app_summeval_full}
\end{table*}

\subsection[SummEval: Small vs.\ Large, Paired]{SummEval: Small vs.\ Large\\ Judges, Paired}

Table~\ref{tab:app_summeval_paired} compares the five most accurate SLMs on the closed-ended task with each large judge on the same 1{,}600 summaries, resampling the 100 source articles so that every judge is scored on the same draws. The SLM is significantly more human-aligned in 15 of the 30 comparisons, the two are not distinguishable in 10, and the large judge is ahead in 5, all of them against Qwen3.5-122B-A10B. Phi-4 is significantly above every large judge except Qwen3.5-122B-A10B, which leads it by 0.022 with an interval of $[0.001, 0.043]$. Figure~\ref{fig:summeval_small_vs_large} plots every judge's $\rho$ and the 30 paired differences.

\begin{table*}[tp]
\centering
\rowcolors{2}{white}{slmrow}
\begin{adjustbox}{max width=\textwidth}
\small
\begin{tabular}{llcccl}
\toprule
\textbf{SLM} & \textbf{Large judge} & \textbf{SLM $\rho$} & \textbf{Large $\rho$} & \textbf{95\% CI of difference} & \textbf{Reading} \\
\midrule
Phi-4 & Qwen3.5-122B-A10B & 0.520 & 0.542 & \cipad{}{$-$0.043}{$-$0.001}{} & large higher \\
Phi-4 & GLM-4.5-Air & 0.520 & 0.442 & \cipad{$-$}{0.056}{0.098}{$-$} & SLM higher \\
Phi-4 & GPT-OSS-120B & 0.520 & 0.473 & \cipad{$-$}{0.024}{0.069}{$-$} & SLM higher \\
Phi-4 & Qwen2.5-72B & 0.520 & 0.478 & \cipad{$-$}{0.019}{0.066}{$-$} & SLM higher \\
Phi-4 & Llama-3.3-70B & 0.520 & 0.439 & \cipad{$-$}{0.059}{0.103}{$-$} & SLM higher \\
Phi-4 & DeepSeek-R1-Distill-70B & 0.520 & 0.323 & \cipad{$-$}{0.168}{0.225}{$-$} & SLM higher \\
\midrule
Qwen3-14B & Qwen3.5-122B-A10B & 0.487 & 0.542 & \cipad{}{$-$0.074}{$-$0.035}{} & large higher \\
\bottomrule
\end{tabular}
\end{adjustbox}
\caption{\textbf{SummEval, each of the five most accurate SLMs against each large judge}, paired on the same 1{,}600 summaries. The interval is on the SLM's $\rho$ minus the large judge's, resampling the 100 source articles (10{,}000 resamples). \emph{n.d.} = the interval contains zero.}
\label{tab:app_summeval_paired}
\end{table*}

\subsection[SummEval: Per Dimension]{SummEval: Per-Dimension Breakdown}

Table~\ref{tab:app_summeval_dims} shows Spearman correlation with the human annotations broken down by the four SummEval dimensions for every judge. Among the eight judges most aligned with the human scores, coherence draws the highest rank correlation for seven and fluency the lowest for all eight, so judges follow logical structure more closely than surface quality; across all 23 judges the counts are 13 and 17. The two measures disagree about which dimension is easiest: Cohen's $\kappa$ puts consistency first for seven of the top eight (16 of 23), since it binarizes both scores at 4 (Appendix~\ref{app:scoring_metrics}) and so rewards agreement on which summaries are good rather than on their ordering.

\begin{table*}[tp]
\centering
\small
\begin{adjustbox}{max width=\textwidth}
\rowcolors{2}{white}{slmrow}
\begin{adjustbox}{max width=\textwidth}
\setlength{\tabcolsep}{4pt}
\begin{tabular}{lcccccccc}
\toprule
\hiderowcolors
& \multicolumn{4}{c}{\textbf{Spearman $\rho$}} & \multicolumn{4}{c}{\textbf{Cohen's $\kappa$}} \\
\cmidrule(lr){2-5} \cmidrule(lr){6-9}
\textbf{Judge} & \textbf{Coher.} & \textbf{Consist.} & \textbf{Fluency} & \textbf{Relev.} & \textbf{Coher.} & \textbf{Consist.} & \textbf{Fluency} & \textbf{Relev.} \\
\midrule
\noalign{\global\rownum=2}\showrowcolors
Qwen3.5-122B-A10B & \phantom{$-$}\textbf{0.656} & \phantom{$-$}\textbf{0.636} & 0.331 & \phantom{$-$}\textbf{0.545} & \textbf{0.495} & \phantom{$-$}\textbf{0.590} & \phantom{$-$}0.024 & \phantom{$-$}\textbf{0.430} \\
Phi-4 & \phantom{$-$}0.619 & \phantom{$-$}0.555 & 0.423 & \phantom{$-$}0.481 & 0.472 & \phantom{$-$}0.519 & \phantom{$-$}0.223 & \phantom{$-$}0.383 \\
Phi-4-Reasoning-Plus & \phantom{$-$}0.583 & \phantom{$-$}0.501 & 0.382 & \phantom{$-$}0.492 & 0.424 & \phantom{$-$}0.452 & \phantom{$-$}0.048 & \phantom{$-$}0.373 \\
Qwen3-14B & \phantom{$-$}0.599 & \phantom{$-$}0.521 & 0.370 & \phantom{$-$}0.459 & 0.427 & \phantom{$-$}0.486 & \phantom{$-$}0.107 & \phantom{$-$}0.380 \\
Phi-4-Reasoning & \phantom{$-$}0.571 & \phantom{$-$}0.486 & 0.367 & \phantom{$-$}0.492 & 0.444 & \phantom{$-$}0.516 & \phantom{$-$}0.071 & \phantom{$-$}0.389 \\
Qwen2.5-72B & \phantom{$-$}0.460 & \phantom{$-$}0.531 & \textbf{0.438} & \phantom{$-$}0.481 & 0.344 & \phantom{$-$}0.409 & \phantom{$-$}0.288 & \phantom{$-$}0.265 \\
\bottomrule
\end{tabular}
\end{adjustbox}
\end{adjustbox}
\caption{\textbf{SummEval per-dimension correlations with the human annotations} for all judges (sorted by overall Spearman). Coherence leads on Spearman $\rho$ and consistency on Cohen's $\kappa$ for most judges; fluency is the weakest dimension on $\rho$. \textbf{Bold} = best per column.}
\label{tab:app_summeval_dims}
\end{table*}

\subsection[MT-Bench: Full Results]{MT-Bench: Full SLM-LLM\\ Correlation Results}

Table~\ref{tab:app_mtbench_full} reports the same five metrics against the reference model's scores, computed per student and averaged over the two, with an interval on $\rho$ that resamples the 80 questions. The two strongest large judges lead (Qwen3.5-122B-A10B 0.592, GPT-OSS-120B 0.566). Among the SLMs the ordering shifts toward reasoning: Qwen3-14B (0.516) and Phi-4-Reasoning (0.505) lead, and Phi-4 falls to sixth. With 80 conversations the intervals are wide and overlap heavily, and the reference is itself noisy (Appendix~\ref{app:large_judges}).

\begin{table*}[tp]
\centering
\small
\begin{adjustbox}{max width=\textwidth}
\rowcolors{2}{white}{slmrow}
\begin{adjustbox}{max width=\textwidth}
\setlength{\tabcolsep}{4pt}
\begin{tabular}{llcc@{\hspace{4pt}}cccc}
\toprule
\textbf{Group} & \textbf{Judge Model} & \textbf{Pearson} & \textbf{Spearman} & \textbf{95\% CI} & \textbf{Cohen's $\kappa$} & \textbf{Accuracy} & \textbf{MSE} \\
\midrule
    & Llama-3.1-8B       & \phantom{$-$}0.056 & \phantom{$-$}0.054 & \cipad{}{$-$0.091}{0.204}{} & $-$0.013 & 77.50 & 1.494 \\
    & Llama-3.2-3B       & $-$0.026 & \phantom{$-$}0.016 & \cipad{}{$-$0.106}{0.140}{} & $-$0.039 & 75.00 & 1.531 \\
\multirow{-3}{*}{Llama 3.x}
    & Llama-3.2-1B       & \phantom{$-$}0.038 & \phantom{$-$}0.018 & \cipad{}{$-$0.182}{0.221}{} & \phantom{$-$}0.073 & 74.38 & 1.638 \\
\midrule
    & Qwen2.5-7B         & \phantom{$-$}0.107 & \phantom{$-$}0.143 & \cipad{}{$-$0.011}{0.302}{} & $-$0.012 & 78.13 & 1.325 \\
    & Qwen2.5-3B         & $-$0.094 & $-$0.120 & \cipad{}{$-$0.285}{0.046}{} & $-$0.058 & 74.38 & 1.794 \\
\multirow{-3}{*}{Qwen2.5}
    & Qwen2.5-1.5B       & $-$0.006 & \phantom{$-$}0.017 & \cipad{}{$-$0.127}{0.158}{} & $-$0.021 & 77.50 & 1.419 \\
\midrule
    & Qwen3-14B          & \phantom{$-$}0.464 & \phantom{$-$}0.516 & \cipad{$-$}{0.369}{0.646}{} & \phantom{$-$}0.285 & 86.25 & \textbf{0.756} \\
\bottomrule
\end{tabular}
\end{adjustbox}
\end{adjustbox}
\caption{\textbf{Full MT-Bench (SLM-LLM) results for all judges.} Correlations are computed between judge scores and the scores of the reference model, Qwen3.5-397B-A17B, per student (80 questions each) and averaged over the two students. The 95\% interval on Spearman $\rho$ resamples the 80 questions (10{,}000 resamples). All judges use $B{=}8{,}192$. \textbf{Bold} = best per column.}
\label{tab:app_mtbench_full}
\end{table*}

\paragraph{Key Observations.}
\textbf{SLM-Human and SLM-LLM rankings diverge}: among the SLMs, Phi-4 (rank 1 on SummEval, $\rho{=}0.520$) ranks sixth on MT-Bench ($\rho{=}0.269$), while Phi-4-Reasoning (rank 4 on SummEval) ranks second ($\rho{=}0.505$). The two share a base model, so the pair is close to a controlled comparison, and the reversal holds under paired resampling: Phi-4-Reasoning is ahead on MT-Bench by 0.235 (95\% interval $[0.033, 0.396]$) and behind on SummEval by 0.040 ($[0.020, 0.061]$). It also holds under every prompt variant we test on MT-Bench (Appendix~\ref{app:prompt_sensitivity}). \textbf{Reasoning-trained models outscore their siblings on MT-Bench}: Phi-4-Reasoning (0.505) and Phi-4-Reasoning-Plus (0.405) both score above Phi-4 (0.269), and Phi-4-mini-Reasoning (0.252) above Phi-4-mini (0.050). \textbf{The two smallest models produce near-random correlations}: Llama-3.2-1B and Qwen3-0.6B both score Spearman $\rho < 0.08$ on both benchmarks, marking a minimum capability threshold for meaningful open-ended evaluation. \textbf{Accuracy and correlation can diverge}: Qwen2.5-1.5B reaches 77.50\% thresholded accuracy on MT-Bench yet $\rho = 0.017$, assigning similar scores to most responses without discriminating quality.

\section{Large Judges: Full Results}
\label{app:large_judges}

\begin{figure*}[!tp]
\centering
\includegraphics[width=\textwidth]{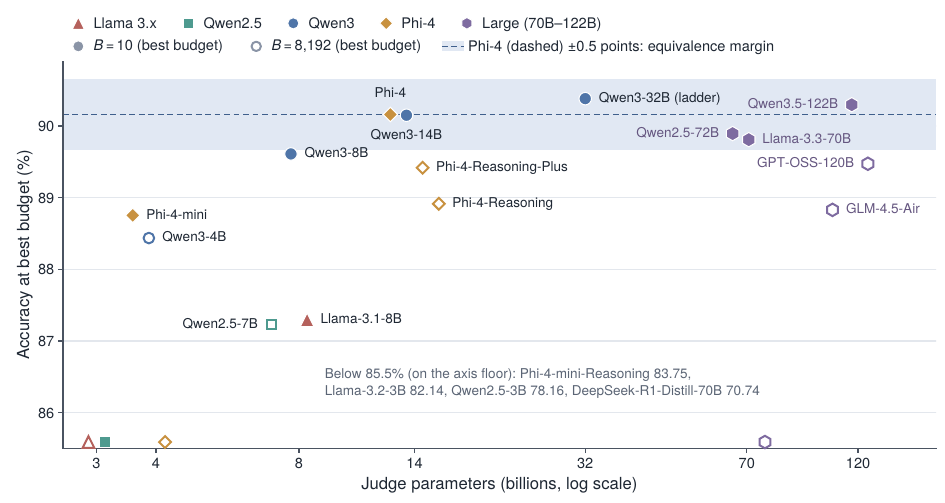}
\caption{\textbf{Accuracy saturates near 8--14B.} Each judge at its best budget on the identical eight datasets (filled: $B{=}10$, open: $B{=}8{,}192$); judges of 3B and more are shown, same-size judges are spread slightly apart, and the four below 85.5\% sit on the axis floor and are listed with their accuracy. The shaded band is the $\pm$0.5-point equivalence margin around the best SLM (Phi-4). Qwen2.5-72B, Llama-3.3-70B, and Qwen3.5-122B-A10B land inside it, as does the ladder judge Qwen3-32B; GPT-OSS-120B and GLM-4.5-Air land below it. The x-axis is logarithmic: going from 14B to 32B or 122B moves accuracy by at most a quarter of a point.}
\label{fig:slm_vs_large}
\end{figure*}

This appendix expands the small-vs-large comparison of \S\ref{sec:scaling}. The six large judges (Llama-3.3-70B-Instruct, DeepSeek-R1-Distill-Llama-70B, Qwen2.5-72B-Instruct, GLM-4.5-Air, GPT-OSS-120B, Qwen3.5-122B-A10B) and the ladder judge Qwen3-32B run under the identical pipeline as the small judges: same students, same prompts, same verdict parser, same oracle. GPT-OSS-120B and DeepSeek-R1-Distill-Llama-70B always reason and, like the Phi-4 reasoning variants, are evaluated at $B{=}8{,}192$ only, as is GLM-4.5-Air. Serving configurations, including the two FP8 checkpoints and GPT-OSS-120B's greedy decoding, appear in Appendix~\ref{app:models}. DeepSeek-R1-Distill-Llama-70B's responses were stored without byte-level detokenization and with its reasoning trace left in; we report its verdicts as stored, and cleaning the text would lower its accuracy from 70.74\% to 67.77\% and its instruction following from 87.06\% to 82.79\%. Table~\ref{tab:app_all_judges} gives every judge's accuracy, instruction-following rate, and open-ended agreement, and Figure~\ref{fig:slm_vs_large} plots accuracy against size.

\begin{table*}[tp]
\centering
\rowcolors{2}{white}{slmrow}
\begin{adjustbox}{max width=\textwidth}
\small
\begin{tabular}{lcccccccc}
\toprule
\hiderowcolors
& & & \multicolumn{2}{c}{\textbf{Accuracy (\%)}} & \multicolumn{2}{c}{\textbf{IFR (\%)}} & \multicolumn{2}{c}{\textbf{Open-ended $\rho$}} \\
\cmidrule(lr){4-5} \cmidrule(lr){6-7} \cmidrule(lr){8-9}
\textbf{Judge} & \textbf{Params} & \textbf{Type} & $B{=}10$ & $B{=}8{,}192$ & $B{=}10$ & $B{=}8{,}192$ & \textbf{SummEval} & \textbf{MT-B.} \\
\midrule
\noalign{\global\rownum=2}\showrowcolors
Qwen3-32B & 32B & Ladder & \textbf{90.38} & 88.14 & 100.00 & \phantom{0}99.99 & \phantom{$-$}0.466 & \phantom{$-$}0.462 \\
Qwen3.5-122B-A10B & 122B & Large & 90.30 & \textbf{89.82} & 100.00 & 100.00 & \phantom{$-$}\textbf{0.542} & \phantom{$-$}\textbf{0.592} \\
Phi-4 & 14B & SLM & 90.16 & 87.80 & \phantom{0}99.98 & 100.00 & \phantom{$-$}0.520 & \phantom{$-$}0.269 \\
Qwen3-14B & 14B & SLM & 90.15 & 88.60 & 100.00 & 100.00 & \phantom{$-$}0.487 & \phantom{$-$}0.516 \\
Qwen2.5-72B & 72B & Large & 89.89 & 88.10 & \phantom{0}99.99 & 100.00 & \phantom{$-$}0.478 & \phantom{$-$}0.218 \\
Llama-3.3-70B & 70B & Large & 89.81 & 88.86 & 100.00 & \phantom{0}99.99 & \phantom{$-$}0.439 & \phantom{$-$}0.190 \\
\bottomrule
\end{tabular}
\end{adjustbox}
\caption{\textbf{All 22 judges and the ladder judge}, sorted by best closed-ended accuracy, all eight datasets and both students ($N{=}64{,}824$ per configuration). \textbf{Bold} = column best on accuracy and $\rho$. SummEval $\rho$ is against the human annotations; MT-Bench $\rho$ is against the reference model Qwen3.5-397B-A17B. Intervals for both are in Appendix~\ref{app:openended_full}.}
\label{tab:app_all_judges}
\end{table*}

\subsection{Paired Comparisons}

Every judge scores the same 64{,}824 items, so differences between judges are tested on paired items with the procedure of Appendix~\ref{app:paired_stats}. Table~\ref{tab:app_paired} compares each large judge with the five most accurate SLMs, each judge at its best budget. Qwen3.5-122B-A10B, the most accurate, is 0.136 points above Phi-4, inside the $\pm$0.5-point margin, so the two are equivalent (McNemar's $p{=}0.012$; paired interval $[+0.032, +0.239]$). Qwen2.5-72B and Llama-3.3-70B trail Phi-4 by 0.27 and 0.35 points, differences that are significant and still equivalent within the margin. GPT-OSS-120B and GLM-4.5-Air are significantly less accurate than Phi-4 and Qwen3-14B, and DeepSeek-R1-Distill-Llama-70B is 18 to 19 points below each of the five.

\begin{table*}[tp]
\centering
\rowcolors{2}{white}{slmrow}
\begin{adjustbox}{max width=\textwidth}
\small
\setlength{\tabcolsep}{4pt}
\begin{tabular}{llcccccl}
\toprule
\textbf{Large judge} & \textbf{SLM} & \textbf{$\Delta$ micro} & \textbf{$\Delta$ macro} & \textbf{Discordant} & \textbf{McNemar $p$} & \textbf{95\% CI} & \textbf{Reading} \\
\midrule
Qwen3.5-122B-A10B & Phi-4 & \phantom{0}$+$0.136 & \phantom{0}$+$0.225 & 638 / 550 & 0.012 & \cipad{$-$0}{0.032}{0.239}{$-$0} & equivalent \\
Qwen3.5-122B-A10B & Qwen3-14B & \phantom{0}$+$0.147 & \phantom{0}$+$0.282 & 448 / 353 & $<$0.001 & \cipad{$-$0}{0.060}{0.231}{$-$0} & equivalent \\
Qwen3.5-122B-A10B & Qwen3-8B & \phantom{0}$+$0.686 & \phantom{0}$+$0.364 & 722 / 277 & $<$0.001 & \cipad{$-$0}{0.589}{0.777}{$-$0} & large better \\
Qwen3.5-122B-A10B & Phi-4-Reasoning-Plus & \phantom{0}$+$0.879 & \phantom{0}$+$0.268 & 788 / 218 & $<$0.001 & \cipad{$-$0}{0.784}{0.973}{$-$0} & large better \\
Qwen3.5-122B-A10B & Phi-4-Reasoning & \phantom{0}$+$1.382 & \phantom{0}$+$0.563 & 1,086 / 190 & $<$0.001 & \cipad{$-$0}{1.276}{1.490}{$-$0} & large better \\
\midrule
Qwen2.5-72B & Phi-4 & \phantom{0}$-$0.267 & \phantom{0}$+$0.023 & 522 / 695 & $<$0.001 & \cipad{0}{$-$0.370}{$-$0.162}{0} & equivalent \\
Qwen2.5-72B & Qwen3-14B & \phantom{0}$-$0.256 & \phantom{0}$+$0.079 & 427 / 593 & $<$0.001 & \cipad{0}{$-$0.350}{$-$0.160}{0} & equivalent \\
\bottomrule
\end{tabular}
\end{adjustbox}
\caption{\textbf{Each large judge against the five most accurate SLMs}, paired on identical items, each judge at its best budget. $\Delta$ = large minus SLM accuracy in points, micro- and macro-averaged. Discordant = items only the large judge gets right / items only the SLM gets right. The 95\% interval is the within-dataset paired bootstrap on $\Delta$ micro. \emph{Equivalent} = TOST rejects at a $\pm$0.5-point margin; \emph{n.d.} = McNemar does not reject.}
\label{tab:app_paired}
\end{table*}

Table~\ref{tab:app_matrix} summarizes all $\binom{22}{2}{=}231$ comparisons among the 22 judges. After Benjamini--Hochberg correction 224 differences are significant, 23 pairs are equivalent within $\pm$0.5 points, and 16 are both: with 64{,}824 paired items, a gap of a few tenths of a point is often significant and still small, which is why the matrix reports the two tests separately. The four most accurate judges at their best budget, Qwen3.5-122B-A10B, Phi-4, Qwen3-14B, and Qwen2.5-72B, are pairwise equivalent.

\begin{table*}[tp]
\centering
\begin{adjustbox}{max width=\textwidth}
\footnotesize
\setlength{\tabcolsep}{2.2pt}
\begin{tabular}{rlccccccccccccccccccccc}
\toprule
 & & \textbf{1} & \textbf{2} & \textbf{3} & \textbf{4} & \textbf{5} & \textbf{6} & \textbf{7} & \textbf{8} & \textbf{9} & \textbf{10} & \textbf{11} & \textbf{12} & \textbf{13} & \textbf{14} & \textbf{15} & \textbf{16} & \textbf{17} & \textbf{18} & \textbf{19} & \textbf{20} & \textbf{21} \\
\midrule
1 & Qwen3.5-122B-A10B$^{L}$ &  &  &  &  &  &  &  &  &  &  &  &  &  &  &  &  &  &  &  &  &  \\
2 & Phi-4 & \hlcell$\approx$ &  &  &  &  &  &  &  &  &  &  &  &  &  &  &  &  &  &  &  &  \\
3 & Qwen3-14B & \hlcell$\approx$ & \hlcell$\circ$ &  &  &  &  &  &  &  &  &  &  &  &  &  &  &  &  &  &  &  \\
4 & Qwen2.5-72B$^{L}$ & \hlcell$\approx$ & \hlcell$\approx$ & \hlcell$\approx$ &  &  &  &  &  &  &  &  &  &  &  &  &  &  &  &  &  &  \\
5 & Llama-3.3-70B$^{L}$ & $-$ & \hlcell$\approx$ & \hlcell$\approx$ & \hlcell$\circ$ &  &  &  &  &  &  &  &  &  &  &  &  &  &  &  &  &  \\
6 & Qwen3-8B & $-$ & $-$ & $-$ & \hlcell$\approx$ & \hlcell$\approx$ &  &  &  &  &  &  &  &  &  &  &  &  &  &  &  &  \\
7 & GPT-OSS-120B$^{L}$ & $-$ & $-$ & $-$ & \hlcell$\approx$ & \hlcell$\approx$ & \hlcell$\approx$ &  &  &  &  &  &  &  &  &  &  &  &  &  &  &  \\
\bottomrule
\end{tabular}
\end{adjustbox}
\caption{\textbf{Paired verdicts for all 231 judge pairs}, judges sorted by best-budget accuracy; each cell compares the row judge with the column judge. $\circ$ = equivalent within $\pm$0.5 points and not distinguishable; $\approx$ = equivalent, with a significant difference; $+$ / $-$ = the row judge is significantly more / less accurate and the pair is not equivalent. A difference counts as significant when McNemar's $p<0.05$ and the paired interval excludes zero; all 224 such pairs also survive Benjamini--Hochberg correction. $^{L}$ = large judge.}
\label{tab:app_matrix}
\end{table*}

\paragraph{Macro averages.} Table~\ref{tab:app_macro} repeats every judge's accuracy as a macro average, the unweighted mean of the eight per-dataset accuracies. HellaSwag, one of the two largest datasets, is also the hardest, so macro averaging raises the judges above 88\% by 0.7 to 1.7 points at their best budget; with every judge at its best budget, macro minus micro ranges from $-$3.14 (Qwen2.5-3B) to $+$3.10 (Llama-3.2-1B). Under either average the leaders form one tight group: the seven most accurate judges by macro average, four large and three small, lie within 0.31 points of each other (90.80--91.11\%).

\begin{table*}[tp]
\centering
\begin{adjustbox}{max width=\textwidth}
\small
\rowcolors{2}{white}{slmrow}
\setlength{\tabcolsep}{4pt}
\begin{tabular}{llcccc}
\toprule
\textbf{Judge} & \textbf{Type} & $B$ & \textbf{Micro} & \textbf{Macro} & \textbf{Macro$-$micro} \\
\midrule
Qwen3.5-122B-A10B & Large & 10 & 90.30 & 91.11 & $+$0.82 \\
Phi-4 & SLM & 10 & 90.16 & 90.89 & $+$0.73 \\
Qwen3-14B & SLM & 10 & 90.15 & 90.83 & $+$0.68 \\
Qwen2.5-72B & Large & 10 & 89.89 & 90.91 & $+$1.02 \\
Qwen3.5-122B-A10B & Large & 8{,}192 & 89.82 & 91.05 & $+$1.24 \\
Llama-3.3-70B & Large & 10 & 89.81 & 90.80 & $+$0.99 \\
Qwen3-8B & SLM & 10 & 89.61 & 90.75 & $+$1.14 \\
GPT-OSS-120B & Large & 8{,}192 & 89.47 & 90.98 & $+$1.51 \\
Phi-4-Reasoning-Plus & SLM & 8{,}192 & 89.42 & 90.85 & $+$1.43 \\
Phi-4-Reasoning & SLM & 8{,}192 & 88.91 & 90.55 & $+$1.64 \\
Llama-3.3-70B & Large & 8{,}192 & 88.86 & 90.48 & $+$1.62 \\
GLM-4.5-Air & Large & 8{,}192 & 88.83 & 90.31 & $+$1.48 \\
Phi-4-mini & SLM & 10 & 88.75 & 89.44 & $+$0.69 \\
Qwen3-14B & SLM & 8{,}192 & 88.60 & 90.16 & $+$1.56 \\
Qwen3-4B & SLM & 8{,}192 & 88.44 & 90.18 & $+$1.74 \\
Qwen3-4B & SLM & 10 & 88.21 & 85.70 & $-$2.51 \\
Phi-4-mini & SLM & 8{,}192 & 88.12 & 89.00 & $+$0.88 \\
Qwen2.5-72B & Large & 8{,}192 & 88.10 & 90.11 & $+$2.01 \\
Qwen3-8B & SLM & 8{,}192 & 88.08 & 90.00 & $+$1.92 \\
\bottomrule
\end{tabular}
\hspace{1.2em}
\rowcolors{2}{white}{slmrow}
\begin{tabular}{llcccc}
\toprule
\textbf{Judge} & \textbf{Type} & $B$ & \textbf{Micro} & \textbf{Macro} & \textbf{Macro$-$micro} \\
\midrule
Phi-4 & SLM & 8{,}192 & 87.80 & 89.68 & $+$1.88 \\
Llama-3.1-8B & SLM & 10 & 87.29 & 88.52 & $+$1.22 \\
Qwen2.5-7B & SLM & 8{,}192 & 87.23 & 87.44 & $+$0.21 \\
Qwen3-1.7B & SLM & 8{,}192 & 86.56 & 88.27 & $+$1.71 \\
Qwen3-1.7B & SLM & 10 & 84.52 & 85.91 & $+$1.39 \\
Llama-3.1-8B & SLM & 8{,}192 & 84.20 & 84.41 & $+$0.22 \\
Phi-4-mini-Reasoning & SLM & 8{,}192 & 83.75 & 84.01 & $+$0.26 \\
Llama-3.2-3B & SLM & 8{,}192 & 82.14 & 83.34 & $+$1.21 \\
Qwen2.5-1.5B & SLM & 8{,}192 & 80.39 & 80.51 & $+$0.12 \\
Llama-3.2-3B & SLM & 10 & 79.92 & 82.91 & $+$2.99 \\
Qwen2.5-3B & SLM & 10 & 78.16 & 75.02 & $-$3.14 \\
Qwen2.5-7B & SLM & 10 & 78.01 & 73.54 & $-$4.47 \\
Qwen2.5-3B & SLM & 8{,}192 & 77.22 & 77.25 & $+$0.03 \\
Qwen3-0.6B & SLM & 8{,}192 & 76.79 & 78.94 & $+$2.15 \\
Qwen2.5-1.5B & SLM & 10 & 74.32 & 71.81 & $-$2.50 \\
DeepSeek-R1-Distill-70B & Large & 8{,}192 & 70.74 & 71.71 & $+$0.97 \\
Qwen3-0.6B & SLM & 10 & 67.99 & 72.58 & $+$4.58 \\
Llama-3.2-1B & SLM & 8{,}192 & 65.61 & 68.72 & $+$3.10 \\
Llama-3.2-1B & SLM & 10 & 43.25 & 43.75 & $+$0.50 \\
\bottomrule
\end{tabular}
\end{adjustbox}
\caption{\textbf{Micro and macro accuracy (\%)} for every judge and budget, sorted by micro accuracy. Micro weights each item equally (the convention throughout the paper); macro weights each of the eight datasets equally. Macro$-$micro is in points.}
\label{tab:app_macro}
\end{table*}

\subsection{The Qwen3 Ladder}

Table~\ref{tab:app_ladder} walks the Qwen3 family up in size, each judge at its best budget and each step paired on identical items. The gains shrink from 9.77 points (0.6B to 1.7B) to 0.54 (8B to 14B) and 0.23 (14B to 32B), the last inside the equivalence margin. Qwen3-235B was beyond our hardware. The next judge we could serve, Qwen3.5-122B-A10B, is a newer generation rather than a larger Qwen3 and a mixture of experts with about 10B active parameters, so the final row is not a clean scale step; it lands 0.08 points below Qwen3-32B, a significant but equivalent difference. Qwen3-32B is 0.22 points above Phi-4, also within the margin, at 2.15 times Phi-4's serving cost (Table~\ref{tab:cost}).

\begin{table}[!htbp]
\centering
\rowcolors{2}{white}{slmrow}
\begin{adjustbox}{max width=\columnwidth}
\small
\setlength{\tabcolsep}{4pt}
\begin{tabular}{lcccc}
\toprule
\textbf{\makecell{Step\\(Qwen3 sizes)}} & \textbf{\makecell{$\Delta$\\(points)}} & \textbf{\makecell{McNemar\\$p$}} & \textbf{95\% CI} & \textbf{TOST} \\
\midrule
0.6B $\to$ 1.7B & $+$9.77 & $<$0.001 & \cipad{$-$}{9.49}{10.05}{$-$} & no \\
1.7B $\to$ 4B & $+$1.88 & $<$0.001 & \cipad{$-$}{1.73}{2.03}{$-$0} & no \\
4B $\to$ 8B & $+$1.17 & $<$0.001 & \cipad{$-$}{1.05}{1.30}{$-$0} & no \\
8B $\to$ 14B & $+$0.54 & $<$0.001 & \cipad{$-$}{0.45}{0.63}{$-$0} & no \\
14B $\to$ 32B & $+$0.23 & $<$0.001 & \cipad{$-$}{0.15}{0.32}{$-$0} & yes \\
32B $\to$ Qwen3.5-122B & $-$0.08 & 0.038 & \cipad{}{$-$0.16}{$-$0.01}{0} & yes \\
Phi-4 $\to$ 32B & $+$0.22 & $<$0.001 & \cipad{$-$}{0.13}{0.32}{$-$0} & yes \\
\bottomrule
\end{tabular}
\end{adjustbox}
\caption{\textbf{The Qwen3 scaling ladder}, each judge at its best budget, each step paired on the same 64{,}824 items. TOST = equivalent within $\pm$0.5 points. The last row compares Qwen3-32B with Phi-4.}
\label{tab:app_ladder}
\end{table}

\subsection{Per-Dataset Accuracy}

Table~\ref{tab:app_large_per_dataset} gives the per-dataset accuracy for the six large judges and Qwen3-32B alongside the three strongest small ones, at every budget each supports. The pattern of \S\ref{sec:individual} repeats at scale: every judge except DeepSeek-R1-Distill-Llama-70B is strongest on GSM8K or MATH and weakest on HellaSwag, and no judge at any size clears 80\% there. DeepSeek-R1-Distill-Llama-70B, the one large judge whose instruction following falls well short (87.06\%), is weakest on ARC-Challenge (68.22\%).

\begin{table*}[tp]
\centering
\input{tables/41_large_per_dataset}
\caption{\textbf{Per-dataset closed-ended accuracy (\%)}, micro-averaged over both students. Five of the six large judges, and Qwen3-32B, show the same dataset profile as the strongest small ones.}
\label{tab:app_large_per_dataset}
\end{table*}

\subsection[The Math-to-General Gap at Scale]{The Math-to-General Gap Does Not Shrink With Scale}

Table~\ref{tab:app_domain_gap} reports the gap between the mean of the three math datasets and the mean of the three general-reasoning datasets, the same unweighted quantity \S\ref{sec:individual} uses to separate specialists from generalists. If generalization improved with parameters, the 70B--122B judges would show a smaller gap than the 14B ones. They do not: the five large judges other than DeepSeek-R1-Distill-Llama-70B sit at 8.5--11.2 points, Qwen3-32B at 11.1 and 8.5, and the strongest small judges at 8.5--12.0. DeepSeek-R1-Distill-Llama-70B shows no gap ($-$0.7) because it is weak on both. The general-reasoning ceiling is a property of the task.

\begin{table}[!htbp]
\centering
\rowcolors{2}{white}{slmrow}
\begin{adjustbox}{max width=\columnwidth}
\small
\setlength{\tabcolsep}{4pt}
\begin{tabular}{llccc}
\toprule
\textbf{Judge} & \textbf{Budget} & \textbf{Math} & \textbf{General} & \textbf{Gap} \\
\midrule
Llama-3.3-70B & $B{=}10$ & 96.6 & 86.8 & \phantom{$-$0}9.7 \\
Llama-3.3-70B & $B{=}8{,}192$ & 95.9 & 86.6 & \phantom{$-$0}9.3 \\
DeepSeek-R1-Distill-70B & $B{=}8{,}192$ & 71.9 & 72.6 & \phantom{0}$-$0.7 \\
Qwen2.5-72B & $B{=}10$ & 97.3 & 86.3 & \phantom{$-$}11.1 \\
Qwen2.5-72B & $B{=}8{,}192$ & 95.0 & 86.5 & \phantom{$-$0}8.5 \\
GLM-4.5-Air & $B{=}8{,}192$ & 95.8 & 86.4 & \phantom{$-$0}9.4 \\
GPT-OSS-120B & $B{=}8{,}192$ & 96.8 & 87.0 & \phantom{$-$0}9.9 \\
Qwen3.5-122B-A10B & $B{=}10$ & 97.7 & 86.5 & \phantom{$-$}11.2 \\
Qwen3.5-122B-A10B & $B{=}8{,}192$ & 97.1 & 86.9 & \phantom{$-$}10.2 \\
Qwen3-32B (ladder) & $B{=}10$ & 97.7 & 86.6 & \phantom{$-$}11.1 \\
Qwen3-32B (ladder) & $B{=}8{,}192$ & 95.0 & 86.4 & \phantom{$-$0}8.5 \\
Phi-4 & $B{=}10$ & 97.2 & 86.5 & \phantom{$-$}10.7 \\
Phi-4 & $B{=}8{,}192$ & 94.6 & 85.8 & \phantom{$-$0}8.8 \\
Qwen3-14B & $B{=}10$ & 97.8 & 85.8 & \phantom{$-$}12.0 \\
Qwen3-14B & $B{=}8{,}192$ & 95.3 & 86.8 & \phantom{$-$0}8.5 \\
Qwen3-8B & $B{=}10$ & 97.0 & 86.4 & \phantom{$-$}10.6 \\
Qwen3-8B & $B{=}8{,}192$ & 95.0 & 86.4 & \phantom{$-$0}8.6 \\
\bottomrule
\end{tabular}
\end{adjustbox}
\caption{\textbf{Math-to-general accuracy gap} (unweighted mean of GSM8K/GSM-Plus/MATH minus unweighted mean of HellaSwag/WinoGrande/TruthfulQA).}
\label{tab:app_domain_gap}
\end{table}

\subsection[MT-Bench Reference Reliability]{How Reliable Is the MT-Bench Reference?}

MT-Bench quality scoring has no programmatic ground truth, so every judge is scored by correlation with one large reference model, Qwen3.5-397B-A17B. To measure how reliable that reference is, a second large model, GPT-OSS-120B, scored the same conversations with the identical prompt. The two agree at Spearman $\rho{=}0.496$ on the Qwen2.5-32B student's answers and $\rho{=}0.652$ on the Llama-3.1-8B student's ($n{=}80$ each), a mean of 0.574 with a 95\% interval of $[0.450, 0.685]$ (2{,}000 resamples, drawing each student's 80 questions separately). Two models at 120B and 397B parameters, given identical instructions, agree only moderately, so MT-Bench agreement is a noisy target: the best judge, Qwen3.5-122B-A10B (0.592), and the best SLM, Qwen3-14B (0.516), both fall inside that interval. We therefore treat MT-Bench as secondary evidence and rest the open-ended comparison on SummEval's human labels (Appendix~\ref{app:openended_full}).

\subsection{Serving Cost}

Table~\ref{tab:cost} gives the serving-cost estimates behind \S\ref{sec:scaling}. A verdict at $B{=}10$ is charged 210 tokens (a 200-token prompt plus the budget) at the judge's decode rate, times the number of GPUs in its serving configuration (Appendix~\ref{app:models}), at \$0.55 per GPU-hour, the same rate for the small judges' 32\,GB V100s and the large judges' 44\,GiB cards, which favors the large judges. The decode rate is an approximation rather than a measurement: one law, 582\,$\mu$s plus 341\,$\mu$s per billion total parameters per token, applied to every judge, with mixture-of-experts judges charged on total parameters because every expert stays resident, and a small-batch penalty for the three Phi-4 reasoning judges. Only the ratios between judges matter for the conclusions. Llama-3.3-70B needs four times Phi-4's GPUs and decodes at under a quarter of its rate, so its verdicts cost 18 times more whatever the hourly rate; Qwen2.5-72B, on four GPUs, costs 9.4 times more. GPU counts are the tensor-parallel widths of each judge's 16-bit serving configuration (Appendix~\ref{app:models}). Served from its FP8 checkpoint on four GPUs instead of eight, Qwen3.5-122B-A10B would cost \$5.41 per thousand verdicts, 16 times Phi-4 rather than 32, and GLM-4.5-Air would halve in the same way.

\begin{table}[!htbp]
\centering
\input{tables/43_serving_cost}
\caption{\textbf{Serving cost per 1{,}000 verdicts at $B{=}10$}, from the approximate cost model above. $^\dagger$Judges evaluated at $B{=}8{,}192$ only (DeepSeek-R1-Distill-Llama-70B, GLM-4.5-Air, GPT-OSS-120B) are priced at that budget. The last row is the ladder judge.}
\label{tab:cost}
\end{table}

\section{Error Taxonomy}
\label{app:error_taxonomy}

Where do judge errors against the oracle come from? We collect every disagreement between a judge and the oracle across the full grid (both students, all eight datasets, both budgets where the judge supports both) and classify each one with a rule-based categorizer that inspects the ground truth, the extracted student answer, and the judge's response. The categories:

\begin{itemize}
    \item \textbf{Hallucinated comparison}: the judge asserts that the student's answer does or does not match the reference where the oracle says otherwise;
    \item \textbf{Critical-thinking failure}: the problem is unanswerable (GSM-Plus perturbations with no valid answer) and the judge accepts a fabricated one;
    \item \textbf{Format mismatch}: the student's answer could not be extracted or uses an incompatible labeling scheme, yet the judge issued a verdict;
    \item \textbf{Semantic equivalence}: the answers are equivalent under lenient matching (fractions, rounding, LaTeX variants) and the strict oracle disagrees;
    \item \textbf{Reasoning distraction}: the judge's response grades the quality of the reasoning rather than the correctness of the answer;
    \item \textbf{Instruction failure}: no parseable verdict;
    \item \textbf{Unclassified}: none of the rules fire (1.2--6.0\% per judge).
\end{itemize}

\begin{table*}[tp]
\centering
\input{tables/44_error_taxonomy}
\caption{\textbf{Disagreement rate and error-category distribution} (\% of each judge's disagreements), for the 16 SLM judges, the six large judges, and the ladder judge under one classifier. $^\dagger$Judges evaluated at $B{=}8{,}192$ only cover one budget, so their item count is half. Columns: CritT = critical-thinking failure, Fmt = format mismatch, SemEq = semantic equivalence, Reas = reasoning distraction, Instr = instruction failure, Halluc = hallucinated comparison, Uncl = unclassified.}
\label{tab:app_e8_dist}
\end{table*}

\begin{table}[!htbp]
\centering
\input{tables/45_error_direction}
\caption{\textbf{Error direction per judge}: accepting answers the oracle rejects (false accepts) vs.\ rejecting answers it accepts (false rejects). $^\dagger$As in Table~\ref{tab:app_e8_dist}.}
\label{tab:app_e8_dir}
\end{table}

Table~\ref{tab:app_e8_dist} shows one failure mode dominating everything else: hallucinated comparisons account for 68--94\% of the errors of every judge but one. The exception, DeepSeek-R1-Distill-Llama-70B, is at 45.7\% because unparseable verdicts make up another 44.2\% of its errors. Judges rarely fail to parse, rarely get distracted by reasoning quality, and rarely stumble on formatting, so most disagreements are direct conflicts with the oracle's verdict. Critical-thinking failures are the second mode for the strongest judges (9--14\%), which is what remains once the easy errors are gone: recognizing that a question has no valid answer is harder than comparing two answers.

\paragraph{Scale changes neither the rate nor the mode.} Two judges can agree on how often they are right and still be wrong about entirely different things, which an accuracy column cannot show. Running the same classifier over the large judges puts their error profiles alongside the small ones. Qwen3.5-122B-A10B disagrees with the oracle on 9.9\% of items, GPT-OSS-120B on 10.5\%, Llama-3.3-70B on 10.7\%, Qwen2.5-72B on 11.0\%, and GLM-4.5-Air on 11.2\%; Phi-4-Reasoning-Plus and Qwen3-14B sit at 10.6\%, Qwen3-32B at 10.7\%, and Phi-4 at 11.0\%. The top small judges and the large ones interleave rather than separating: Phi-4-Reasoning-Plus and Qwen3-14B fall between GPT-OSS-120B and Llama-3.3-70B, and Qwen2.5-72B and Phi-4 are both at 11.0\%. The composition matches too: hallucinated comparison accounts for 80.0--86.0\% of these five large judges' errors against 82.1--85.8\% for the top three SLMs, and critical-thinking failures for 8.5--14.5\% against 8.6--12.5\%. DeepSeek-R1-Distill-Llama-70B is the outlier at 29.3\%, and nearly half its errors are the instruction failures noted above, rather than a different kind of judging mistake. The dominant failure mode is the same at every scale we test, so the two groups agree not only on how often they match the oracle but on how they depart from it, which is consistent with the accuracy parity of \S\ref{sec:scaling}.

Error direction is family-dependent rather than universal (Table~\ref{tab:app_e8_dir}). Phi-4 and the Qwen3 family err heavily toward acceptance, from 2.8$\times$ to 30.7$\times$ more false accepts than false rejects, and so does every large judge (8.18$\times$ to 15.35$\times$), Qwen2.5-72B included. The small Qwen2.5 judges lean the opposite way, rejecting answers the oracle accepts more often than they accept ones it rejects (0.54--0.94$\times$), consistent with their low false-accept rates in the family-bias analysis (\S\ref{sec:scaling}). A deployment that cares about the direction of errors, not just their count, should pick the judge family accordingly.

\subsection{A Manual Audit of the Oracle}
\label{app:oracle_audit}

The oracle is a program, so where it and a judge disagree, either could be wrong. To measure which, we use 622 student answers labeled manually by the authors, each judged from the question, the reference, and the student's solution alone, answering one question per item: is the student's final answer correct? The labeling sheet showed no oracle verdict, judge verdict, or extracted answer, though it did name each item's sampling stratum. The labels are 468 correct, 145 incorrect, and 9 unclear. Six judges are audited: Phi-4, Qwen3-14B, Qwen3.5-122B-A10B, GPT-OSS-120B, Llama-3.3-70B, and Qwen2.5-72B. Items come from two strata, which are never pooled:

\begin{itemize}
    \item \textbf{Disagreements}: 89 to 95 of the judge's disagreements with the oracle, sampled within strata of answer type, budget, and error direction;
    \item \textbf{Unreadable answers}: 100 of the 437 student answers the extractor cannot read and the oracle scores \texttt{Incorrect} (Appendix~\ref{app:oracle_reliability}), the same 100 for every judge, drawn in turn from each of the eight student--dataset cells that contain such answers.
\end{itemize}

An item sampled for two judges is labeled once and scores both. Table~\ref{tab:app_oracle_audit} reports, per stratum, how often the program was wrong among the decided items, with a Wilson 95\% interval. On sampled disagreements the human label sides with the judge in 84.6--95.6\% of cases per judge. Beyond these two strata, a separate sample of MATH disagreements, where answer formats vary most, sides with the oracle in 28 of 42 decided judge--item pairs. Of the unreadable answers, 44 of 100 are correct (34.7--53.8\%), and the six judges get these items right in 79--100\% of cases. Because the sample takes about equally from each cell, it over-represents the small ones: 310 of the 437 are Llama-3.1-8B MATH solutions, and none of the 17 sampled from that cell is correct. Weighting each cell by its share of the 437 puts the correct share at 13.9\%, so the default \texttt{Incorrect} is wrong for about one unreadable answer in seven.

The audit bounds the oracle on contested items, not overall. The disagreement samples are stratified, so their rates cannot be multiplied by a judge's disagreement rate to estimate the oracle's overall error without reweighting. Because the same program scores every judge, the comparisons in the paper remain like for like, while absolute accuracies on contested items are conservative.

\begin{table*}[tp]
\centering
\input{tables/46_oracle_audit}
\caption{\textbf{Manual audit of the oracle}, per judge and stratum. Unclear = items that could not be decided, excluded from the percentages. Program wrong = share of decided items where the oracle's verdict contradicts the manual label (Wilson 95\% interval); Judge right = share where the judge's verdict agrees with it. On the disagreement stratum the program and the judge always disagree, so the two columns coincide.}
\label{tab:app_oracle_audit}
\end{table*}

\section{Family Bias: Full Matrices}
\label{app:self_preference}

This appendix expands the family-bias analysis of \S\ref{sec:scaling}. Thirteen judges, the 13 dual-setting SLMs, are scored against all five students, with every cell computed over the identical eight-dataset grid (32{,}412 items per judge, student, and budget). Family bias is only measurable for judge families that have a same-lineage student in the panel, so the family summary covers the eleven Llama, Qwen2.5, and Qwen3 judges; Phi-4 and Phi-4-mini appear in the matrices, but their family has no student. Table~\ref{tab:self_preference} summarizes the result by family.

\begin{table}[!htbp]
\centering
\input{tables/47_family_preference}
\caption{\textbf{False-accept rate (FAR) by judge family.} $\Delta$ = FAR on the same-family student minus FAR on the others; the solver-accuracy columns give the mean solver accuracy (\%) of the same-family student and of the rest of the panel. $\Delta$ mixes lineage with the solver-accuracy mismatch; Appendix~\ref{app:lineage} separates the two. The Phi-4 family has no same-lineage student and is not measurable.}
\label{tab:self_preference}
\end{table}

\subsection{Student Panel}

Table~\ref{tab:app_solver} lists every student used in this appendix, with its family and its own solver accuracy under the same oracle that scores the judges. Five of them carry the eight-dataset panel; Qwen3-14B (student) joins it for the lineage test of Appendix~\ref{app:lineage}; the last two join only the wider grid of the final subsection, since they cover five of the eight benchmarks. The spread is deliberate: on the eight-dataset grid the panel covers a 55-point capability range, which is what lets us separate ``the judge is lenient toward this student'' from ``this student is weak.'' DeepSeek-R1-Qwen3-8B's GSM-Plus, HellaSwag, and MATH solutions were stored without byte-level detokenization, which leaves 3{,}710 of its HellaSwag and 173 of its MATH answers unreadable and lowers its solver accuracy; every judge sees the same text, so its cells stay comparable across judges.

\begin{table}[!htbp]
\centering
\input{tables/48_student_solvers}
\caption{\textbf{Students used in the family-bias analysis}, with solver accuracy (\%) on the eight-dataset grid (32{,}412 problems) and on the five datasets the wider grid shares (6{,}818 problems). Qwen3-14B (student) is used on the eight-dataset grid only, for the lineage test; the last two students appear on the five-dataset grid only.}
\label{tab:app_solver}
\end{table}

\subsection{False-Accept Rate Matrices}

Tables~\ref{tab:app_far_t10} and~\ref{tab:app_far_t8192} give the full judge$\times$student false-accept matrices at both budgets. Reading across any row, the DeepSeek-R1-Qwen3-8B student draws the highest false-accept rates from most judges, including Llama and Qwen2.5 judges that share no lineage with it. It is the student Llama-3.1-8B accepts most readily when it is wrong, at 0.490 against 0.233 for Mistral-7B, and a Llama judge has no lineage to favor in a Qwen3 derivative. Leniency follows the student's answer style and quality, not the judge's family.

\begin{table*}[tp]
\centering
\input{tables/49_far_b10}
\caption{\textbf{False-accept rate at $B{=}10$}: probability the judge accepts an oracle-incorrect answer, per judge and student (all eight datasets).}
\label{tab:app_far_t10}
\end{table*}

\begin{table*}[tp]
\centering
\input{tables/50_far_b8192}
\caption{\textbf{False-accept rate at $B{=}8{,}192$}, as in Table~\ref{tab:app_far_t10}.}
\label{tab:app_far_t8192}
\end{table*}

\subsection{False-Reject Rate Matrices}

A judge can err in two directions, and the false-accept matrices only show one of them. Tables~\ref{tab:app_frr_t10} and~\ref{tab:app_frr_t8192} give the false-reject rate, the probability that a judge rejects an answer the oracle accepts, over the same grid.

The two directions separate by family rather than by lineage, which is the same conclusion the accept side reaches from the other end. At $B{=}10$ the Qwen2.5 judges reject correct answers on 13.1--66.2\% of items while the Qwen3 judges stay between 0.1\% and 39.6\%, and Qwen3-0.6B is the extreme case at 0.1--0.6\%: it accepts nearly everything, which is also why its false-accept rates are the highest in the panel. Each family sits at a characteristic operating point on the accept-reject trade-off, and that point does not move when the student happens to share its lineage. Appendix~\ref{app:error_taxonomy} reaches the same split from the error counts.

\begin{table*}[tp]
\centering
\input{tables/51_frr_b10}
\caption{\textbf{False-reject rate at $B{=}10$}: probability the judge rejects an oracle-correct answer, per judge and student (all eight datasets).}
\label{tab:app_frr_t10}
\end{table*}

\begin{table*}[tp]
\centering
\input{tables/52_frr_b8192}
\caption{\textbf{False-reject rate at $B{=}8{,}192$}, as in Table~\ref{tab:app_frr_t10}.}
\label{tab:app_frr_t8192}
\end{table*}

\subsection{Judge Accuracy per Student}

Table~\ref{tab:app_e2_acc} shows plain judge accuracy on each student at $B{=}10$. Among the eleven capable judges (all but Llama-3.2-1B and Qwen3-0.6B) the per-student means span 30.4 points while the per-judge means span 8.8, so who produced the answers matters more than three times as much as which judge grades them. A Qwen judge does not lift a Qwen-lineage student above where a Llama judge places it.

\begin{table*}[tp]
\centering
\input{tables/53_judge_student_accuracy}
\caption{\textbf{Judge accuracy (\%) per student at $B{=}10$}, all eight datasets.}
\label{tab:app_e2_acc}
\end{table*}

\subsection{Do Judges Agree on the Ordering?}

If leniency were a family effect, judges of different families would rank the students differently. Table~\ref{tab:app_e2_ordering} measures that directly, as the Kendall $\tau$ between the orderings two judges produce, averaged over every pair, and against the students' own solver accuracy.

The budget decides the answer. At $B{=}8{,}192$ the eleven capable judges produce only two distinct orderings of the five students and agree at $\tau{=}0.96$, and the ordering they agree on is the students' solver accuracy at $\tau{=}0.98$. Ten of the eleven reproduce it exactly; the eleventh, Llama-3.2-3B, transposes the adjacent pair Mistral-7B and Llama-3.1-8B, which is the single swap a $\tau$ of 0.8 over five students allows. At $B{=}10$ the same judges fall to $\tau{=}0.52$ over four orderings, which is the pattern of \S\ref{sec:robustness} and Appendix~\ref{app:protocols} again: the reasoning budget buys agreement, not accuracy.

\begin{table}[!htbp]
\centering
\input{tables/54_student_ordering}
\caption{\textbf{Agreement between judges on how to rank the students}, over the capable judges of each panel. \textbf{Orderings} counts the distinct rankings produced; \textbf{Mean $\tau$} averages Kendall's $\tau$ over every judge pair; the last column compares each judge's ranking against the students' own solver accuracy.}
\label{tab:app_e2_ordering}
\end{table}

\subsection[Lineage vs.\ Student Quality]{Separating Lineage From Student Quality}
\label{app:lineage}

In the five-student panel, family and student quality are entangled: the one Qwen3-lineage student, DeepSeek-R1-Qwen3-8B, is also the weakest (29.4\% solver accuracy), so a Qwen3 judge's high false-accept rate on it could reflect either. To separate the two, we add a strong Qwen3-family student, Qwen3-14B (59.3\% solver accuracy), and score the eleven Llama, Qwen2.5, and Qwen3 judges on all six students over the same eight datasets (Tables~\ref{tab:app_full8p_far_t10} and~\ref{tab:app_full8p_far_t8192}).

For each family $F$ with lineage student $s$, the lineage effect is a difference in differences of false-accept rate: how much more $F$'s judges accept wrong answers from $s$ than from the students outside $F$, minus the same quantity for the judges outside $F$. Subtracting the other judges' difference removes how hard $s$'s wrong answers are to catch, and subtracting each group's rate on the other students removes its general leniency. Table~\ref{tab:app_lineage} gives every term. At $B{=}10$ the effect is $+0.010$ for Qwen3, $-0.075$ for Llama, and $+0.010$ for Qwen2.5, so no family favors its own lineage at the quick verdict. With room to reason the Qwen3 judges grow more lenient toward the Qwen3 student than other families do ($+0.206$), and the Llama and Qwen2.5 judges slightly so ($+0.089$ and $+0.045$). It cannot inflate any accuracy in the paper, since accepting a wrong answer counts as an error against the oracle.

\begin{table*}[tp]
\centering
\input{tables/55_lineage_did}
\caption{\textbf{Lineage effect on the false-accept rate}, as a difference in differences over the six-student panel. \emph{own} = mean FAR on the family's lineage student; \emph{others} = mean FAR on the students outside the family. DiD = (family own $-$ family others) $-$ (other own $-$ other others).}
\label{tab:app_lineage}
\end{table*}

\begin{table*}[tp]
\centering
\input{tables/56_lineage_far_b10}
\caption{\textbf{False-accept rate at $B{=}10$ on the six-student panel}, including the strong Qwen3-14B student (all eight datasets).}
\label{tab:app_full8p_far_t10}
\end{table*}

\begin{table*}[tp]
\centering
\input{tables/57_lineage_far_b8192}
\caption{\textbf{False-accept rate at $B{=}8{,}192$ on the six-student panel}.}
\label{tab:app_full8p_far_t8192}
\end{table*}

\subsection{A Wider Panel}

The eight-dataset grid is what caps the main panel at five students, since the two remaining students cover only five of the eight datasets. Trading dataset coverage for panel width gives a second grid of the same thirteen judges against seven students on those five datasets (6{,}818 items per cell), which adds Qwen3.5-35B and GLM-4.7-Flash as students. Table~\ref{tab:app_e2_wide_acc} gives the full 13$\times$7 accuracy matrix and Table~\ref{tab:app_e2_wide_family} repeats the family summary on it.

\begin{table*}[tp]
\centering
\input{tables/58_wide_panel_accuracy}
\caption{\textbf{Judge accuracy (\%) per student on the wider 13$\times$7 grid at $B{=}10$}, over the five datasets all seven students share. Accuracy is micro-averaged over every judgment record, including the unparseable ones, exactly as in Table~\ref{tab:individual}; the last column is each judge's mean across students and the last row each student's mean across judges.}
\label{tab:app_e2_wide_acc}
\end{table*}

The conclusion does not move. On this grid Llama and Qwen2.5 are stricter on their own lineage at both budgets, by 0.275 and 0.092 at $B{=}10$, and Qwen3 remains the exception in the same direction and for the same reason, its two lineage students averaging 62.9\% solver accuracy against 73.6\% for the rest. The accuracy matrix says the same thing from the other side. The student means span 25.8 points (58.7 to 84.5), and the six judges that average above 75\% span 4.2 among themselves (75.3 to 79.4); every one of those six puts the same two students at the top of its row, Qwen2.5-32B and Gemma-4-31B, and the same two at the bottom, Qwen3.5-35B and GLM-4.7-Flash, whatever family they belong to. The ordering result also survives the wider grid: at $B{=}8{,}192$ the eleven capable judges again produce two orderings and agree at $\tau{=}0.95$ (Table~\ref{tab:app_e2_ordering}). Their agreement with the students' own solver accuracy is looser here ($\tau{=}0.68$), which the narrower grid explains: on five datasets the seven students compress into a 28-point band rather than the 55-point one of Table~\ref{tab:app_solver}, and adjacent pairs are correspondingly easier to transpose.

\begin{table}[!htbp]
\centering
\input{tables/59_wide_panel_family}
\caption{\textbf{False-accept rate by judge family on the wider 13$\times$7 grid}, over the five datasets it covers uniformly. Columns are as in Table~\ref{tab:self_preference}.}
\label{tab:app_e2_wide_family}
\end{table}

\section{An Adaptive Budget Policy}
\label{app:adaptive_budget}

The budget sweep of Appendix~\ref{app:budget_factorial} raises a practical question: if accuracy plateaus by 512 tokens and $B{=}10$ is three orders of magnitude cheaper, can a policy pick budgets per item instead of globally? The natural proposal is a per-domain budget $B^{*}(\mathcal{D})$, since the sweep shows different task types reaching their knee at different budgets. We test that proposal against a second one suggested by the factor decomposition of \S\ref{sec:individual}, and simulate both item by item over the 13 dual-setting judges and all eight datasets, 842{,}712 judge decisions per policy:

\begin{itemize}
    \item \textbf{Always $B{=}10$} and \textbf{always $B{=}8{,}192$}, the two fixed baselines;
    \item \textbf{Per-domain knee budget}: for each judge and dataset the knee is the smallest sweep budget within 0.5 points of its best accuracy, and each answer type (numeric, LaTeX, multiple choice) gets the median of its knees, rounded up to a sweep budget;
    \item \textbf{Escalate on failure}: every item starts at $B{=}10$; only when the verdict is unparseable is that single item retried at $B{=}256$ with the reasoned prompt, as in the sweep. This is deployable as-is, since an unparseable verdict is observable without knowing the right answer.
\end{itemize}

\begin{table}[!htbp]
\centering
\input{tables/60_budget_policies}
\caption{\textbf{Budget policies} simulated per item over 13 judges and eight datasets. Relative cost counts output tokens only, normalized to always-$B{=}8{,}192$.}
\label{tab:app_policies}
\end{table}

Table~\ref{tab:app_policies} gives the outcome. The per-domain policy reaches 77.86\%, which is 5.24 points below always-$B{=}8{,}192$ and 2.08 points below always asking for a quick verdict, while costing 3.6 times as much as the latter, so it is worse than the quick verdict on both accuracy and cost. As the factor decomposition shows, spending more tokens on an item does not by itself make a judge more accurate, so allocating tokens by domain buys nothing, and the mid-range budgets the knee analysis selects fall inside the 64--256 token region where truncated reasoning does the most damage.

Escalation works because, as the same decomposition shows, what a small budget costs a judge is unparseable output rather than faulty reasoning, and escalation retries an item only when its verdict cannot be read. Just 0.76\% of items trigger a retry, and the parseable-verdict rate rises from 99.24\% to 99.80\% for 20\% more than the always-quick cost, recovering $+0.39$ points.

Neither policy matches full reasoning on accuracy. Always reasoning buys 3.16 points over always-quick at 819 times the output tokens, while escalation gives up 2.78 points against always-$B{=}8{,}192$ for a 99.85\% cost reduction, about 1/685 of the output tokens. Which of these trades is acceptable depends on the deployment. An adaptive policy should retry unreadable verdicts rather than allocate budget by domain.

\section{Reproducibility and Licensing}
\makeatletter\setlength{\@fptop}{0pt}\setlength{\@fpsep}{14pt}\makeatother
\label{app:reproducibility}

\subsection{Code and Data Availability}

{\raggedright The complete \slmjury{} framework is publicly available under the Apache 2.0 license at \url{https://github.com/anishh15/SLMJury}. An interactive leaderboard with all evaluation results is hosted at \url{https://anishh15.github.io/SLMJury/}, and the framework is also distributed as a Python package on PyPI (\url{https://pypi.org/project/slmjury/}). The raw judge verdicts behind every table and the analysis scripts that turn those verdicts into the numbers reported here will be released alongside it.\par}

The study is built so that it can be reproduced without our code. Every component is either public or fully specified in this appendix: the judges and the two main students are released checkpoints identified by exact HuggingFace name (Appendix~\ref{app:models}), and the other six students are public releases, the benchmarks are public datasets (Appendix~\ref{app:datasets}), the judging instructions are reproduced verbatim rather than described (Appendix~\ref{app:prompts}), the verdict parser is given as an explicit cascade (Appendix~\ref{app:verdict_cascade}), and the correctness oracle is stated as an algorithm (Appendices~\ref{app:answer_extraction} and~\ref{app:normalization}). Judge decoding is greedy except in thinking mode, which samples at one fixed setting; students, the debate's Variant~B, and the temperature condition of the decomposition also sample (below). An independent implementation that follows this appendix should recover our numbers from the same checkpoints, and the conventions most likely to cause a mismatch are stated explicitly below, together with the resolution a single run actually has.

\paragraph{Aggregation conventions.} Two choices affect every reported number. First, accuracy is \textit{micro}-averaged: correct judgments are summed over all items and divided by the total item count, so each dataset contributes in proportion to its size rather than each dataset contributing equally. Second, the denominator includes every judgment record, counting an unparseable verdict as incorrect rather than discarding it. Reporting instruction-following rate alongside accuracy (Table~\ref{tab:individual}) makes the size of that effect visible for each judge. Macro averages, which weight each dataset equally, are reported beside the micro averages for every judge in Table~\ref{tab:app_macro}.

\subsection{Run-to-Run Variation}
\label{app:run_variance}

Greedy decoding is reproducible in the sampler, not in the serving stack. vLLM batches requests continuously, and the composition of a batch sets the order in which the floating-point reductions inside each matrix multiplication accumulate. A token whose two most likely continuations are almost tied can land either way from one run to the next, and one flipped token sends the rest of the generation down a different path. The single-factor decomposition of Appendix~\ref{app:budget_factorial} produced its own baselines, which leaves us two independent runs of the thirteen dual-setting judges at both budgets, and of Phi-4-mini-Reasoning at $B{=}8{,}192$, over the same items, prompts, parser, and oracle, so the size of this effect can be measured rather than assumed.

Table~\ref{tab:app_run_variance} reports how often the two runs disagree, and capability sets the scale. At $B{=}10$, where every judge decodes greedily, the six judges above 87\% accuracy change at most 0.04\% of their verdicts between runs, while Llama-3.2-1B changes 9.45\%. At $B{=}8{,}192$ the two strongest judges that still decode greedily, Phi-4 and Phi-4-mini, sit at 0.08\% and 0.06\%, and Llama-3.2-1B reaches 12.71\%. A judge with wide logit margins has nothing close enough to a tie to flip, while a judge that is unsure decides many items on small ones.

The text moves far more than the verdict it carries. Phi-4 rewrites 7.43\% of its reasoned responses between runs yet changes its mind on 0.08\% of them, arriving at the same answer along a different route. Judges with thinking enabled sample at $\tau{=}0.6$ at $B{=}8{,}192$ instead of decoding greedily, so their variation there is a property of the sampler and is marked separately.

\begin{table}[!htbp]
\centering
\input{tables/61_run_variance}
\caption{\textbf{Disagreement between two independent runs} of the same configuration, as a percentage of items. \textbf{Verdict} is how often the parsed verdict differs; \textbf{Text} is how often the response differs at all. $^\dagger$Thinking enabled, so decoding samples at $\tau{=}0.6$ rather than greedily.}
\label{tab:app_run_variance}
\end{table}

This sets a floor on what one measurement resolves. For the configurations carrying our conclusions, Phi-4 and Qwen3-14B at $B{=}10$, that floor is a hundredth of a point, far inside the $\pm$0.5-point equivalence margin of \S\ref{sec:metrics}. It is larger where thinking judges sample (1.84\% of verdicts for Qwen3-14B at $B{=}8{,}192$) and for the weakest judges (up to 12.71\% for Llama-3.2-1B and 14.95\% for Qwen3-0.6B), which is a reason to treat their figures as approximate and to separate them by several points before calling one better than another.

\subsection{Installation and Usage}

\slmjury{} can be installed directly from PyPI:

{\small
\begin{verbatim}
pip install slmjury
\end{verbatim}
}

\noindent Optional extras are available for GPU inference and oracle scoring:

{\small
\begin{verbatim}
pip install slmjury[vllm]     # GPU inference
pip install slmjury[together] # Oracle API
pip install slmjury[full]     # All extras
\end{verbatim}
}

\noindent\textbf{Python API quickstart.} The following example demonstrates the three-step pipeline (solve, judge, evaluate):

{\small
\begin{verbatim}
from slmjury import load_dataset
from slmjury.core.solver import StudentSolver
from slmjury.core.judge import JudgeModel
from slmjury.core.evaluator import JudgeEvaluator

# Step 1: Solve problems with a student model
problems = load_dataset("gsm8k")
solver = StudentSolver("qwen2.5-32b")
results = solver.solve_batch(problems, "gsm8k")
solver.save_results(results, "gsm8k")
solver.cleanup()

# Step 2: Judge the solutions
judge = JudgeModel("qwen3-4b")
judgements = judge.evaluate_batch(
    results, max_tokens=10
)
judge.save_results(
    judgements, "qwen2.5-32b", "gsm8k", 10
)
judge.cleanup()

# Step 3: Evaluate judge accuracy
evaluator = JudgeEvaluator(
    "qwen3-4b", "qwen2.5-32b",
    "gsm8k", 10, judgements
)
summary = evaluator.evaluate()
\end{verbatim}
}

\noindent\textbf{CLI scripts.} Equivalent command-line scripts are provided for batch experiments:

{\small
\begin{verbatim}
python scripts/run_student.py \
    --model qwen2.5-32b --datasets gsm8k
python scripts/run_judge.py \
    --judge qwen3-4b --max-tokens 10 8192
python scripts/run_evaluation.py
\end{verbatim}
}

\noindent Full documentation, including advanced features (majority voting, multi-agent debate, persona evaluation), is available in the repository README.

\subsection{Reproducibility Checklist}

To reproduce our experiments, the following are required:
\begin{enumerate}
    \item \textbf{Hardware}: The core experiments ran on an NVIDIA DGX-1 server with 8 $\times$ Tesla V100-SXM2-32GB GPUs; the small judges use 1--2 GPUs each, and models at 0.6B--4B run on a single GPU. The large judges, the ladder judge, and most extended analyses ran on A40-, L40S-, A100-, and H200-class GPUs. With 16-bit weights on 44\,GiB GPUs Qwen2.5-72B uses 4, Qwen3-32B 2, and every other large judge 8, the GPU counts that enter the cost estimate of Appendix~\ref{app:large_judges}.
    \item \textbf{Software}: Python 3.10+ (3.11+ for the analysis scripts), vLLM $\geq$ 0.8.5 (0.24.0 for the large judges and the extended analyses), Transformers $\geq$ 4.57, SymPy 1.13.1 with \texttt{antlr4-python3-runtime} 4.11, PyTorch $\geq$ 2.6 with CUDA 12.4.
    \item \textbf{Model Access}: All 22 judges (16 small, 6 large), the ladder judge Qwen3-32B, and all 8 students are publicly available on HuggingFace under the licenses listed in Table~\ref{tab:app_licenses}; GLM-4.5-Air and Qwen3.5-122B-A10B are served from their vendors' FP8 checkpoints and GPT-OSS-120B from its MXFP4 release.
    \item \textbf{Datasets}: All 8 closed-ended benchmarks and both open-ended benchmarks (SummEval, MT-Bench) are public, at the versions given in Appendix~\ref{app:datasets}.
    \item \textbf{Evaluation grid}: Three grids appear in the paper and each number belongs to exactly one. Every main-text result outside the family-bias analysis averages 2 students $\times$ 8 datasets, giving $N{=}64{,}824$ judgments per judge and budget. The family-bias panel of \S\ref{sec:scaling} uses 5 students $\times$ 8 datasets at 32{,}412 judgments per judge, student, and budget, extended to 6 students for the lineage test, and its wider grid uses 7 students $\times$ 5 datasets at 6{,}818. The prompt-formulation, stress-variant, and protocol analyses use the full 2-student, 8-dataset grid.
    \item \textbf{Determinism}: Non-thinking experiments use \texttt{temperature=0.0}, which fixes the sampler, although batched serving still moves a small fraction of verdicts between runs (Appendix~\ref{app:run_variance}). GPT-OSS-120B is decoded the same way at $B{=}8{,}192$. Thinking-enabled models use \texttt{temperature=0.6}, \texttt{top\_p=0.95}, \texttt{top\_k=20}, \texttt{min\_p=0}, the Qwen3 model card's setting, applied to every thinking judge, and are therefore reproducible in distribution rather than bit for bit; no sampling seed is set beyond the serving library's default. Multi-agent debate Variant~B uses temperatures 0.0,~0.4, and~0.9.
\end{enumerate}

\subsection{Model Licenses}

Table~\ref{tab:app_licenses} lists the license of each model family, covering the judges and the students alike.

\subsection{Dataset Licenses}

Table~\ref{tab:app_dataset_licenses} lists the license of each dataset.

\begin{table}[!htb]
\centering
\input{tables/62_model_licenses}
\caption{\textbf{Model licenses.} All models are used for inference only, and all licenses permit research use. The Llama row covers the 3.1, 3.2, and 3.3 releases, each under its own version of the community license. $^\dagger$The Qwen2.5 family is mixed: the releases we use are Apache 2.0 except the 3B and 72B checkpoints, which carry the Qwen license. $^\ddagger$DeepSeek-R1-Distill-Llama-70B is derived from Llama-3.3-70B-Instruct and remains subject to the Llama 3.3 license.}
\label{tab:app_licenses}

\vspace{8pt}
\centering
\small
\begin{adjustbox}{max width=\columnwidth}
\rowcolors{2}{white}{slmrow}
\begin{tabular}{ll}
\toprule
\textbf{Dataset} & \textbf{License} \\
\midrule
GSM8K & MIT \\
GSM-Plus & CC-BY-SA 4.0 \\
MATH & MIT \\
ARC (Easy/Challenge) & CC-BY-SA 4.0 \\
HellaSwag & MIT \\
WinoGrande & CC-BY 4.0 \\
TruthfulQA & Apache 2.0 \\
SummEval & MIT \\
MT-Bench & CC-BY 4.0 \\
\bottomrule
\end{tabular}
\end{adjustbox}
\caption{\textbf{Dataset licenses.} All datasets are publicly available for research use.}
\label{tab:app_dataset_licenses}
\end{table}

\end{document}